\documentclass[10pt,twocolumn,letterpaper]{article}

\usepackage[pagenumbers]{wacv} 

\usepackage{booktabs,threeparttable}
\usepackage{array} 
\newcolumntype{C}[1]{>{\centering\arraybackslash}p{#1}} 

\usepackage{epsfig}
\usepackage{tcolorbox}
\usepackage{graphicx}
\usepackage{algorithm}
\usepackage{algcompatible}
\usepackage{algpseudocode}
\usepackage{enumitem}
\usepackage{amsmath}
\usepackage{amssymb}
\usepackage{tabularx}
\usepackage{pifont}
\usepackage{placeins}
\usepackage{adjustbox}
\usepackage{subcaption}
\usepackage{tikz}
\usepackage{xcolor}
\usepackage{rotating}

\usetikzlibrary{shapes.geometric, arrows, calc}

\tikzstyle{block} = [rectangle, draw, text centered, rounded corners, minimum height=1cm, minimum width=1cm, text width=1.5cm]
\tikzstyle{arrow} = [thick,->,>=stealth]

\usepackage{ocgx2}    

\newcounter{toggleimages}

\newcommand{\ToggleImages}[3][]{%
  \stepcounter{toggleimages}%
  \edef\toggleA{toggleA-\thetoggleimages}%
  \edef\toggleB{toggleB-\thetoggleimages}%
  \let\toggleboxA\undefined
  \let\toggleboxB\undefined
  \newsavebox{\toggleboxA}%
  \newsavebox{\toggleboxB}%
  \savebox{\toggleboxA}{\includegraphics[#1]{#2}}%
  \savebox{\toggleboxB}{\includegraphics[#1]{#3}}%
  \begin{picture}(\wd\toggleboxA,\ht\toggleboxA)
    \put(0,0){%
      \begin{ocg}{Image A \thetoggleimages}{\toggleA}{1}%
        \switchocg{\toggleA,\toggleB}{\usebox{\toggleboxA}}%
      \end{ocg}%
    }
    \put(0,0){%
      \begin{ocg}{Image B \thetoggleimages}{\toggleB}{0}%
        \switchocg{\toggleA,\toggleB}{\usebox{\toggleboxB}}%
      \end{ocg}%
    }
  \end{picture}%
}

\definecolor{wacvblue}{rgb}{0.21,0.49,0.74}
\usepackage[pagebackref,breaklinks,colorlinks,allcolors=wacvblue]{hyperref}

\definecolor{mydarkgreen}{RGB}{0,110,0}
\newcommand{\cmark}{\textcolor{mydarkgreen}{\ding{51}}} 
\newcommand{\xmark}{\textcolor{red}{\ding{55}}} 

\def\wacvPaperID{1098} 
\def\confName{WACV}
\def\confYear{2026}

\title{Pairwise Matching of Intermediate Representations for Fine-grained Explainability}

\author{Lauren Shrack$^{1, \dagger}$, Timm Haucke$^{1}$, Antoine Salaün$^{1}$, Arjun Subramonian$^{2}$, Sara Beery$^{1}$ \\
$^{1}$Massachusetts Institute of Technology, $^{2}$University of California, Los Angeles \\
$^{\dagger}$ \small{corresponding author: \texttt{lshrack@mit.edu}}}

\begin{document}
\maketitle
\begin{abstract}
The differences between images belonging to fine-grained categories are often subtle and highly localized, and existing explainability techniques for deep learning models are often too diffuse to provide useful and interpretable explanations. 
We propose a new explainability method (PAIR-X) that leverages both intermediate model activations and backpropagated relevance scores to generate fine-grained, highly-localized pairwise visual explanations. We use animal and building re-identification (re-ID) as a primary case study of our method, and we demonstrate qualitatively improved results over a diverse set of explainability baselines on 35 public re-ID datasets. In interviews, animal re-ID experts found PAIR-X to be a meaningful improvement over existing baselines for deep model explainability, and suggested that its visualizations would be directly applicable to their work. We also propose a novel quantitative evaluation metric for our method, and demonstrate that PAIR-X visualizations appear more plausible for correct image matches than incorrect ones even when the model similarity score for the pairs is the same. By improving interpretability, PAIR-X enables humans to better distinguish correct and incorrect matches.\footnote{Our code is available at: \url{https://github.com/pairx-explains/pairx}}
\end{abstract}
    
\section{Introduction}
\begin{figure*}

    \centering
    \begin{tabular}{m{.13\textwidth}|m{.24\textwidth}m{.24\textwidth}m{.24\textwidth}}
        \toprule
        \textbf{Technique} & \textbf{Giraffes} & \textbf{Cows} & \textbf{Buildings} \\
        \midrule
        \textbf{Classical} &
        \includegraphics[width=.24\textwidth]{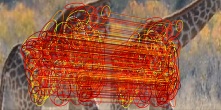} & \includegraphics[width=.24\textwidth]{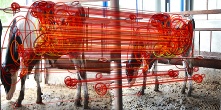} & \includegraphics[width=.24\textwidth]{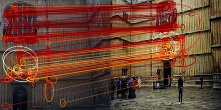}\\
        \textbf{KPCA-CAM} &
        \includegraphics[width=.24\textwidth]{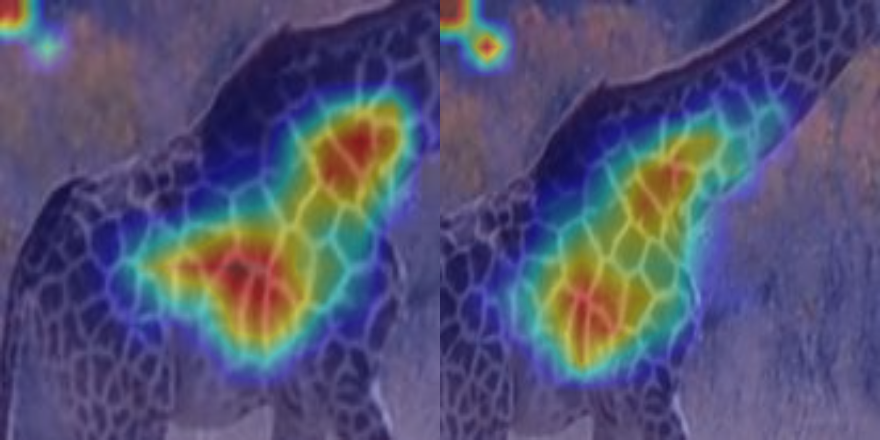} & \includegraphics[width=.24\textwidth]{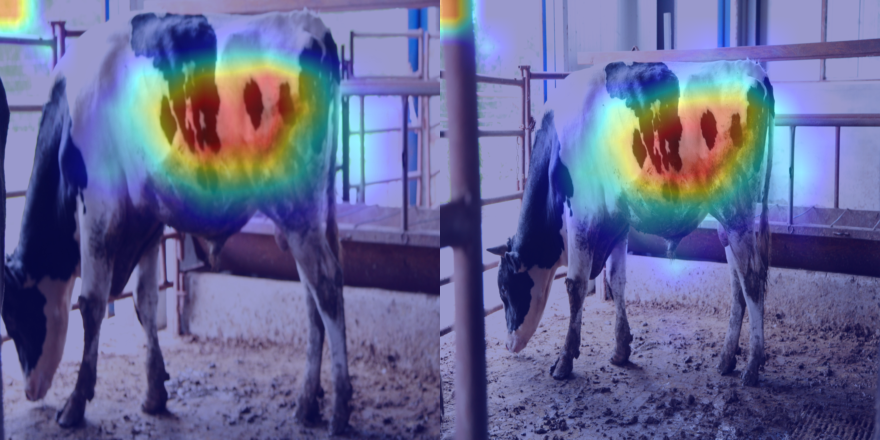} & \includegraphics[width=.24\textwidth]{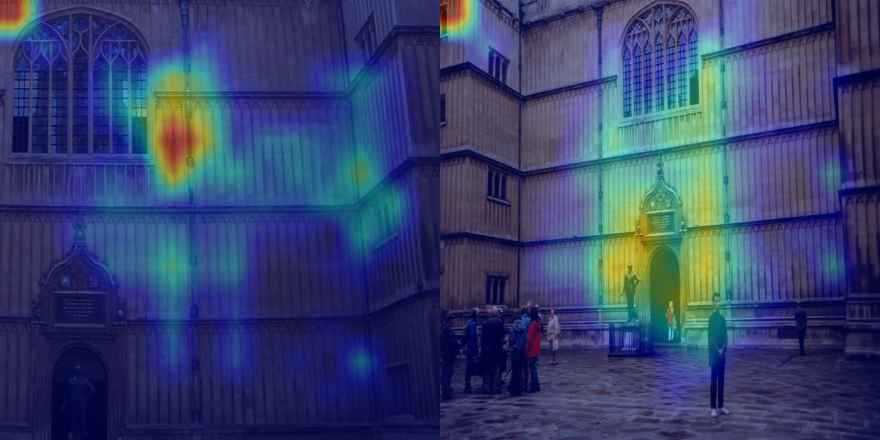}\\
        \textbf{Kernel SHAP} &
        \includegraphics[width=.24\textwidth]{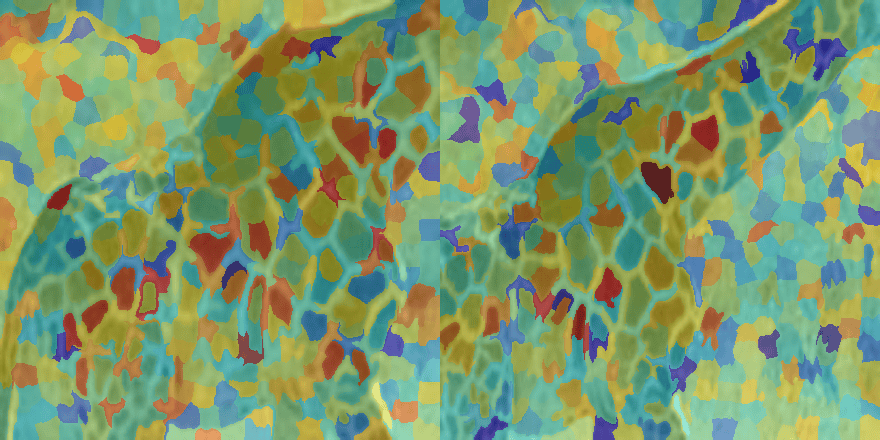} & \includegraphics[width=.24\textwidth]{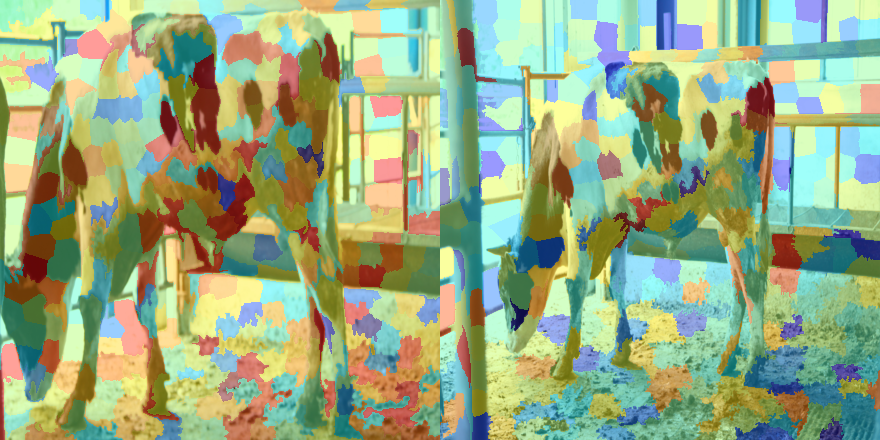} & \includegraphics[width=.24\textwidth]{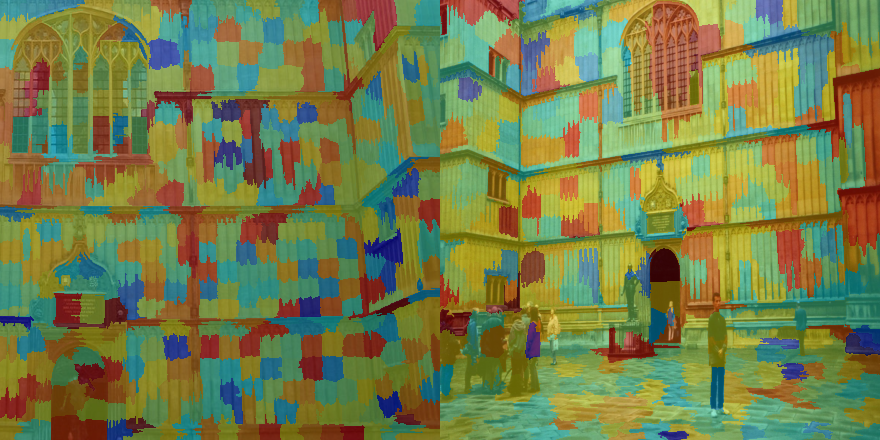}\\
        \textbf{LRP} &
        \includegraphics[width=.24\textwidth]{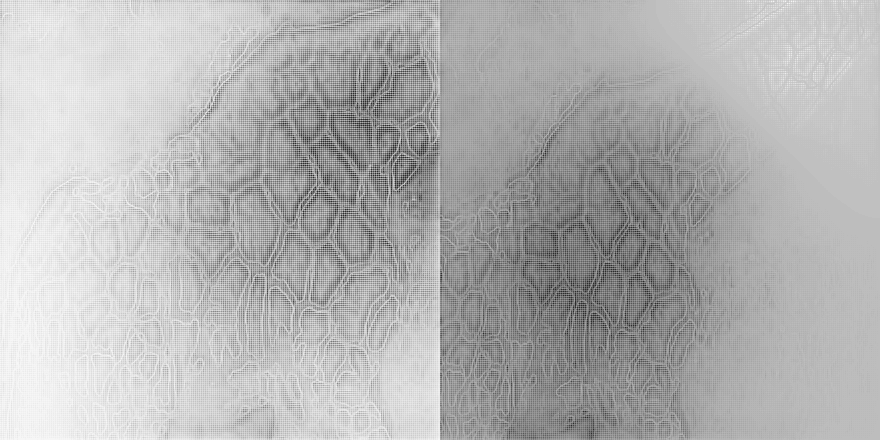} & \includegraphics[width=.24\textwidth]{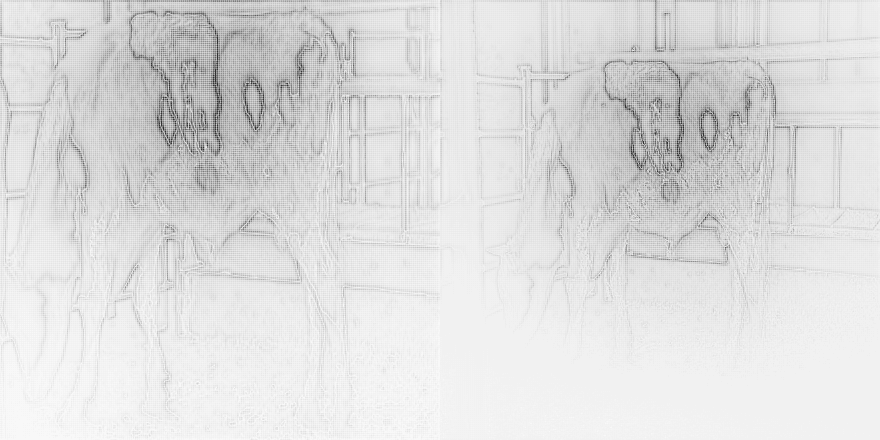} & \includegraphics[width=.24\textwidth]{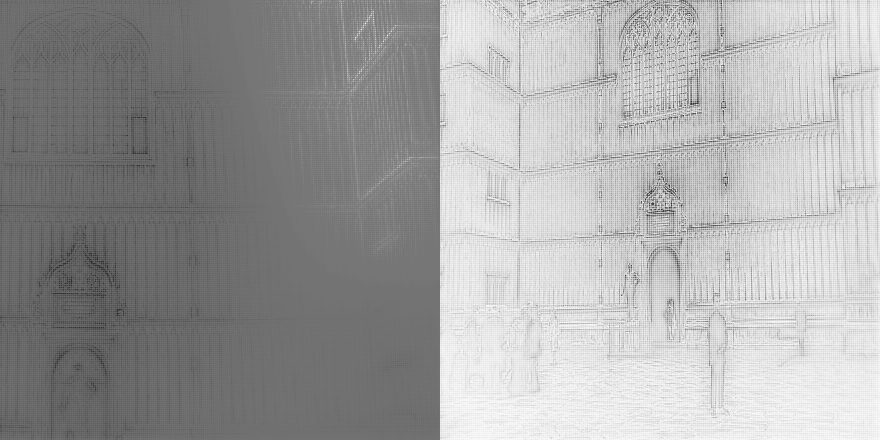}\\
        \textbf{Point-to-Point Activation Intensities} &
        \includegraphics[width=.24\textwidth]{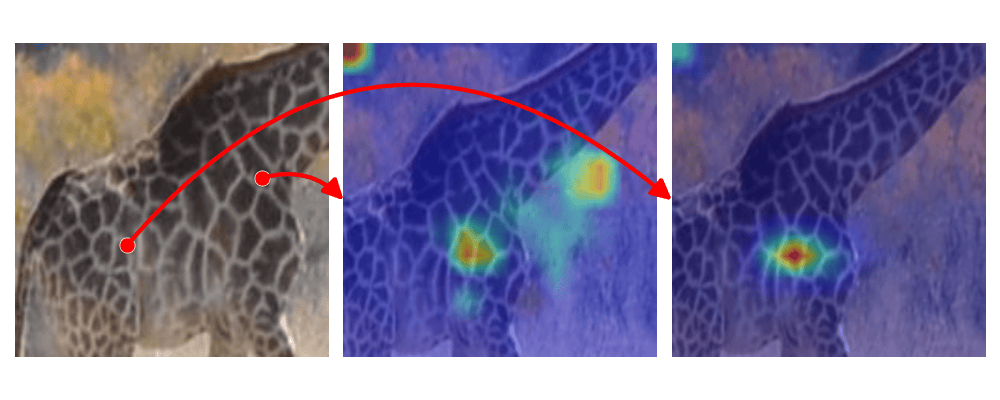} & \includegraphics[width=.24\textwidth]{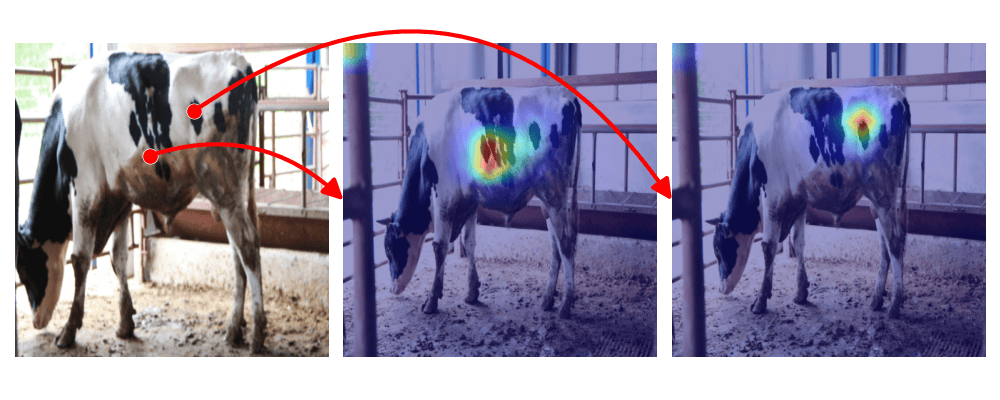} & \includegraphics[width=.24\textwidth]{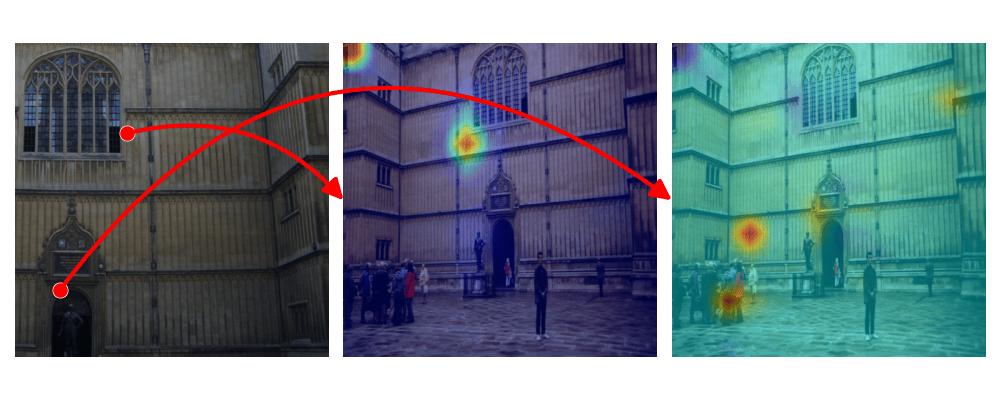}\\
        \textbf{PAIR-X} &
        \ToggleImages[width=.24\textwidth]{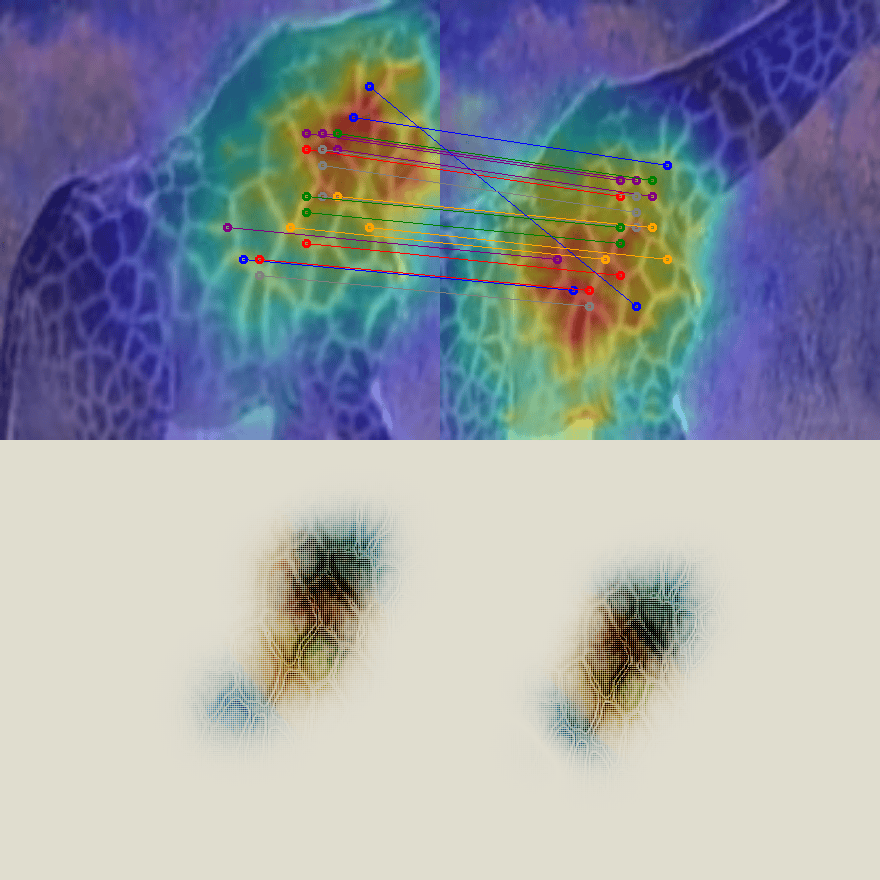}{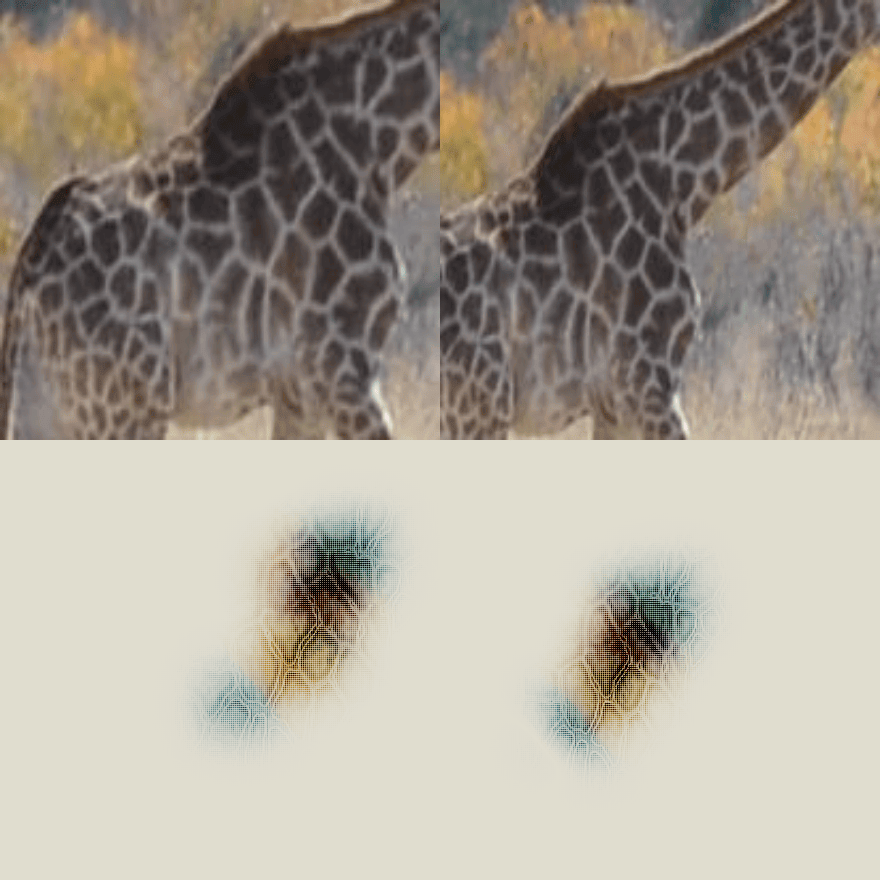} & \ToggleImages[width=.24\textwidth]{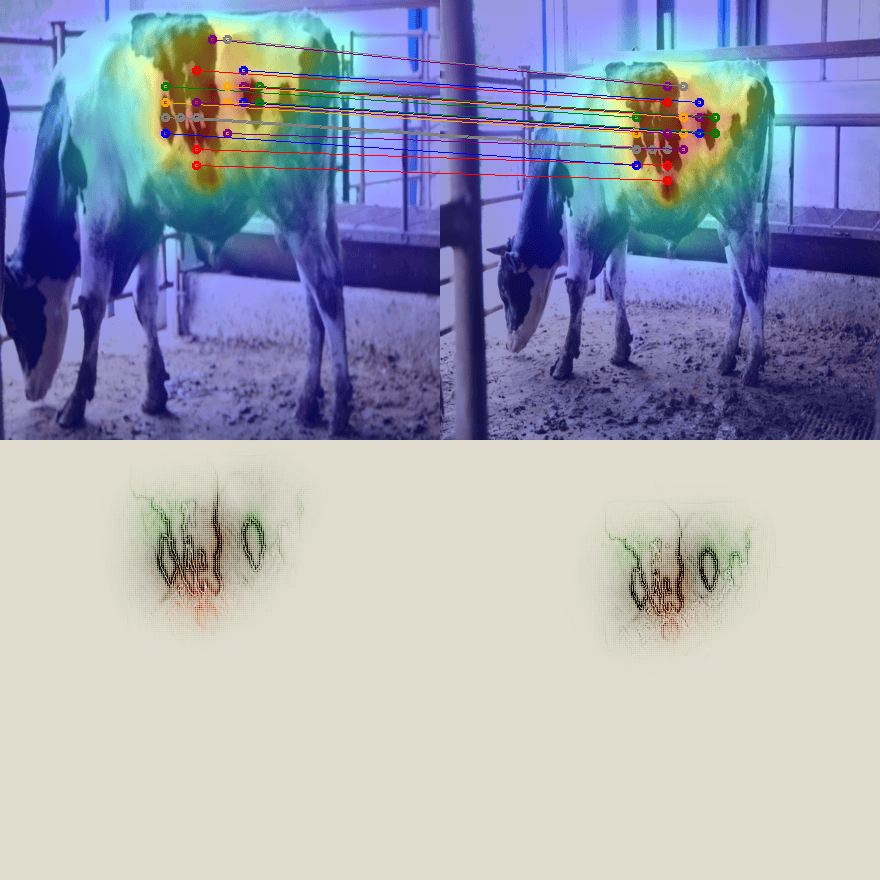}{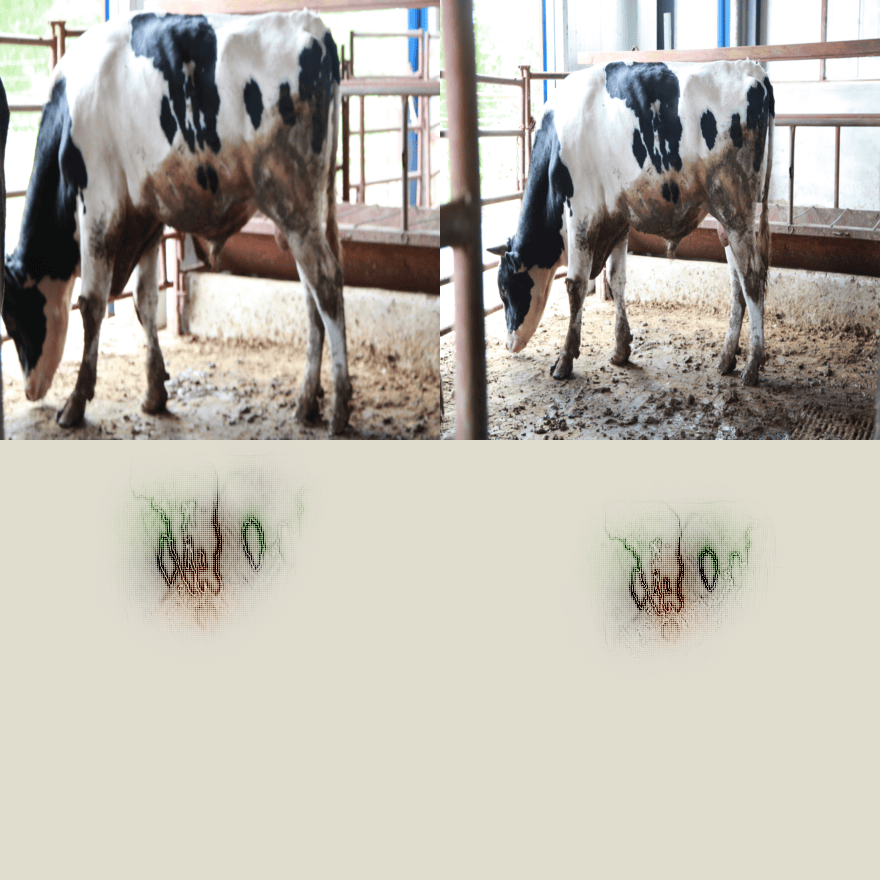} & \ToggleImages[width=.24\textwidth]{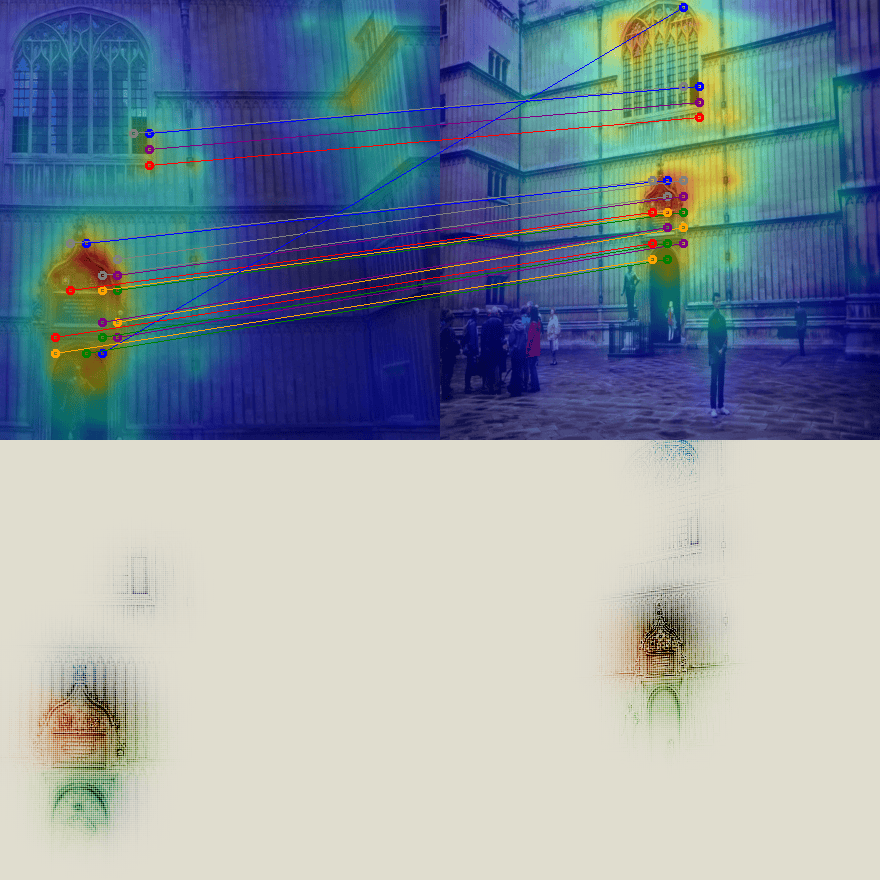}{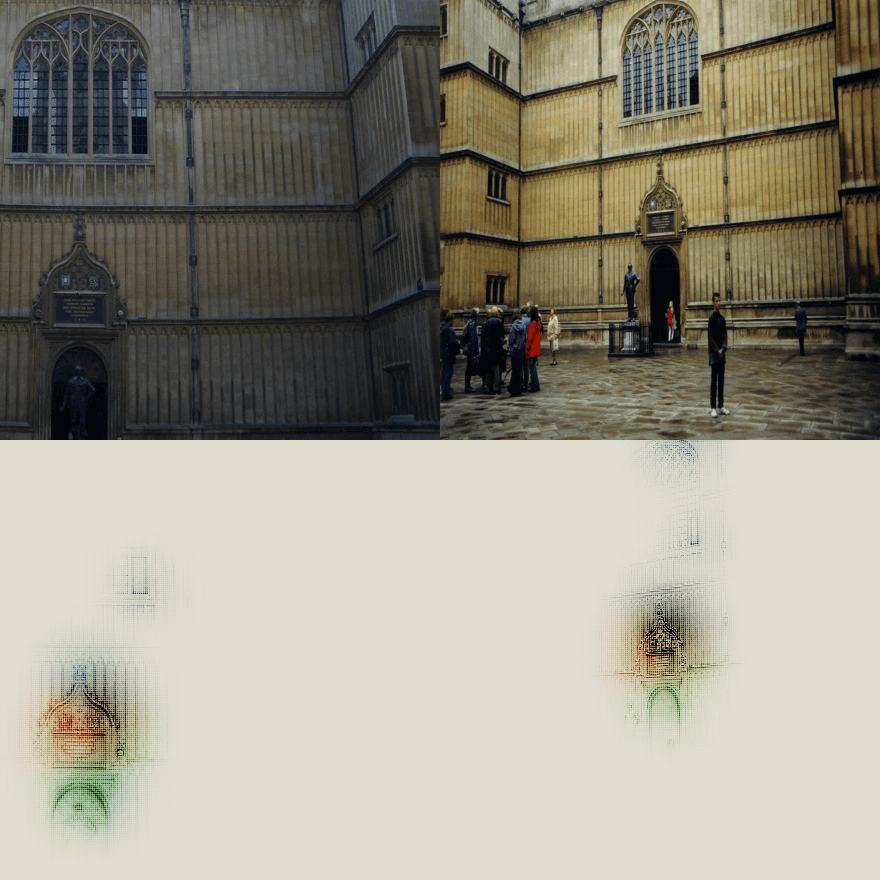}\\
        \hline
    \end{tabular}
    \caption{Qualitative comparison of  explainability techniques across image pairs selected randomly from three datasets. See Supplementary for a more expansive justification of baselines, including an ablation over 12 CAM-based techniques and a hyperparameter search for SHAP.}
    \label{fig:methods_ablation}

\end{figure*}

Similarity-based deep metric learning has proven to be highly effective for a variety of tasks, particularly fine-grained tasks including image retrieval~\cite{wang2014learning}, facial recognition~\cite{Hu_2014_CVPR,schroff2015facenet}, and open-set categorization problems such as animal re-identification (re-ID)~\cite{cermak2023wildlifedatasetsopensourcetoolkitanimal, schneider_2018_reid, Haurum_2020_zebra_fish, Andrew_2021}. However, for robust, trustworthy deployment of these systems, interpretability is key. Existing explainability techniques are often insufficient, producing coarse visualizations that do not adequately capture the fine-grained details important to many tasks~\cite{CRP}. As shown in Figure \ref{fig:methods_ablation}, this makes it difficult to precisely interpret which factors contribute most to the predicted similarity between a given pair of images~\cite{zhu2021visualexplanationdeepmetric}. 


\begin{figure}
\centering
\captionsetup{aboveskip=12pt, belowskip=-8pt, font=small}
\subfloat[Closest correct match]{\ToggleImages[width=0.49\columnwidth]{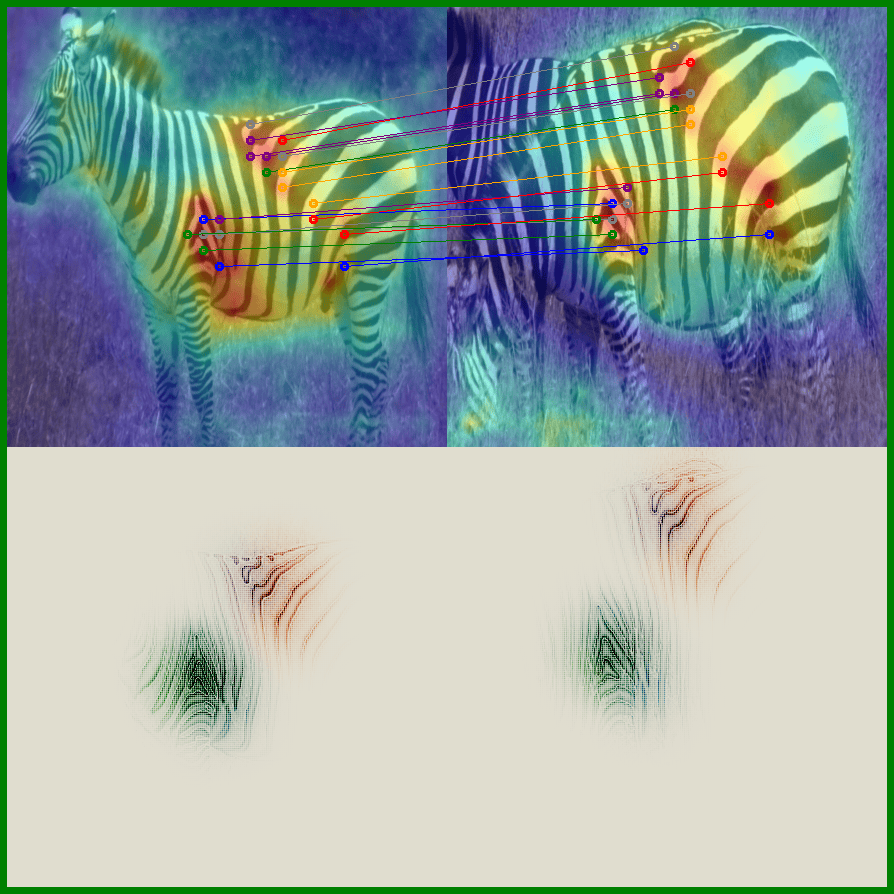}{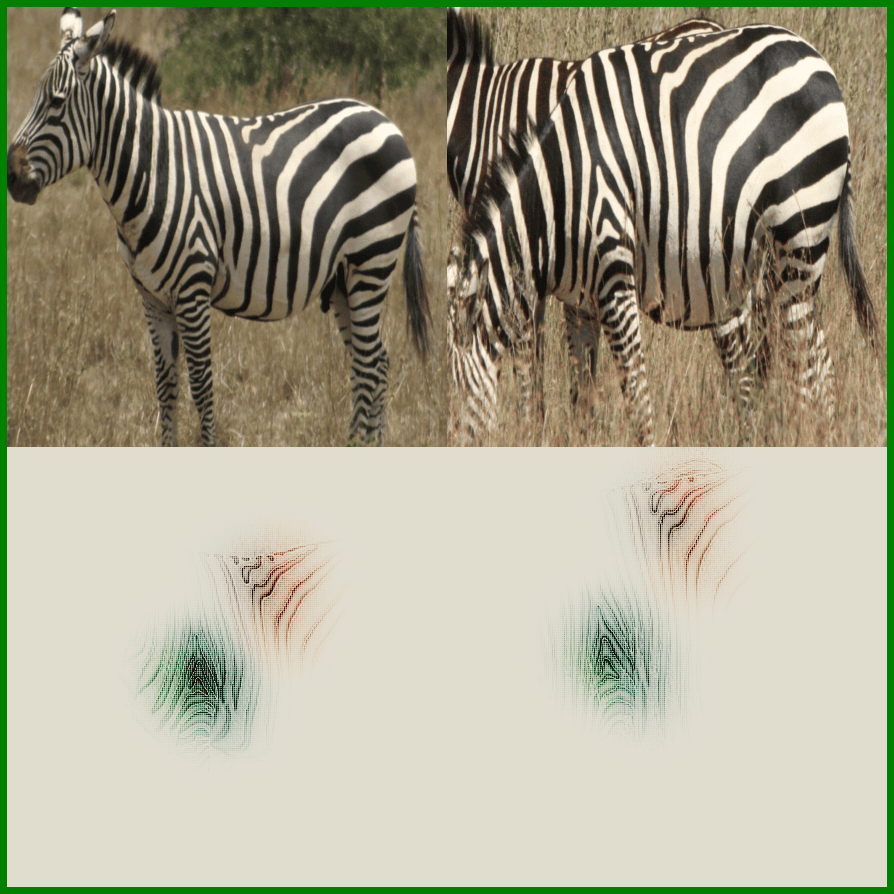}}
\hspace{0.01\columnwidth}
\subfloat[Closest incorrect match]{\ToggleImages[width=0.49\columnwidth]{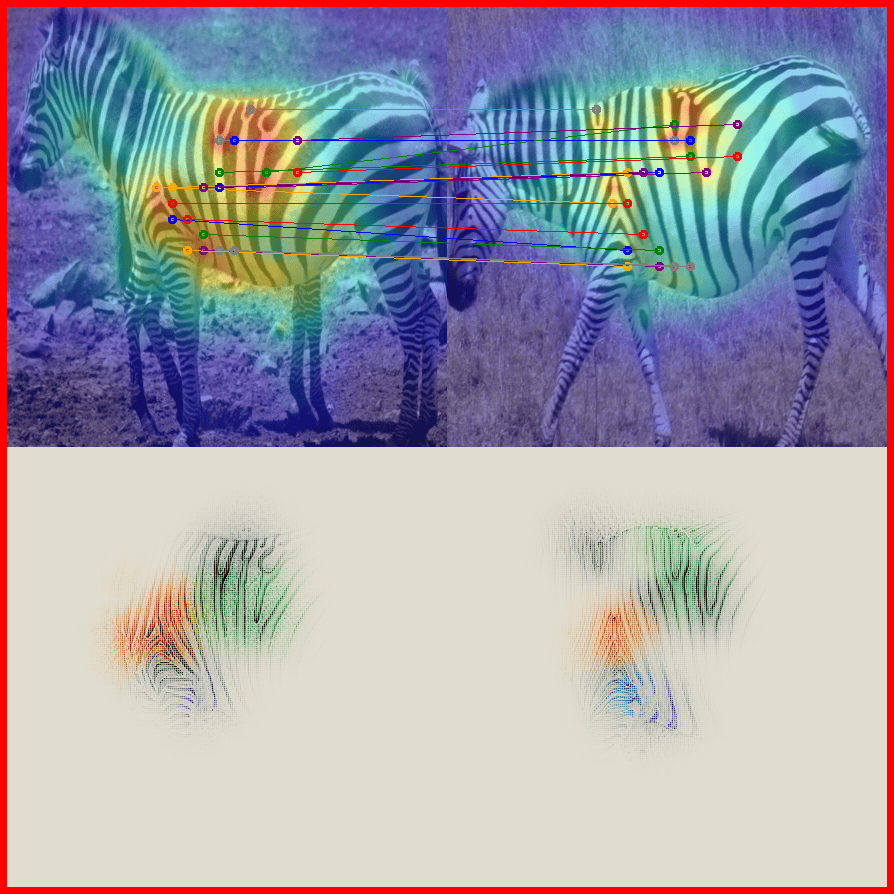}{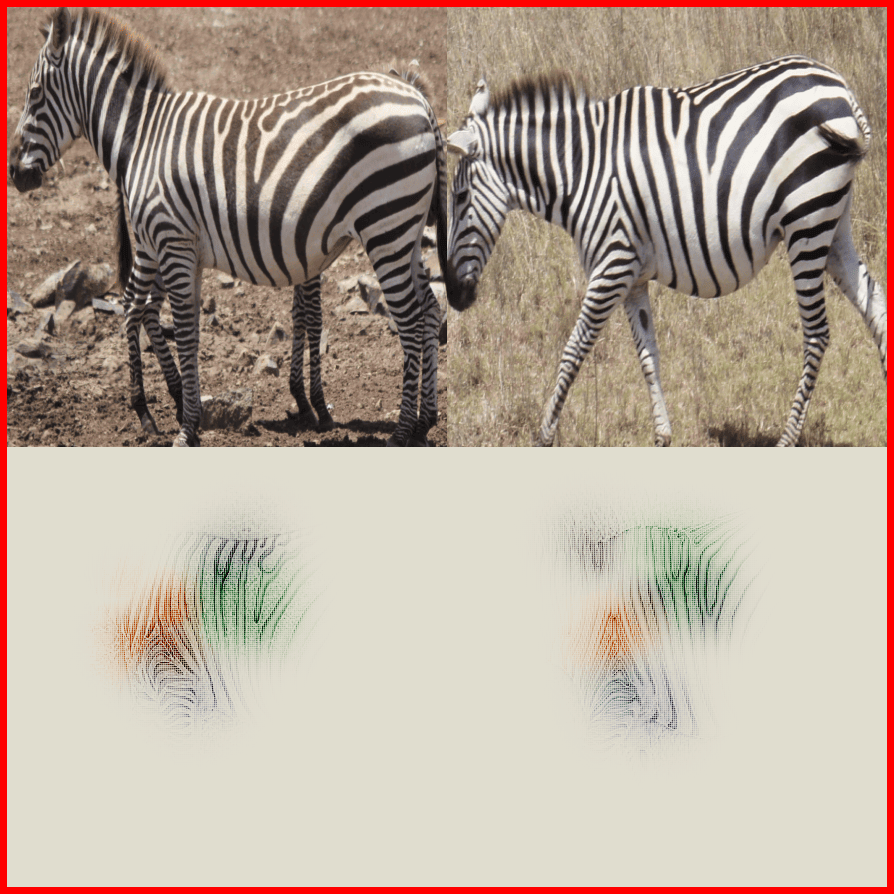}}
\vspace{10pt}

\subfloat[Closest correct match]{\ToggleImages[width=0.49\columnwidth]{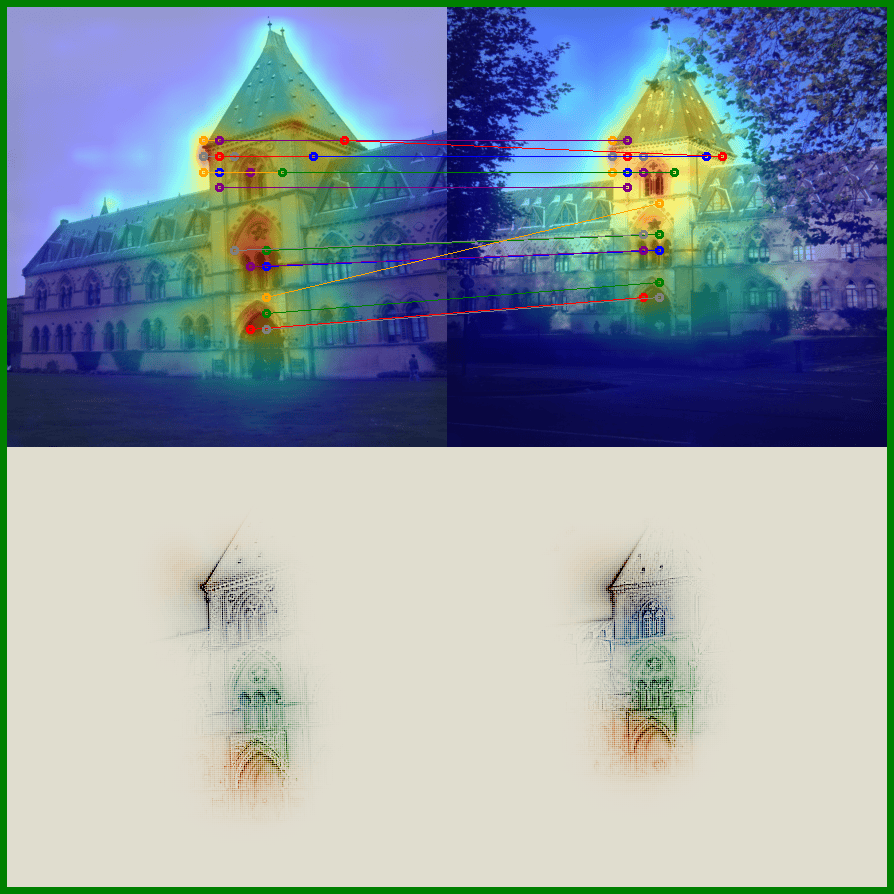}{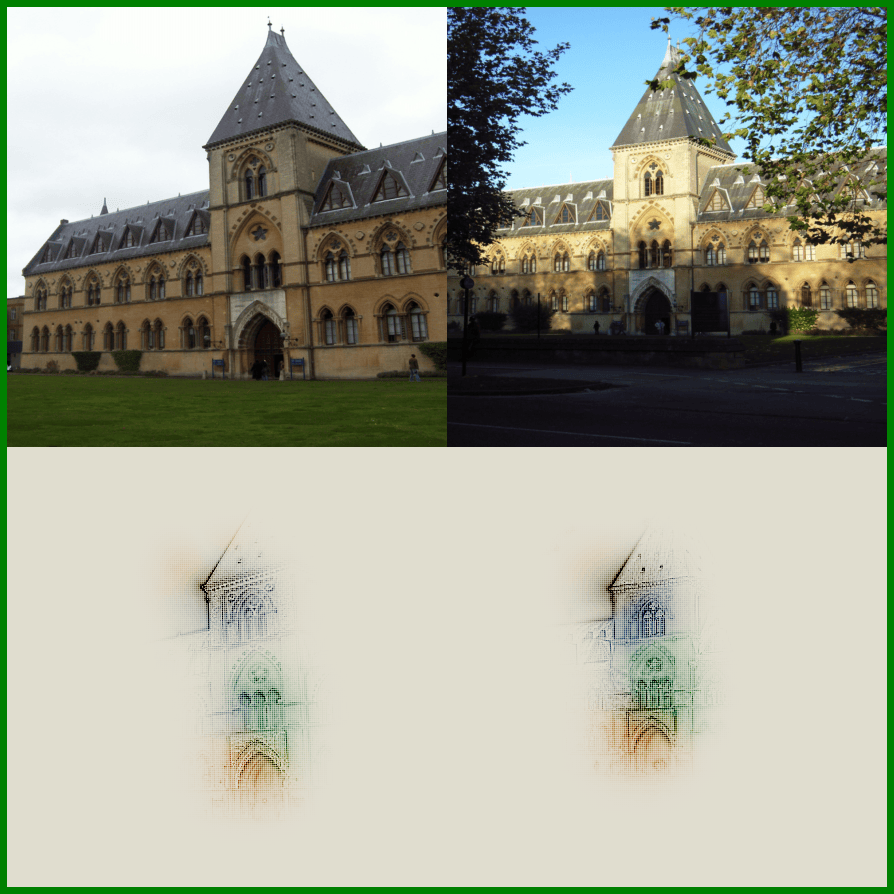}}
\hspace{0.01\columnwidth}
\subfloat[Closest incorrect match]{\ToggleImages[width=0.49\columnwidth]{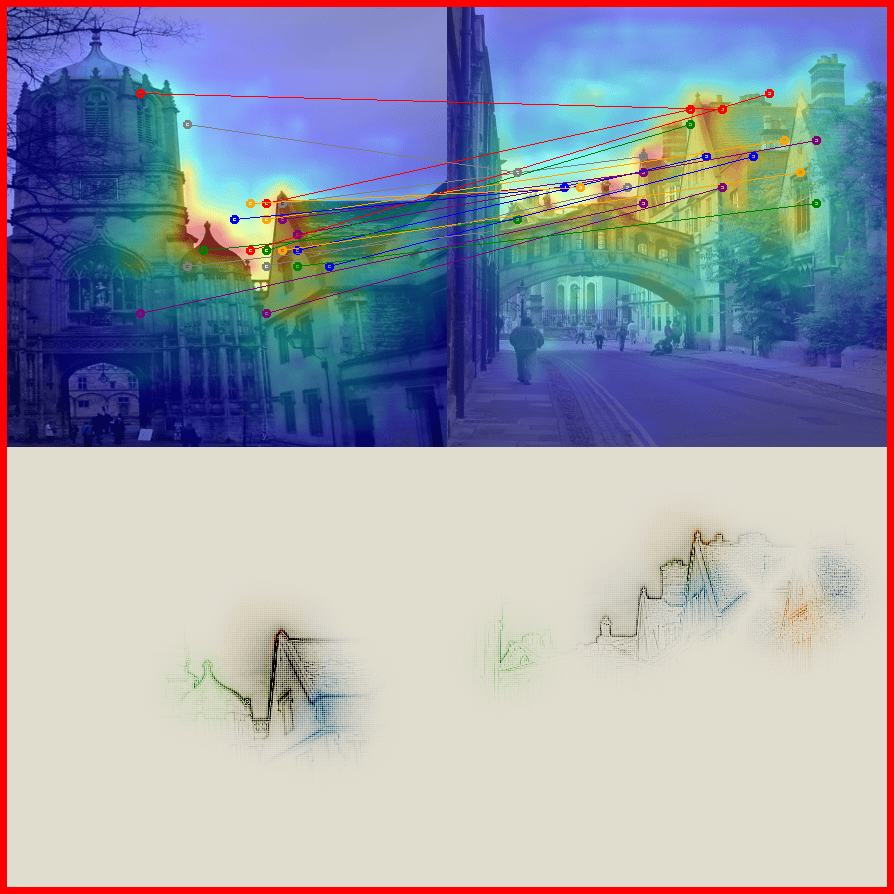}{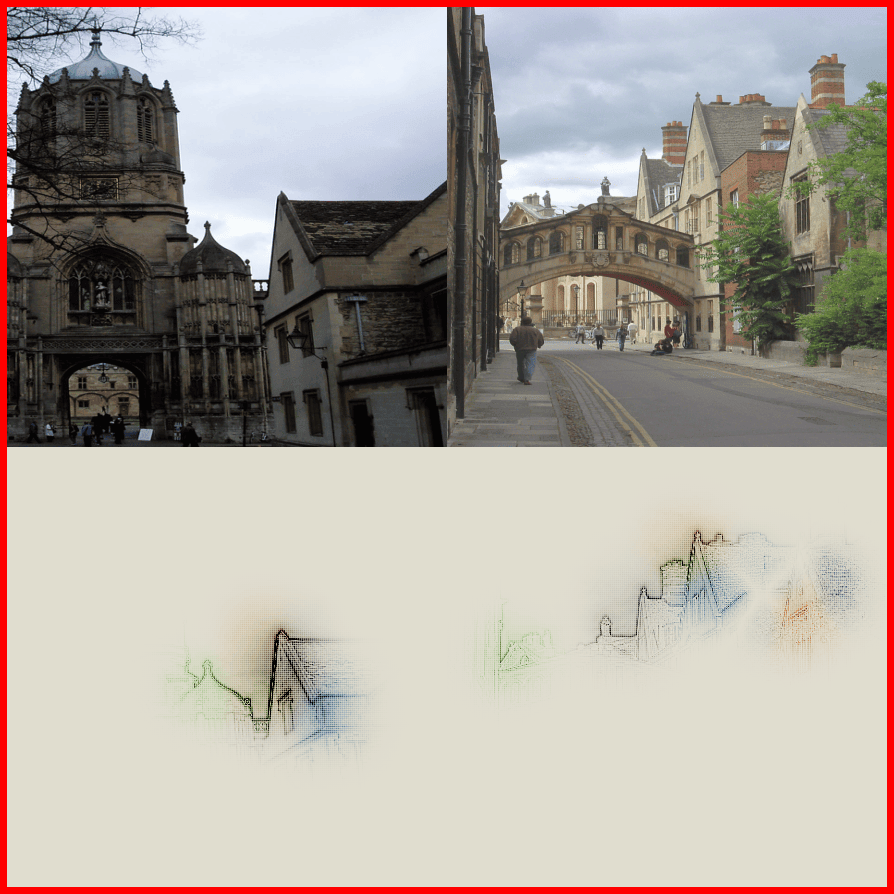}}
\caption{\textbf{PAIR-X provides interpretable, fine-grained, and highly-localized explainations which enable both correct and incorrect matches to be quickly identified.} The top half of each explanation shows pairwise-matched high-contribution deep features, and the bottom half shows a color-coded backpropagation to the original image pixels, highlighting plausible or implausible orientation shifts between fine-grained features.}

\label{fig:teaser}
\end{figure}

One application where explainability for deep metric learning models on fine-grained images is \textit{necessary} for trustworthy deployment is animal re-identification (re-ID)---the task of distinguishing individual members of a species. Animal re-ID is a crucial tool in ecology and conservation, serving as the foundation for key applications such as monitoring population trends and analyzing both individual and collective behaviors.~\cite{Tan2022-tj}. It is an active area of machine learning research, and recent years have seen steady growth in the performance of deep vision models on this task~\cite{schneider_2018_reid}. Animal re-ID typically relies on subtle, highly-localized fine-grained features such as subtle variations in spot or stripe patterns, or contours of fins, flukes, or ears. 
In contrast, explainability techniques like Grad-CAM frequently highlight broad image regions relevant to identification across all individuals (\eg highlighting the entire side of a giraffe, as in Figure \ref{fig:methods_ablation}), but fail to capture localized details that vary between individuals (\eg the placement of specific patterns on the giraffe).

The demand for better explainability techniques for animal re-ID is driven by an ongoing shift from classical re-ID techniques, such as HotSpotter or CurvRank~\cite{hotspotter, curvrank}, towards deep models that improve predictive accuracy and scalability to large datasets~\cite{otarashvili2024multispeciesanimalreidusing, cermak2023wildlifedatasetsopensourcetoolkitanimal, schneider_2018_reid, Haurum_2020_zebra_fish, Andrew_2021}. Classical techniques for re-ID typically rely on algorithmic matching of localized features (\eg SIFT~\cite{sift}), and are inherently explainable, as the local feature matches used for identification can be explicitly visualized. The resulting explanations can facilitate efficient manual review of model predictions, which is particularly necessary for high-profile applications such as population estimation for endangered species using visual mark-recapture, where incorrect labels can lead to significant errors in population size estimates~\cite{doi:10.1139/f01-131}). In interviews with giraffe re-ID experts, we found that the shift from classical techniques to deep models had resulted in a significant loss of explainability, which made it more challenging and thus slower for experts to verify predictions.

To address this gap, we propose \textbf{P}airwise m\textbf{A}tching of \textbf{I}ntermediate \textbf{R}epresentations for e\textbf{X}plainability (\textbf{PAIR-X}), a post-processing method that produces fine-grained visual explanations for deep models that mimic and extend those of classical techniques, without retraining or architectural changes (see Figure \ref{fig:teaser}). PAIR-X combines techniques from classic local feature matching with insights from modern explainability techniques for deep models~\cite{LRP, zhu2021visualexplanationdeepmetric, CRP, amir2022deepvitfeaturesdense}, and produces visualizations with the following key qualities:

\begin{itemize}[leftmargin=0.4cm, itemsep=0em]
\item \textbf{Local pairwise correspondences.} PAIR-X mimics the explainability visualizations produced by classical feature matching techniques (as shown in Figure \ref{fig:methods_ablation}). Correspondences between local image regions can be explicitly visualized.

\item \textbf{Fine-grained resolution.} PAIR-X produces explanations in the resolution of the original input image, thus capturing details in full resolution and shifting focus from broadly relevant regions to highly discriminative details.


\item \textbf{Quantifiable metrics.} Previous explainability approaches have often relied solely on manual, qualitative review to measure performance. For PAIR-X, we additionally propose a set of quantitative metrics (see Section \ref{sec:quantitative_metrics}) to compare its performance across models and datasets. Our metrics are designed to approximate how plausible a given visualization will appear to a user.
\end{itemize}

We evaluate PAIR-X across 34 public datasets for animal re-ID from WildlifeDatasets~\cite{cermak2023wildlifedatasetsopensourcetoolkitanimal} as well as the Oxford Building 5k dataset as a proof-of-concept outside of animal re-ID~\cite{Oxford5k}. Qualitatively, PAIR-X enables the visualization of interpretable local correspondences between image pairs, which are useful for both efficiently verifying correct matches and flagging high-scoring incorrect matches. 
We additionally propose a novel metric for our method which quantitatively demonstrates that PAIR-X produces measurably more plausible explanations for matching image pairs than for non-matching image pairs. PAIR-X can distinguish correct and incorect pairs \textbf{even in some cases where the deep metric model fails} (\eg high model match score for an incorrect pair).






\section{Related work}
\setlength{\columnsep}{0pt}  





\noindent 
\textbf{Animal re-ID.}
Historically, classical, inherently-explainable techniques based on homography-aligned local feature matching were used for individual re-ID of patterned~\cite{hotspotter,lahiri2011,kelly2001computer} and contoured species~\cite{curvrank,hughes2017automated}. Recently, deep models have been applied to a broad range of animal re-ID datasets~\cite{otarashvili2024multispeciesanimalreidusing, schneider_2018_reid, cermak2023wildlifedatasetsopensourcetoolkitanimal, Andrew_2021, Haurum_2020_zebra_fish, deb2018facerecognitionprimateswild, Dlamini_2020}. Deep metric learning methods~\cite{wang2014learning} enable recognition of animals not seen during training, and have been shown to both generalize across species~\cite{cermak2023wildlifedatasetsopensourcetoolkitanimal}, and scale more efficiently to large populations~\cite{otarashvili2024multispeciesanimalreidusing}. 
Concepts from local feature matching have been applied to deep vision model features for use in co-segmentation~\cite{co-segmentation} and semantic correspondence~\cite{semantic-matching}. Like local features, these deep features can be matched between image pairs~\cite{fischer2015descriptormatchingconvolutionalneural, amir2022deepvitfeaturesdense, Balntas_2017_CVPR}, but exploration into their use for explainability has been limited.


 \noindent 
\textbf{Explainability for deep vision models}
Methods for generating saliency heat maps (\eg, Grad-CAM, Grad-CAM++, etc.)~\cite{gradcam, gradcam++,  kpcacam,  hirescam} are used to highlight image regions that contribute to the prediction of a certain class or to the similarity of features between pairs of images, but struggle to capture fine-grained features~\cite{CRP}.
Explainability methods which repeatedly perturb an image and measure the change in model output~\cite{lime, shap} can be both computationally expensive and qualitatively ineffective for fine-grained details (see Fig.~\ref{fig:methods_ablation}).
Layer-wise relevance propagation (LRP) backpropagates relevance back to the original image pixels and can capture fine-grained details~\cite{LRP}. However, as seen in Figure \ref{fig:methods_ablation}, LRP fails to demonstrate how or why these details contribute to the model prediction.
Concept relevance propagation (CRP)~\cite{CRP} combines the fine-grained, localized explanations of LRP with encoded, interpretable “concepts”. 
While CRP can offer precise insights into a model's internal workings, it requires substantial human review to identify relevant and understandable concepts.
To directly compare pairwise feature correspondences, \citet{zhu2021visualexplanationdeepmetric} compute point-to-point activation intensity between pairs of images, and \citet{nguyen2023visualcorrespondencebasedexplanationsimprove} use optimal transport to match image regions. Both approaches rely on final-layer features, which are often coarse, and do not visualize receptive fields.
\section{Our method}
\begin{figure}
\centering
    \includegraphics[width=\columnwidth]{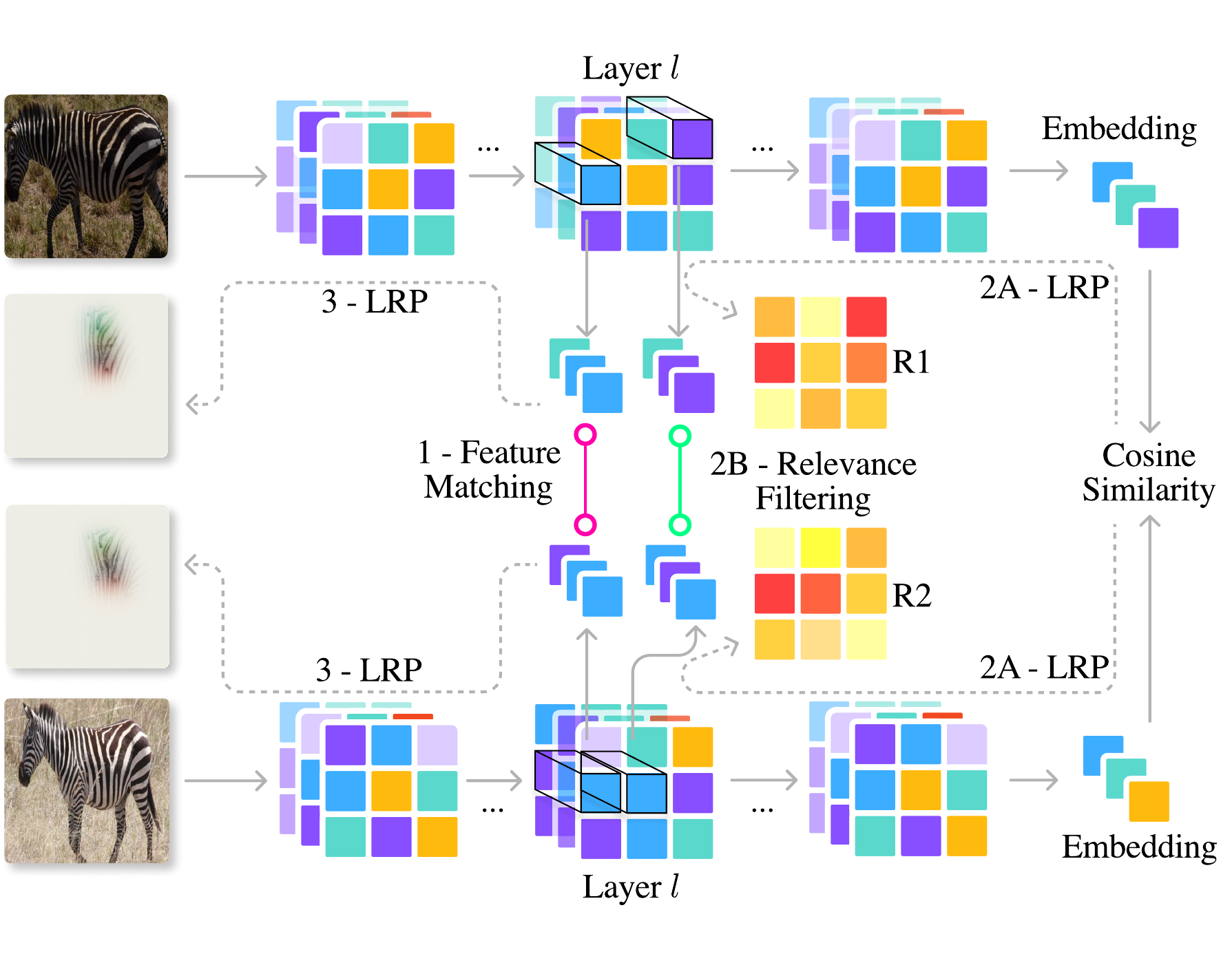}
    \caption{Overview of PAIR-X. In step 1, we match deep features derived from layer $l$. In step 2A, we perform LRP to obtain the relevance of each feature to the final cosine similarity. In step 2B, we filter the matched features acccording to their estimated relevance. Finally, in step 3, we use LRP to visualize which original image pixels are relevant to a filtered subset of matches.}
\label{fig:main_method}
\end{figure}

PAIR-X assumes access to a pretrained deep metric learning model (\ie~\cite{kaya2019deep}) for the fine-grained task of interest, and, taking inspiration from classical feature-matching techniques, constructs an interpretable, highly localized post-hoc explanation of similarity between image pairs by combining intermediate deep feature matching~\cite{fischer2015descriptormatchingconvolutionalneural} and layerwise relevance propagation~\cite{ amir2022deepvitfeaturesdense}. A visual description of our method can be found in Figure \ref{fig:main_method}, and additional methodological details are captured in Suppl. Sec.~\ref{supsec:method}.



\subsection{Deep feature matching}
\label{sec:deep_feature_matching}

Local feature matching techniques typically operate on sets of keypoints $\mathcal{K}$ and descriptors $\mathcal{D}$, where keypoints describe spatial locations within an image, and descriptors provide information about features present at the keypoints. PAIR-X uses a simple spatial decomposition to produce keypoint-descriptor sets $\mathcal{K}$ and $\mathcal{D}$ using the intermediate activations of a model~\cite{fischer2015descriptormatchingconvolutionalneural, amir2022deepvitfeaturesdense}. Given an intermediate activation matrix $A^l$ at selected layer $l$ of shape $w\times h\times c$ (where $w \times h$ represents the downsampled spatial dimensions), we decompose $A^l$ into $N=w\times h$ descriptors of length $c$, thus $\mathcal{D} = \{ A^l_{i,j} \mid i \in \{1,...,w\}, j \in \{1,...,h\} \}$.
The keypoints $\mathcal{K}$ for each descriptor are simply defined according to the estimated location in the image, \ie, $\mathcal{K} = \{ (i,j) \mid i  \in \{1,...,w\}, j \in \{1,...,h\} \}$.
As this definition of keypoint locations does not fully capture the true receptive fields for each neuron, we additionally utilize LRP to create more precise visualizations (see Sec.~\ref{sec:LRP}).
Given the complete sets of keypoints and descriptors, we perform brute-force matching with cross-checking, \ie, descriptors $(x,y)$ will only be returned for a match if $x$ is the closest match to $y$, and vice versa. This yields a set of matches $M$.

\subsection{Layerwise relevance propagation}
\label{sec:LRP}

Naive feature matching results in a large set of matches, some of which are unimportant to the model prediction.
To filter this large set of candidate matches, we first use LRP to determine relevance values for each neuron in the selected intermediate layer $l$. The resulting matrix takes shape $w\times h\times c$, and represents the estimated relevances of the values in the intermediate feature map at $l$. Given a keypoint match between $(i_1,j_1)$ and $(i_2,j_2)$ in the first and second images, respectively, we compute a relevance score for the match:

{\small\begin{equation}
    Rel((i_1,j_1), (i_2,j_2)) = \left(\sum_{k} R^{1,l}_{i_1,j_1,k} \right) \times \left(\sum_{k} R^{2,l}_{i_2,j_2,k}\right)
    \label{eq:intermediate_rel}
\end{equation}}
where $R^{1,l}$ and $R^{2,l}$ are the intermediate relevance matrices for the two images, and we keep the $n$ highest-scoring matches ($n = 20$ in our figures).

Neurons typically draw information from a wider surrounding region, or receptive field, which is not captured by a visualization with lines connecting approximated keypoints. To precisely visualize the pixel-space contributions of matched features, we use LRP to backpropagate from the selected intermediate layer $l$ to the original image for a set of top matches (ranked according to the relevance metric presented in Equation \ref{eq:intermediate_rel}). Given a feature match between keypoints $(i_1,j_1)$ and $(i_2,j_2)$, we backpropagate for each image in the pair separately. We first mask the intermediate activation matrix, $A^l$, to the values included in the matched keypoint descriptor. This takes the form:
\begin{equation}
    \mathcal{M}(A^l_{i,j}) =
    \begin{cases} 
    A^l_{i,j}, & \text{if } (i,j) = (i_1,j_1) \\
    0, & \text{otherwise}
    \end{cases}
\end{equation}
We then backpropagate from $\mathcal{M}(A^l_{i,j})$. The result of this backpropagation takes the same shape as the original input image, which can be summed channel-wise for RGB images to produce a 2D heatmap. We visualize the pixel-wise relevances by assigning a different color map to each match, then combining all matches into a single color-coded visualization.

\subsection{Quantitative explainability metrics}
\label{sec:quantitative_metrics}

Because explainability is inherently qualitative, it is difficult to define metrics that can quantify explainability performance. We propose two quantitative metrics which seek to capture the ``plausibility'' of PAIR-X explainations. 


\noindent \textbf{Inverted residual mean.} Our first metric, the inverted residual mean, aims to capture whether the feature matches follow a ``ground truth'' homography $\mathcal{H}$, as a proxy to understand whether the matches correctly align the image subject. Matches that do not follow a homography will typically appear less plausible. $\mathcal{H}$ is calculated using classical techniques; we use a SuperPoint extractor to extract local features and a LightGlue matcher to find feature matches, then estimate $\mathcal{H}$ from the matches~\cite{superpoint_detone_2017, lindenberger2023lightgluelocalfeaturematching}. Each keypoint $p_1 = (i_1, j_1)$ is projected from the first image using this ``ground truth" homography as:

\begin{equation}
    \mathcal{H} \left( \begin{bmatrix} i_1 \\ j_1 \\ 1 \end{bmatrix}\right) =
    \begin{bmatrix} i_1' \\ j_1' \\ w' \end{bmatrix}
    \rightarrow
    p_1' = \frac{1}{w'}\begin{bmatrix} i_1' \\ j_1' \end{bmatrix}.
\end{equation}
For the final score $S_1$, we take the reciprocal of the average of the residuals on the second image (between the projected points and the matched points) across all feature matches $M$ (before LRP filtering):

\begin{equation}
    S_1 = \frac{|M|}{\sum_{(p_1,p_2) \in M} ||p_1'-p_2||}. 
\end{equation}

We take the reciprocal because residual means are typically concentrated around small values, with a long tail of high-value outliers. Taking the reciprocal allows for improved separability of the smaller values.

\noindent \textbf{Relevance-weighted match coverage.} The second metric we propose, relevance-weighted match coverage, aims to understand what proportion of relevant regions are successfully matched by PAIR-X. Visualizations that fail to show matches between the most important regions of an image will also appear less informative. Each feature that has been matched by PAIR-X is weighted by its relevance score, summed, and then divided by the sum of all relevance scores:
\begin{equation}
    \frac{\sum_{(i,j) \in \mathcal{K}_1}R^1_{i,j} + \sum_{(i,j) \in \mathcal{K}_2}R^2_{i,j}}{\sum_{i,j}R^1_{i,j} + \sum_{i,j}R^2_{i,j}},
    \label{eq:rel_weighted_match_coverage}
\end{equation}
where $\mathcal{K}_1$ and $\mathcal{K}_2$ denote the sets of matched keypoints of the two images, before being filtered by relevance.
For each dataset evaluated, we compute these metrics across top-ranked correct and incorrect pairs following the procedure described in Suppl. Sec.~\ref{sec:pair_selection_appendix}.

\captionsetup[subfloat]{labelformat=empty}
\begin{figure*}
\small
    \hrulefill\vspace{.5cm}\\
    \begin{minipage}{0.18\textwidth}
        \subsection*{Patterned species}
        \vspace{-5pt}
        PAIR-X performs best on fine-grained tasks with highly-localized and structured details, e.g. re-ID for patterned species such as giraffes, cows, zebras, and sea turtles. These patterns capture unique spatial arrangements for individuals, thus the visualizations for correct and incorrect matches are interpretably different. \textcolor{mydarkgreen}{Green outlines} indicate correct image matches (same individual), and \textcolor{red}{red outlines} indicate incorrect matches (different individuals).
    \end{minipage}
    \hspace{.02\textwidth}
    \begin{minipage}{0.79\textwidth}
    \subfloat[Giraffes - Correct and Incorrect Matches]{\ToggleImages[width=0.24\textwidth]{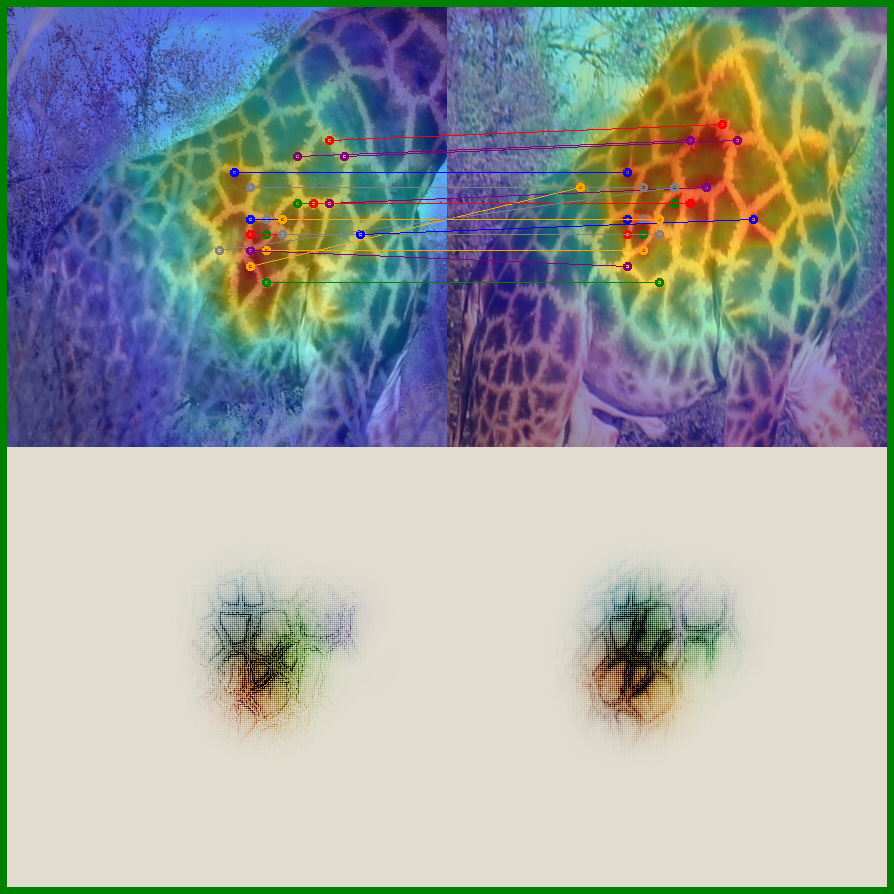}{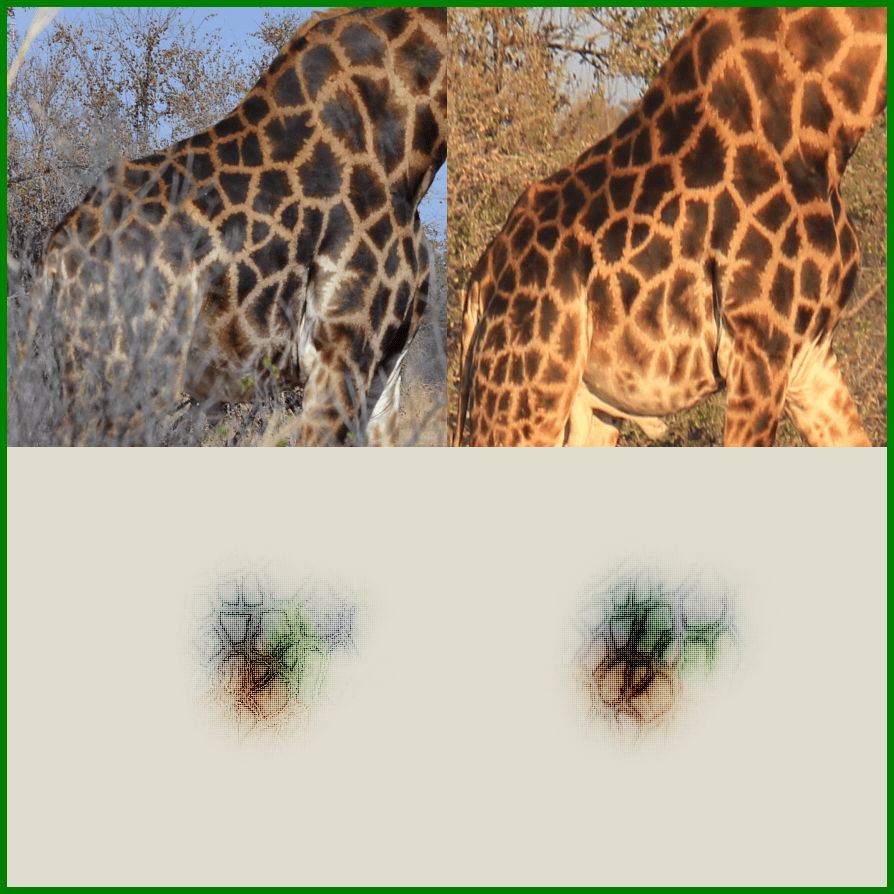}
    \ToggleImages[width=0.24\textwidth]{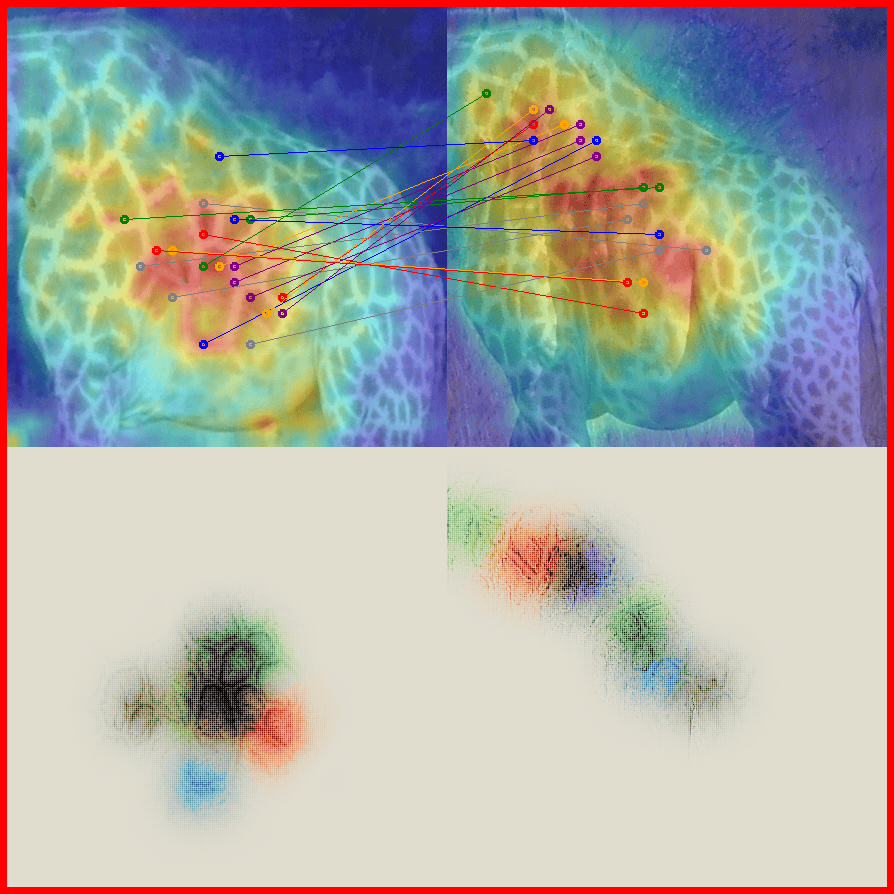}{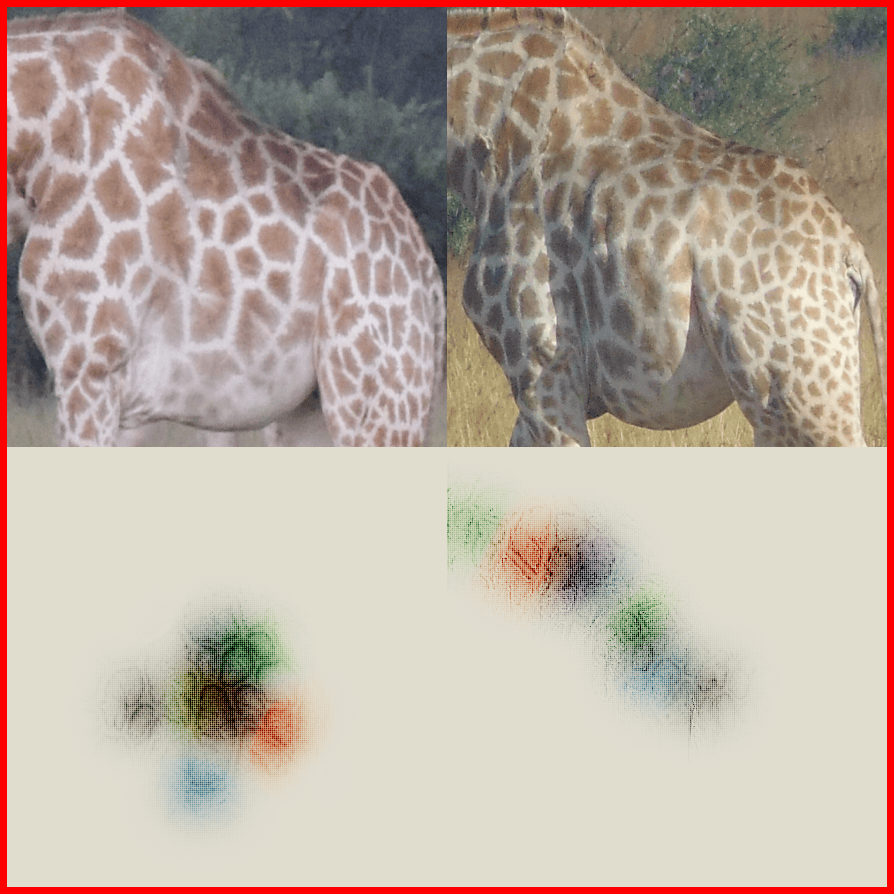}} \hspace{0.02\textwidth}
    \subfloat[Cows - Correct and Incorrect Matches]{\ToggleImages[width=0.24\textwidth]{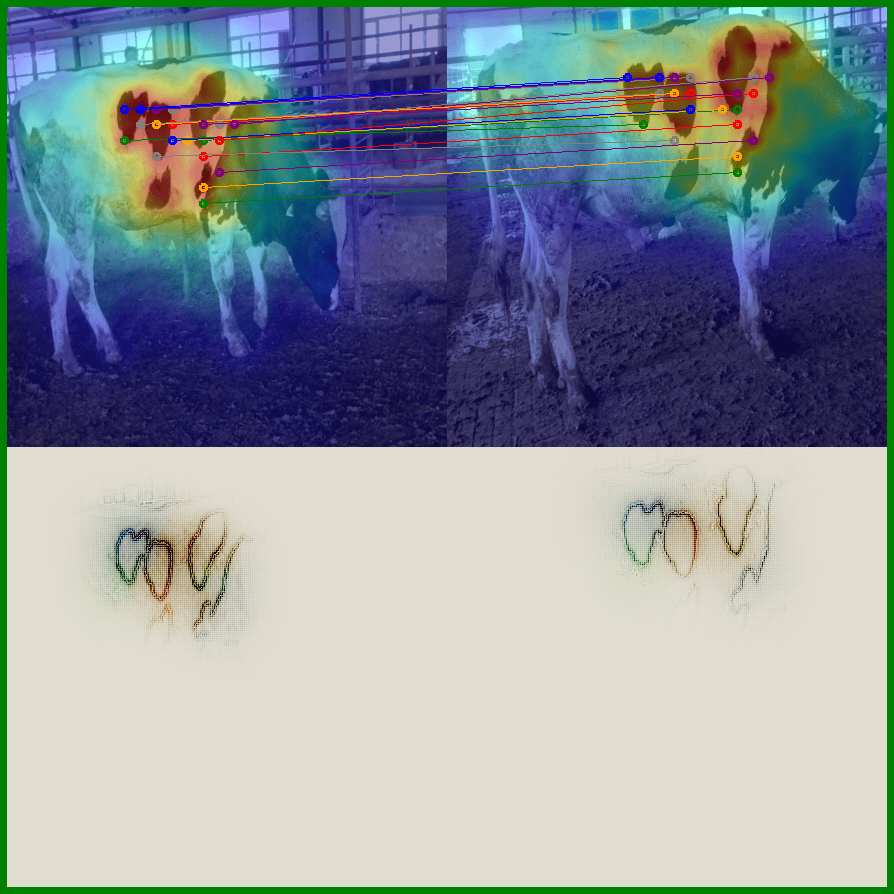}{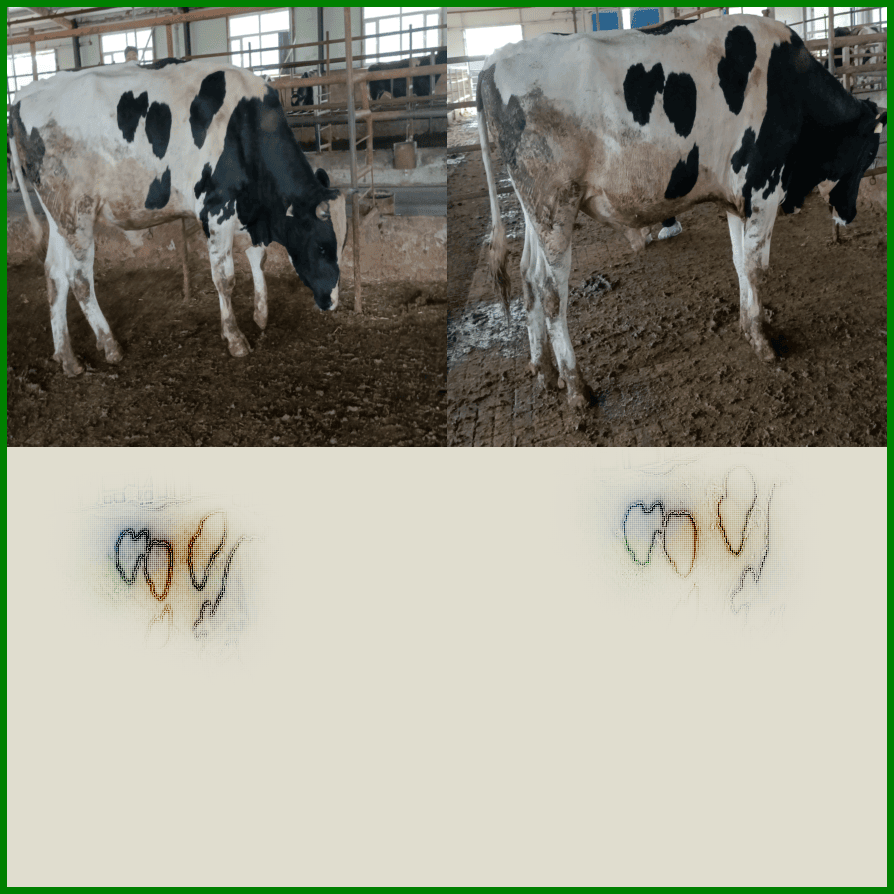}
    \ToggleImages[width=0.24\textwidth]{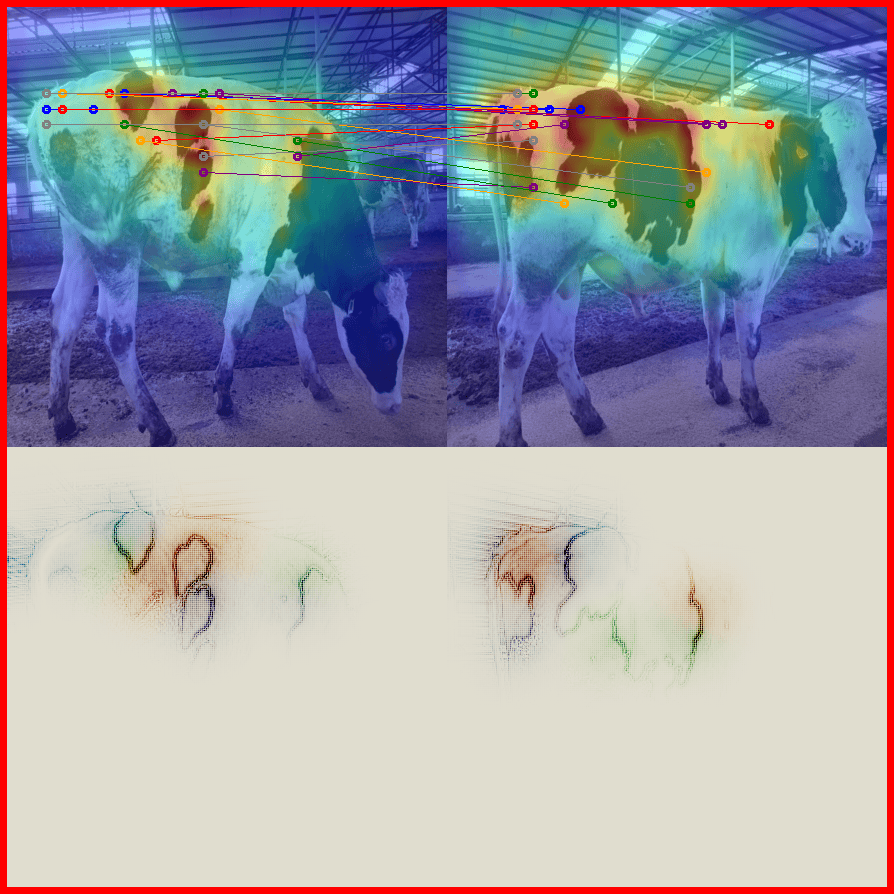}{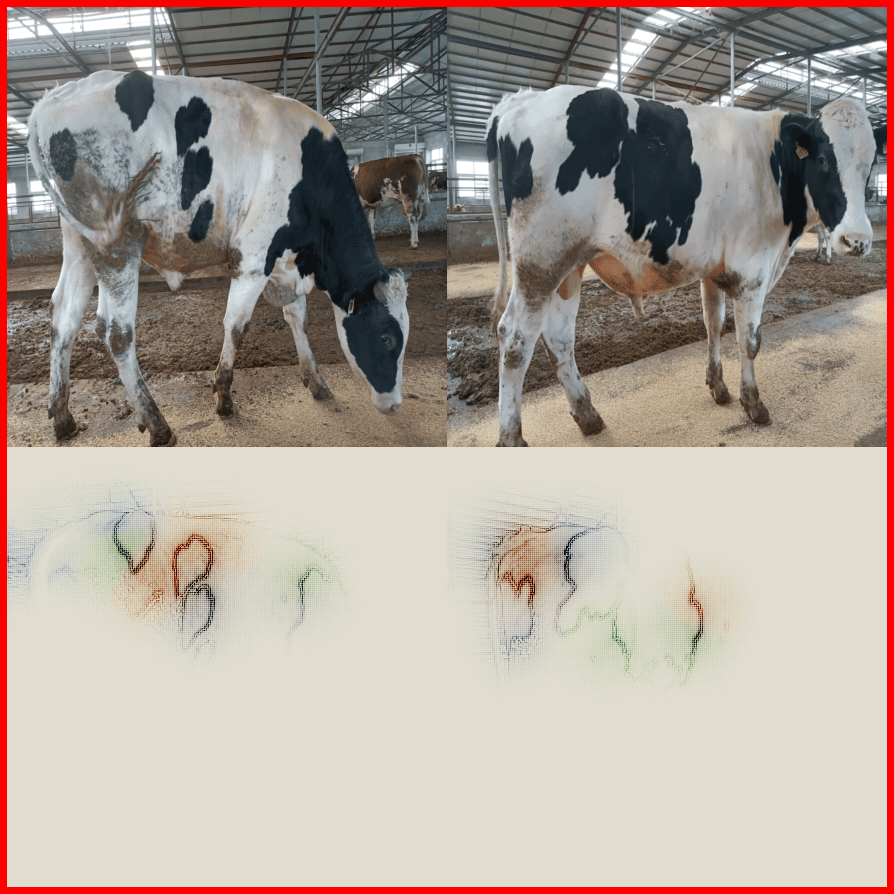}}

    \subfloat[Plains Zebras - Correct and Incorrect Matches]{\ToggleImages[width=0.24\textwidth]{figures/improved_color_maps/zebra_correct_teaser_dataset_ablation_pairx.png}{figures/improved_color_maps/zebra_correct_teaser_dataset_ablation_raw.png} \ToggleImages[width=0.24\textwidth]{figures/improved_color_maps/zebra_incorrect_pairx.png}{figures/improved_color_maps/zebra_incorrect_raw.png}} \hspace{0.02\textwidth}
    \subfloat[Sea Turtles - Correct and Incorrect Matches]{\ToggleImages[width=0.24\textwidth]{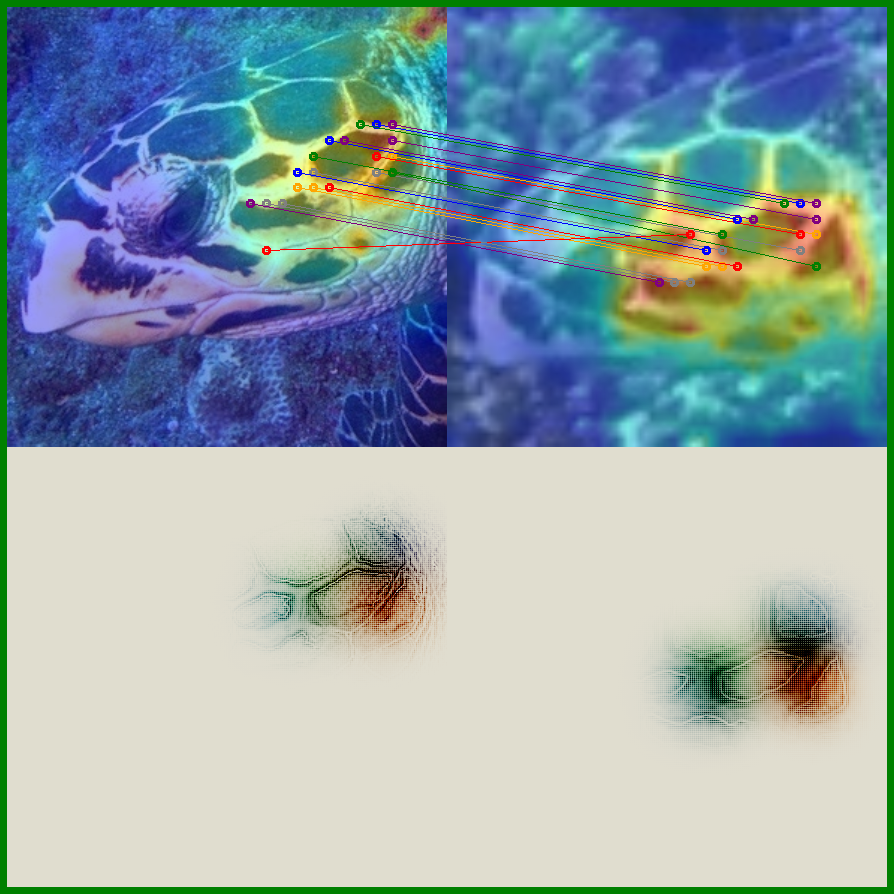}{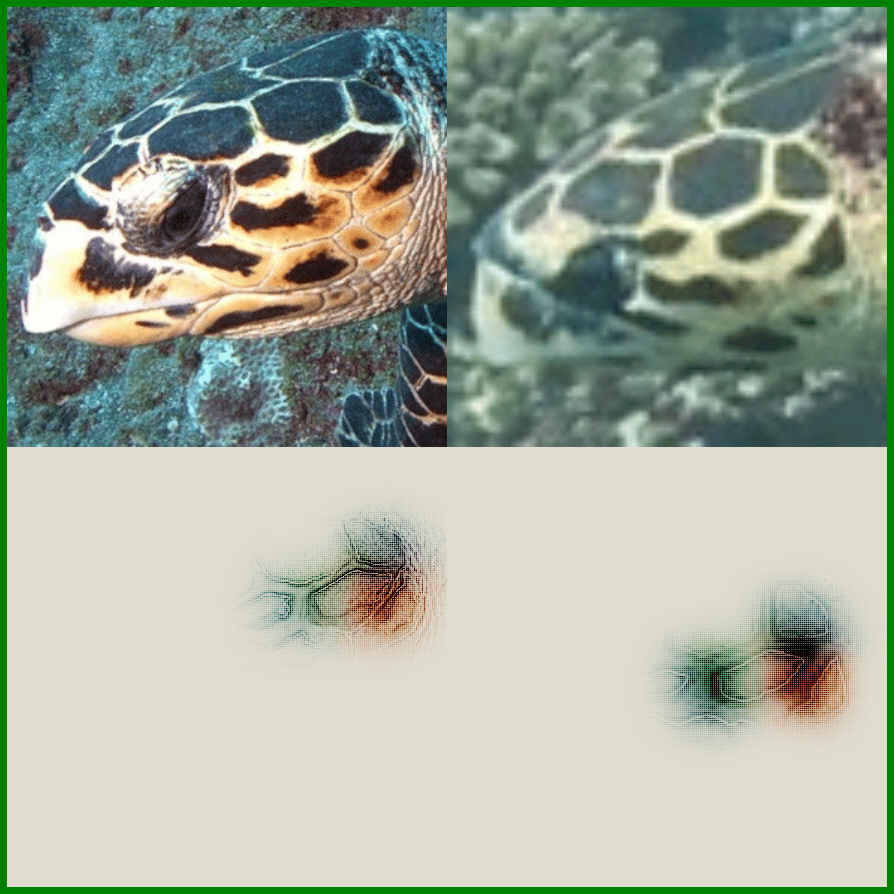} \ToggleImages[width=0.24\textwidth]{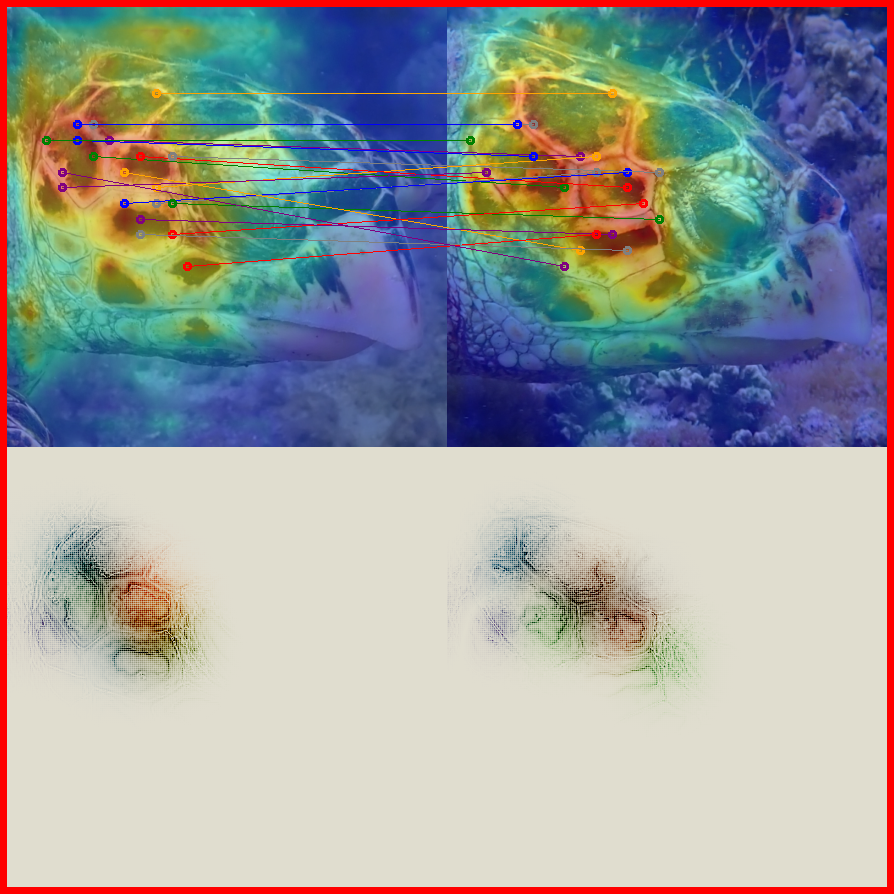}{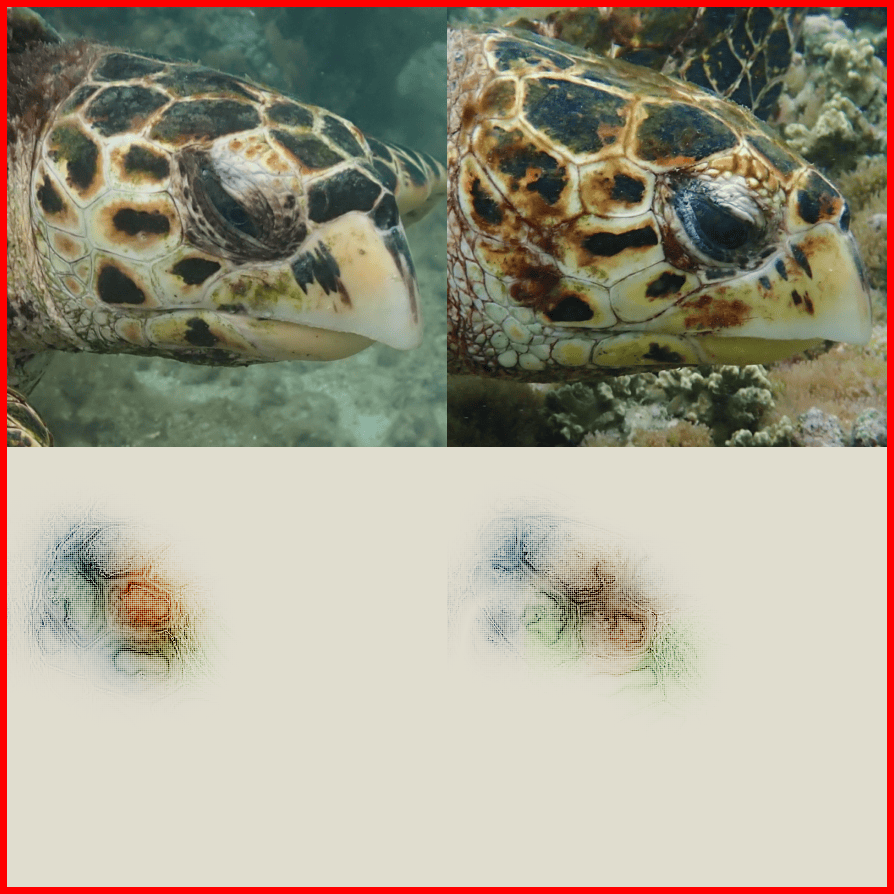}}
    \end{minipage}

    \vspace{.5cm}
    \hrulefill\vspace{.2cm}\\
    \begin{minipage}{0.19\textwidth}
        \subsection*{Unpatterned species}
        PAIR-X is designed to optimally explain fine-grained, localized features that follow unique spatial arrangements for different categories. \textbf{For species without structured biometric patterns such as stripes, we find PAIR-X explanations are more useful for individuals with unique markings.} Without these markings, PAIR-X can sometimes highlight features that are invariant between individuals, producing misleading explanations for incorrect matches.
    \end{minipage}
    \hspace{.02\textwidth}
    \begin{minipage}{0.385\textwidth}
        \subfloat[Cat and Sea Star Images with Distinctive Markings]{\ToggleImages[width=0.48\textwidth]{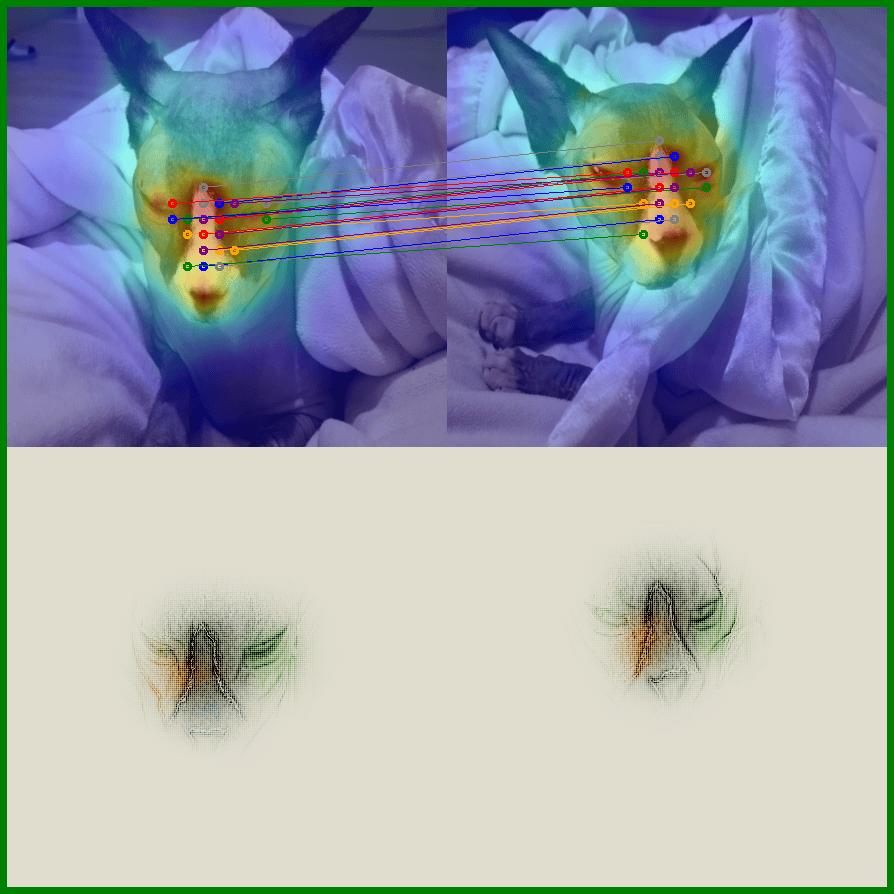}{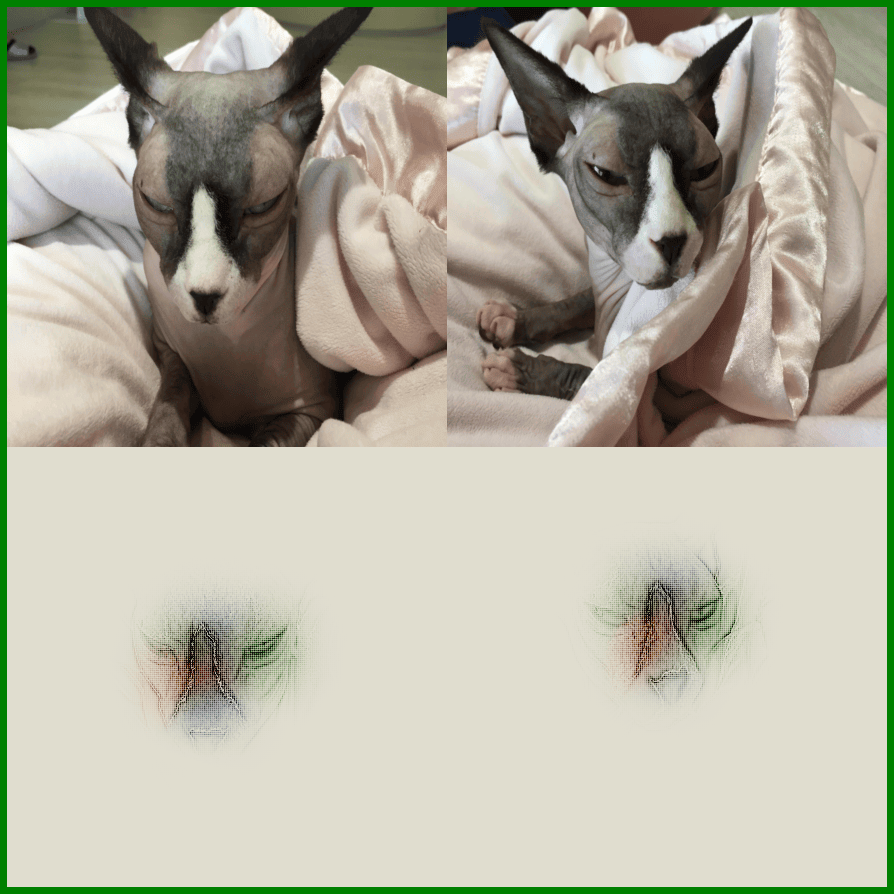} \ToggleImages[width=0.48\textwidth]{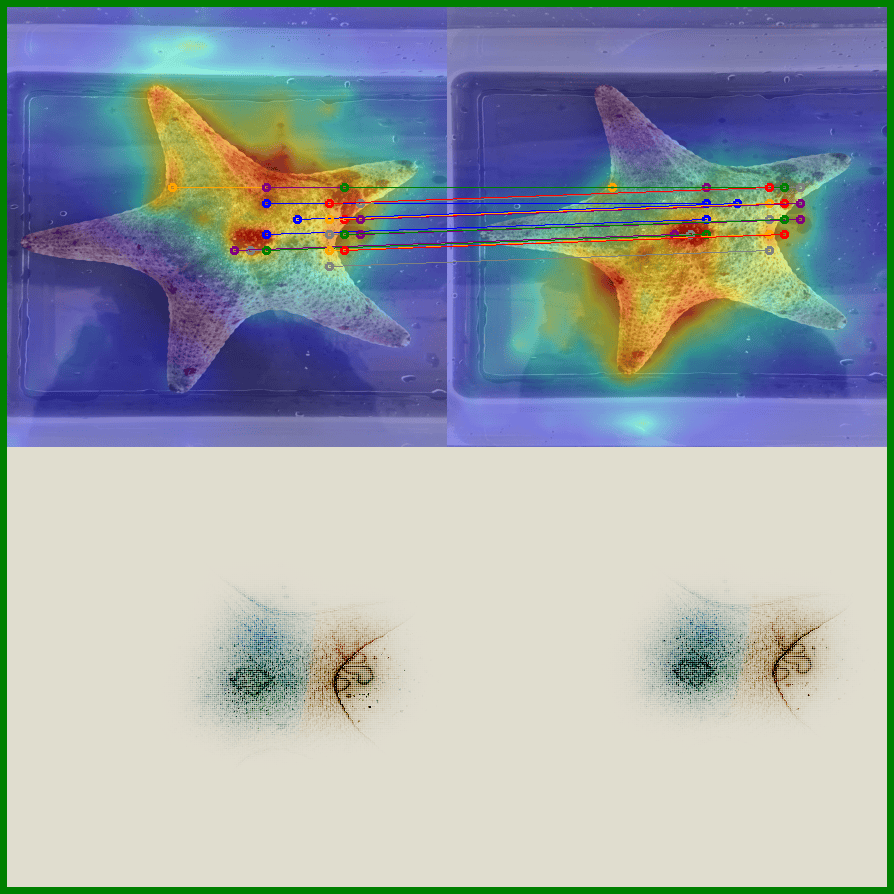}{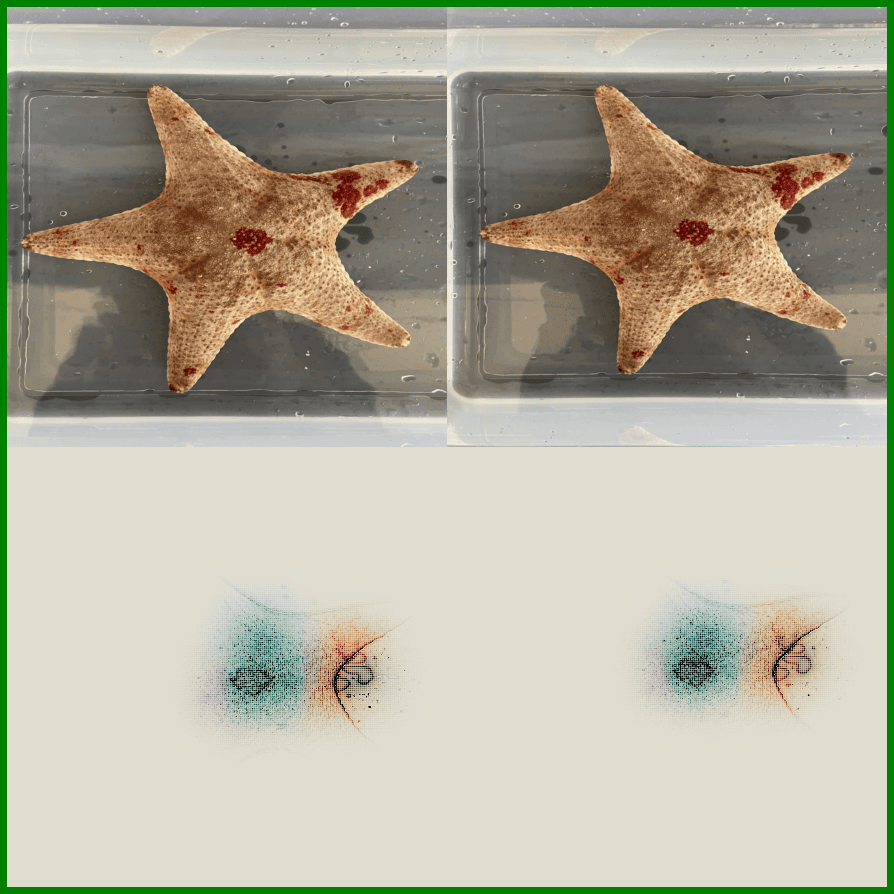}} 

        \subfloat[Cat and Sea Star Images without Distinctive Markings]{\ToggleImages[width=0.48\textwidth]{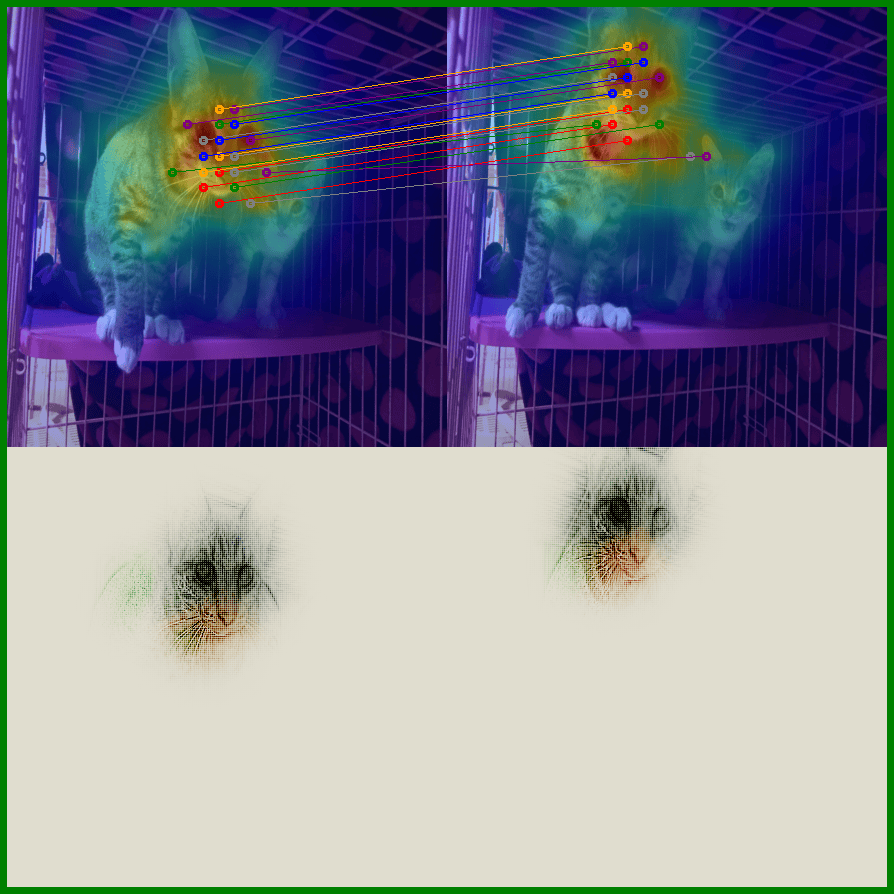}{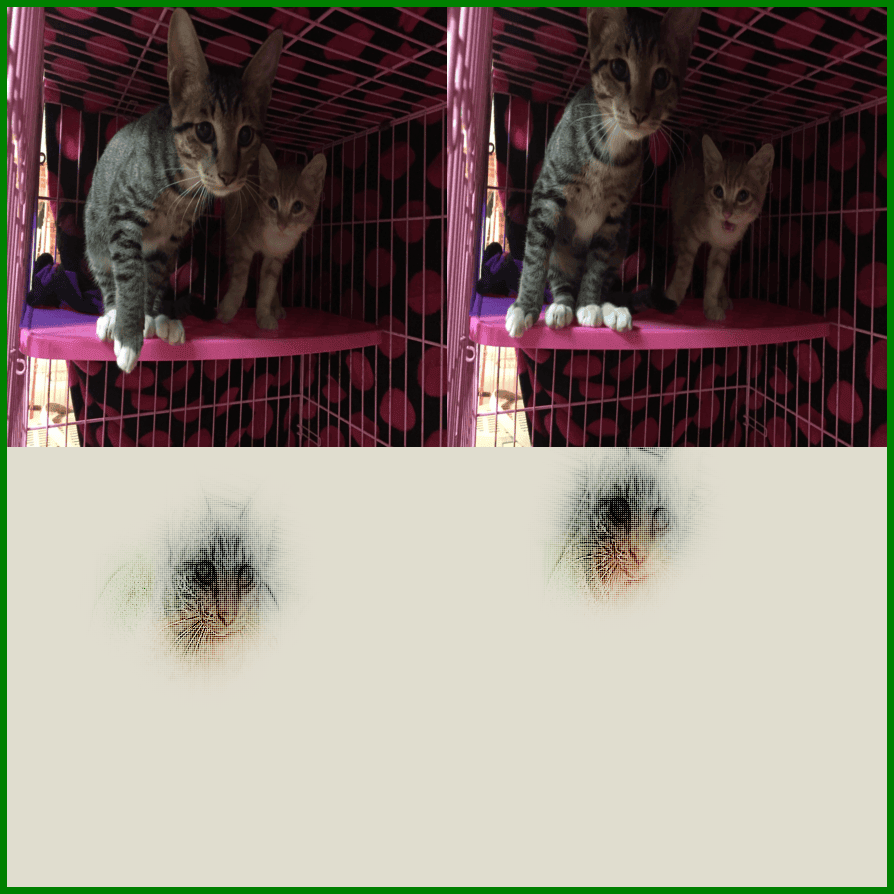}
        \ToggleImages[width=0.48\textwidth]{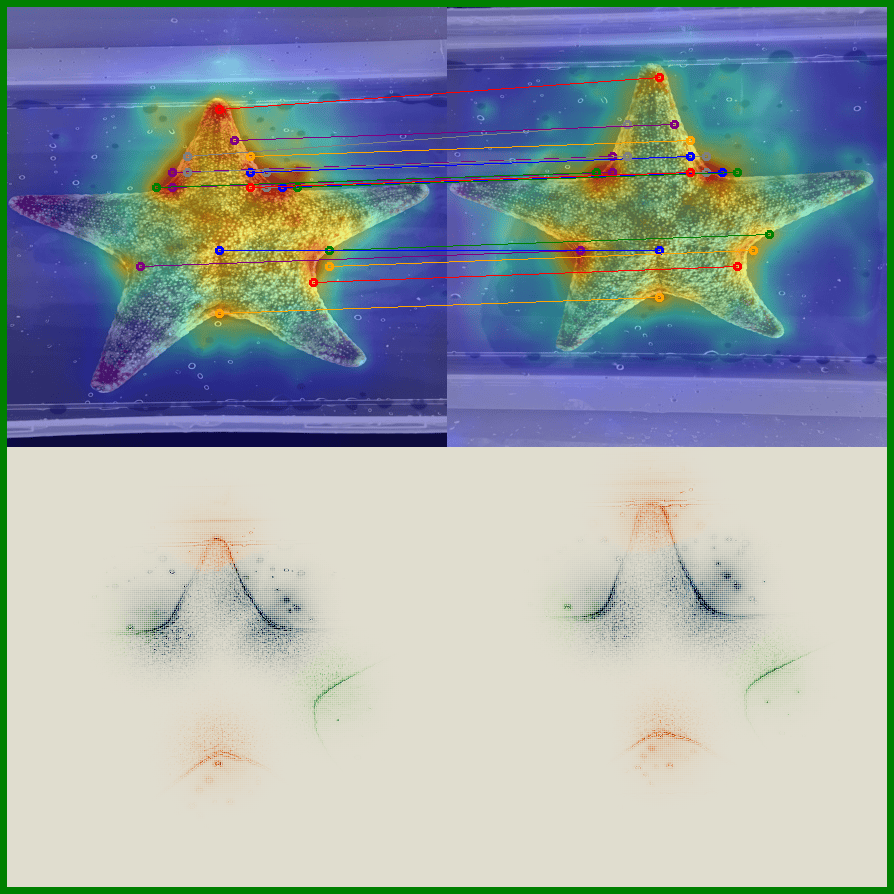}{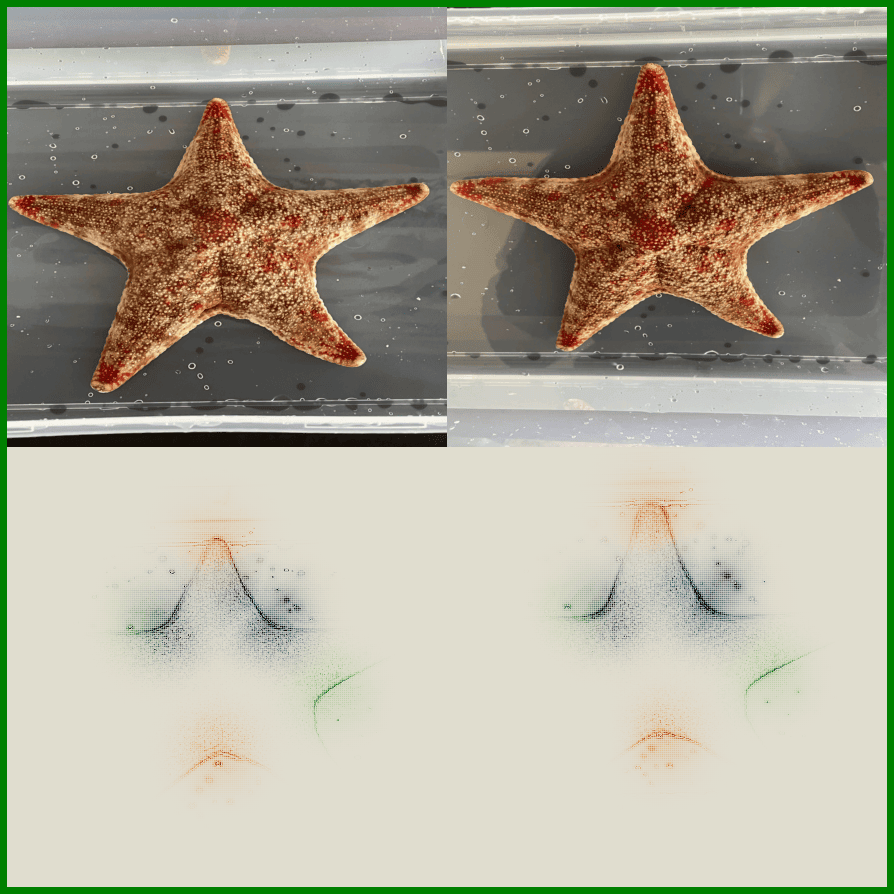}}
    \end{minipage}
    \vrule width .75pt height 120pt
    \hspace{.01\textwidth}
    \begin{minipage}{0.375\textwidth}        
        \subsection*{Identifying spurious correpondences with PAIR-X}
        \begin{minipage}{0.48\textwidth} 
            A key use case for explainability is to identify when model decisions are based on irrelevant information. In the  images to the right, PAIR-X helps visualize the model's focus on foreground information.
        \end{minipage}
        \begin{minipage}{0.48\textwidth} {\ToggleImages[width=\textwidth]{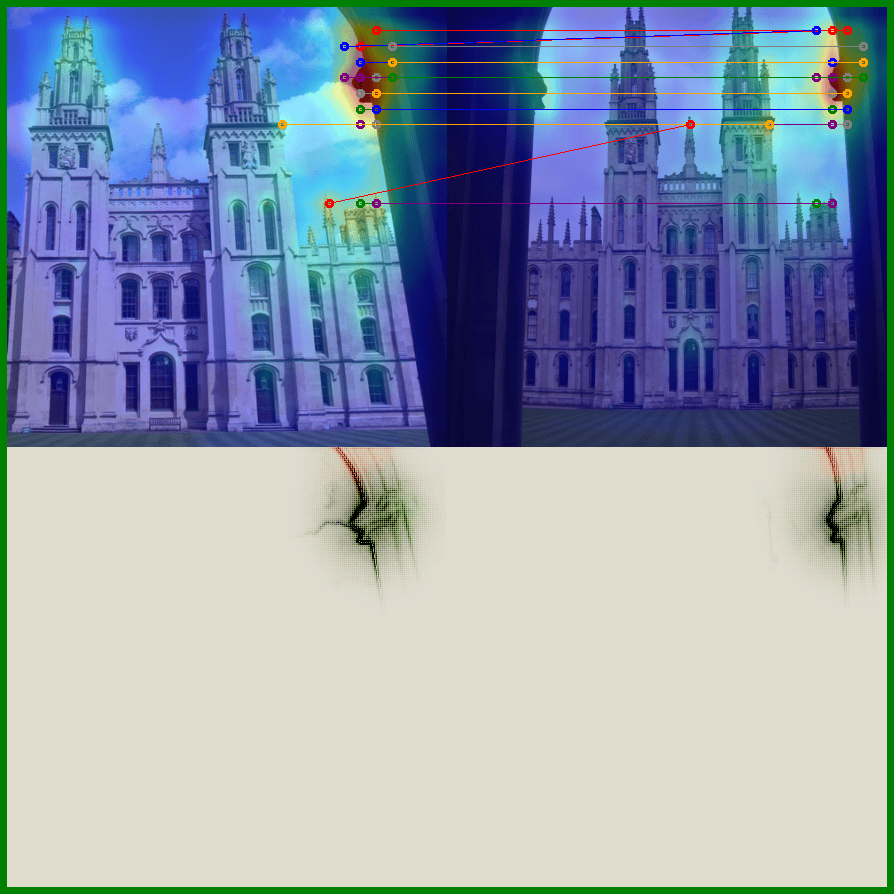}{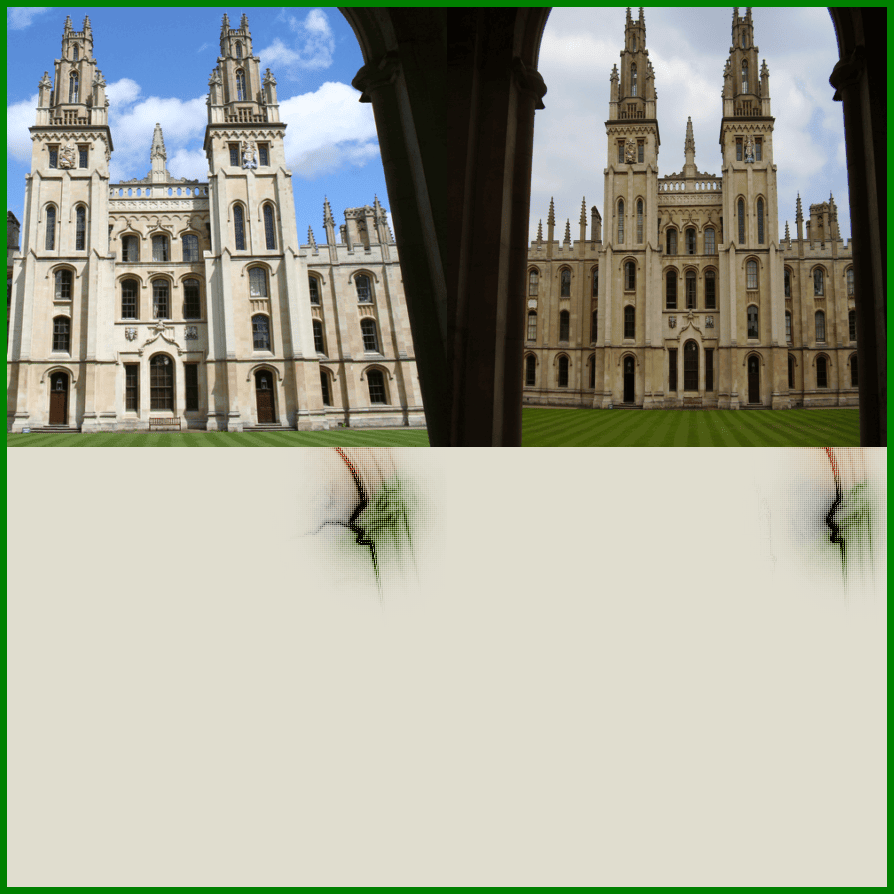}}
        \end{minipage}
        
        \subfloat[For these bird images, PAIR-X captures the background information contributing to the model prediction.]{\ToggleImages[width=0.48\textwidth]{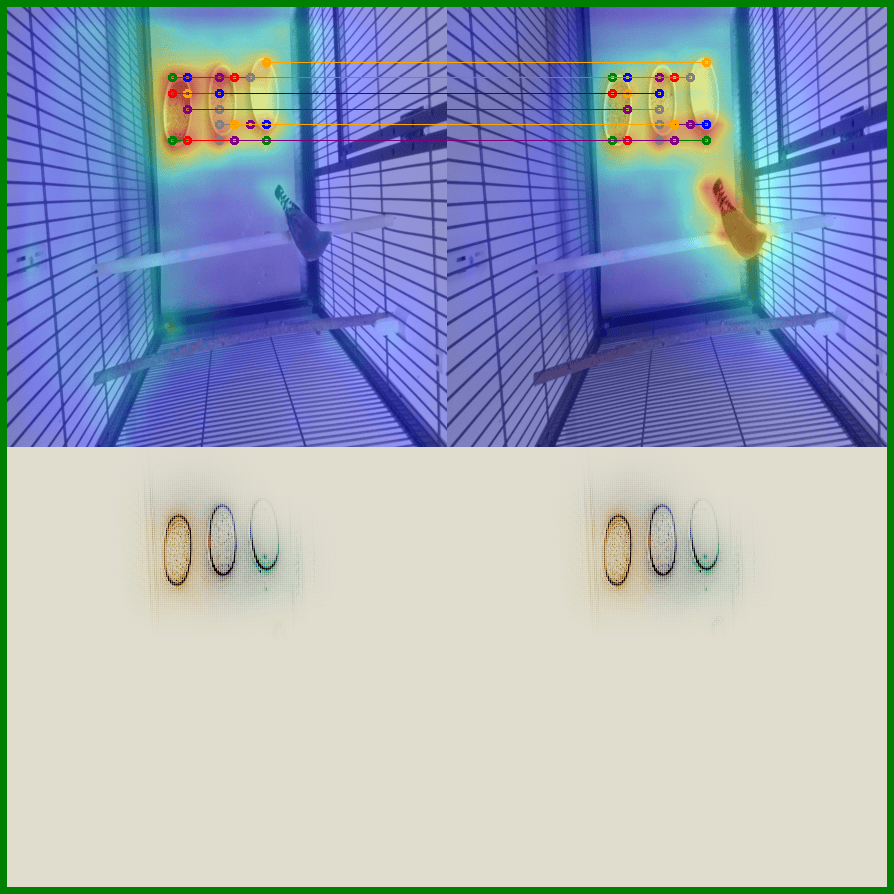}{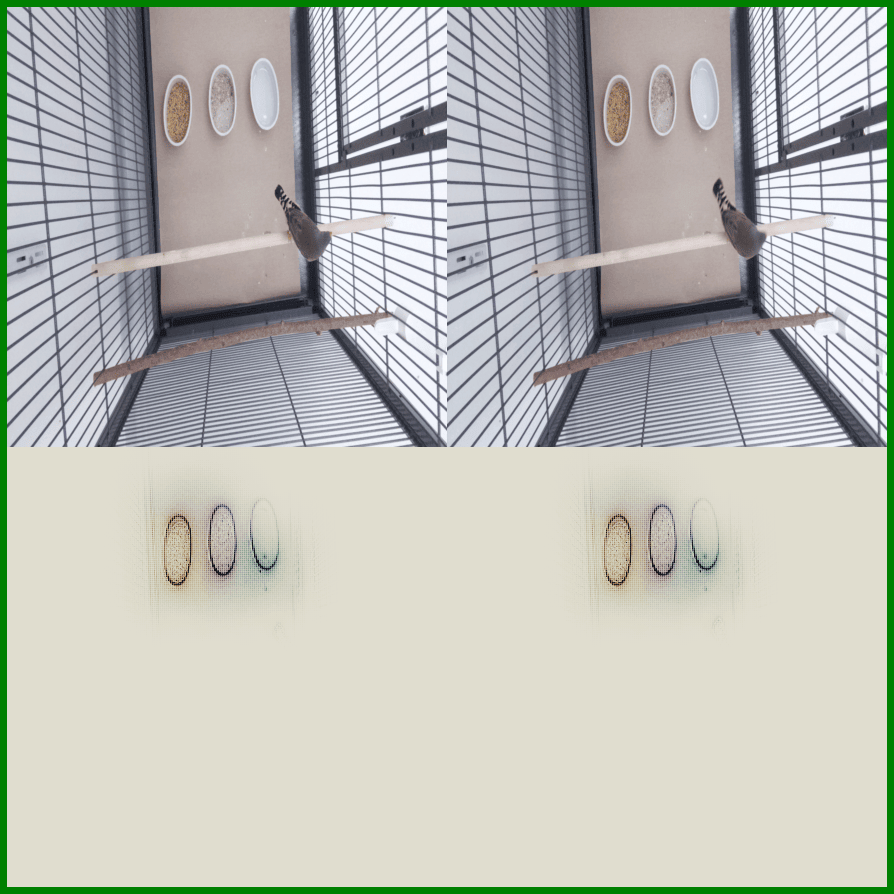} \ToggleImages[width=0.48\textwidth]{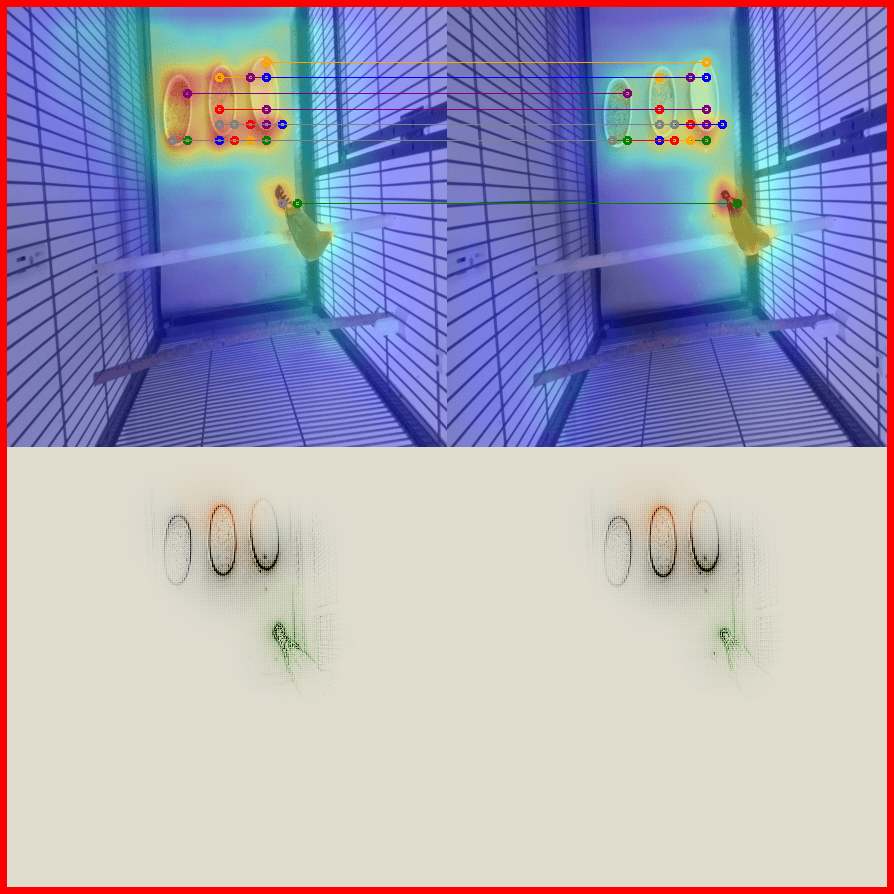}{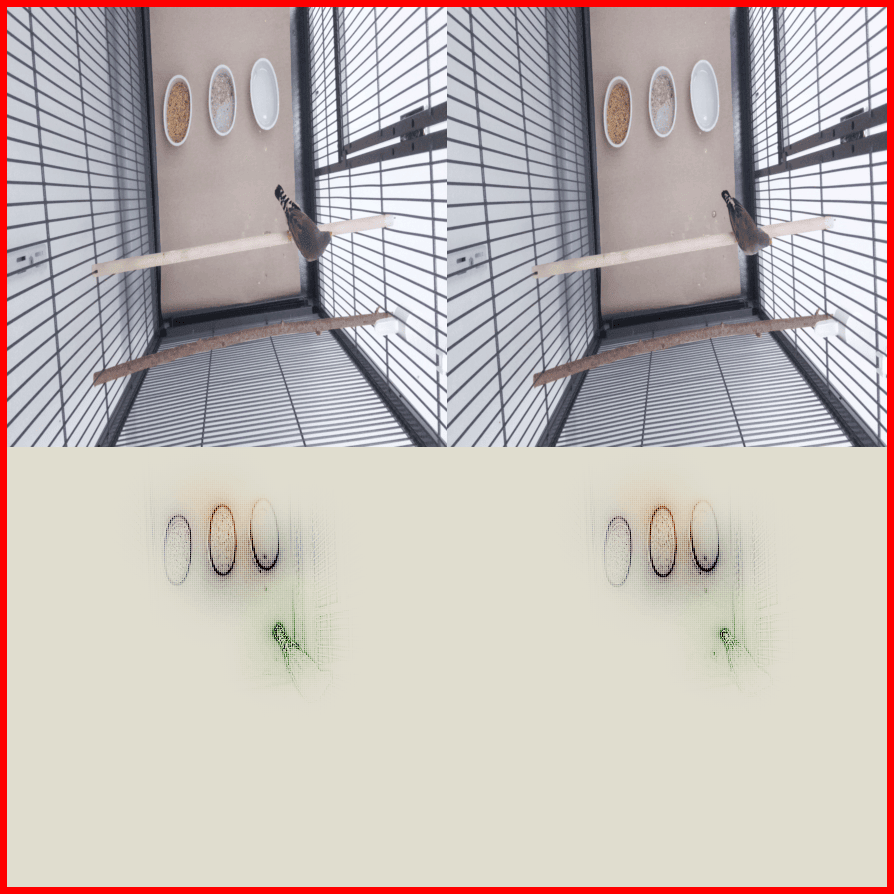}}

    \end{minipage}
    \vspace{.1cm}
    \hrulefill\vspace{.5cm}\\
    \begin{minipage}{0.18\textwidth}
        \subsection*{Failure mode: \\extreme pose\\ variation}
        As with many classical feature-matching techniques, PAIR-X performance degrades as pose variation becomes increasingly extreme.
    \end{minipage}
    \hspace{0.03\textwidth}
    \begin{minipage}{0.79\textwidth}
        \subfloat[Minor pose difference]{\ToggleImages[width=0.235\textwidth]{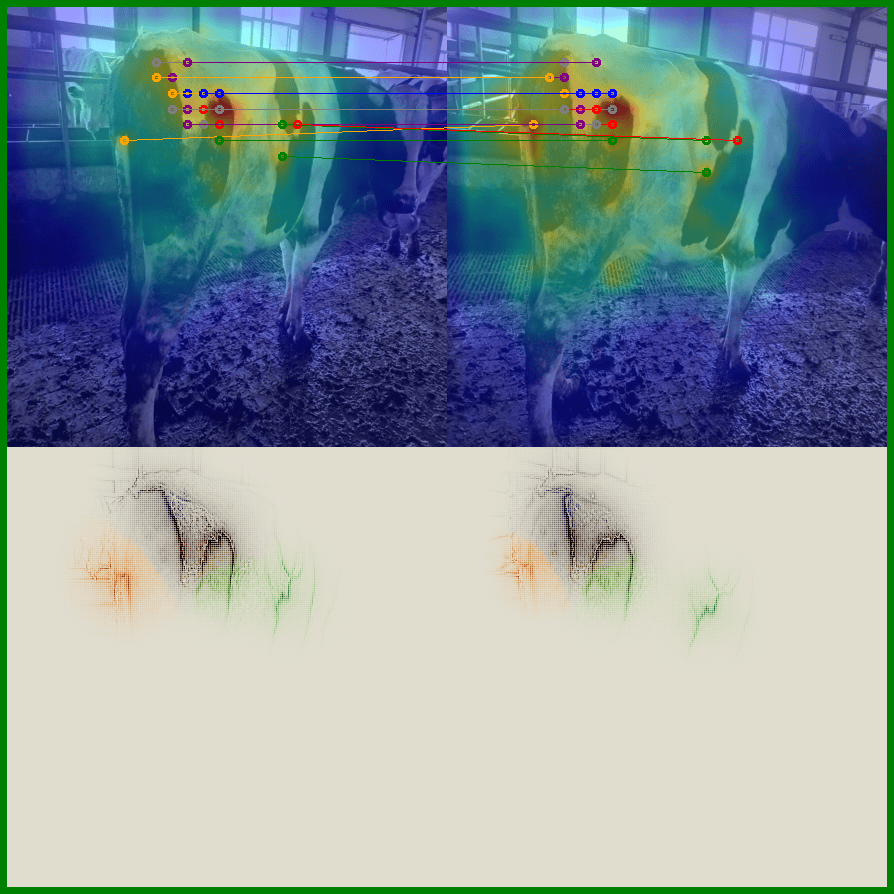}{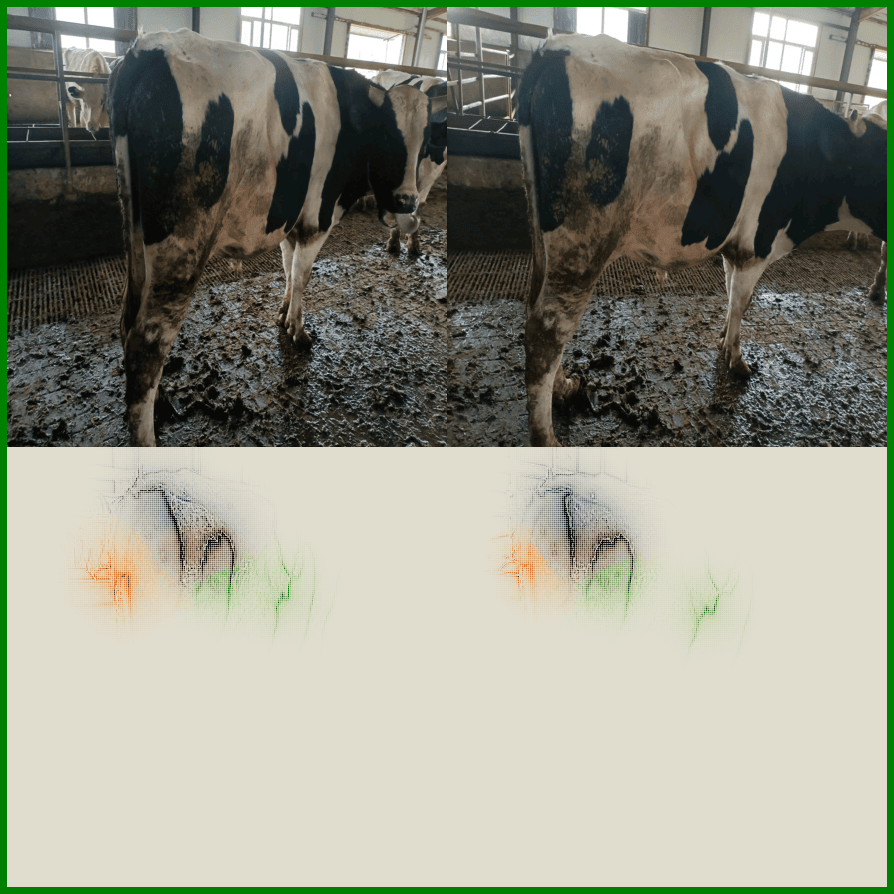}} \hspace{0.01\textwidth}
        \subfloat[Moderate pose difference]{
        \ToggleImages[width=0.235\textwidth]{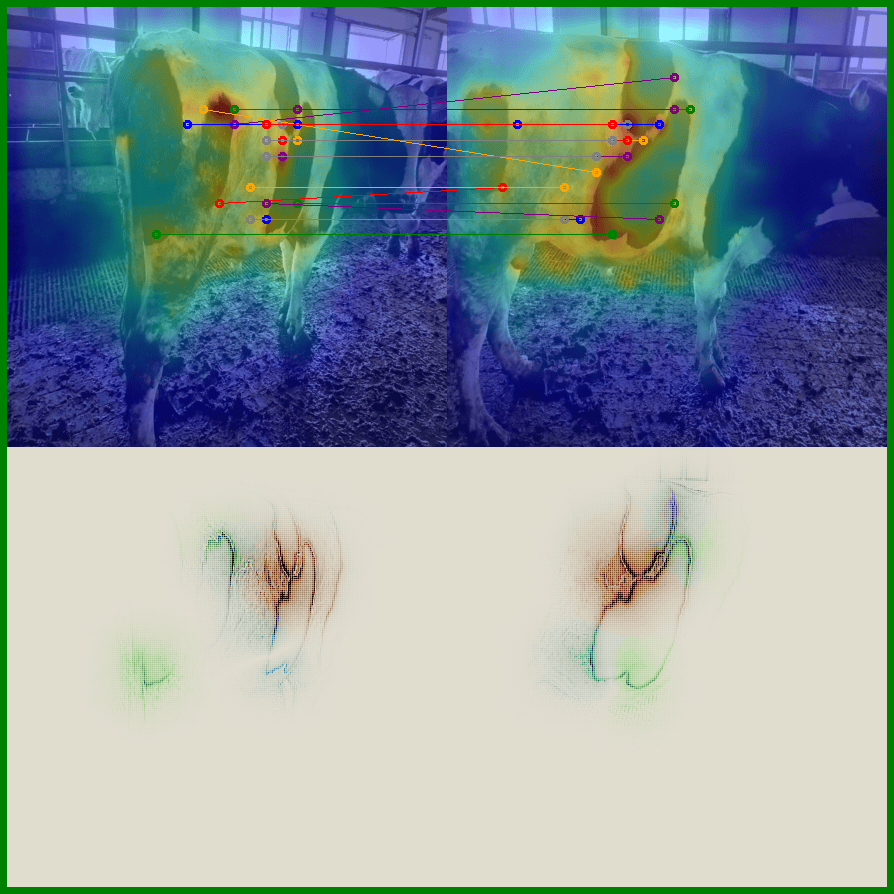}{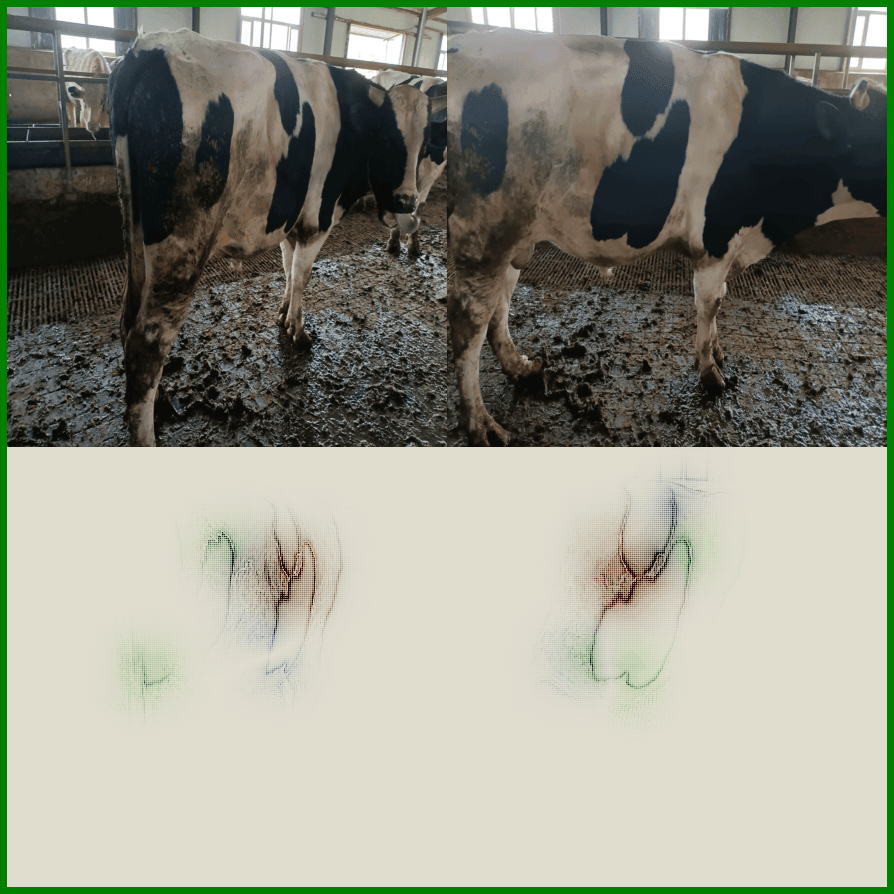}} \hspace{0.01\textwidth}
        \subfloat[Major pose difference]{\ToggleImages[width=0.235\textwidth]{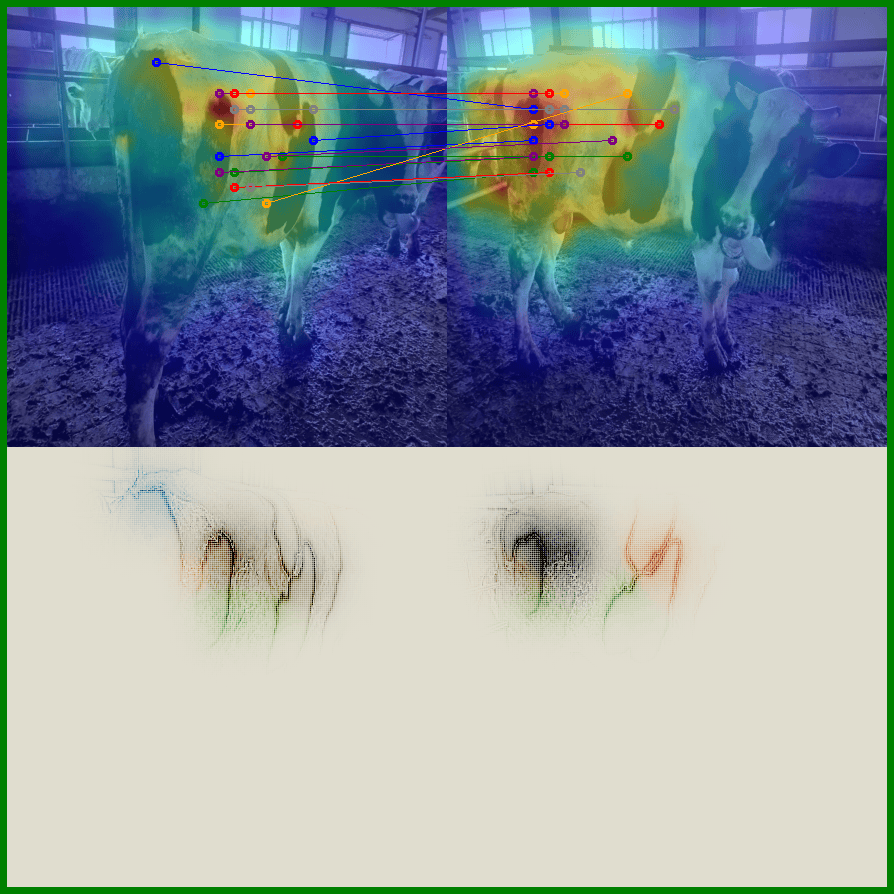}{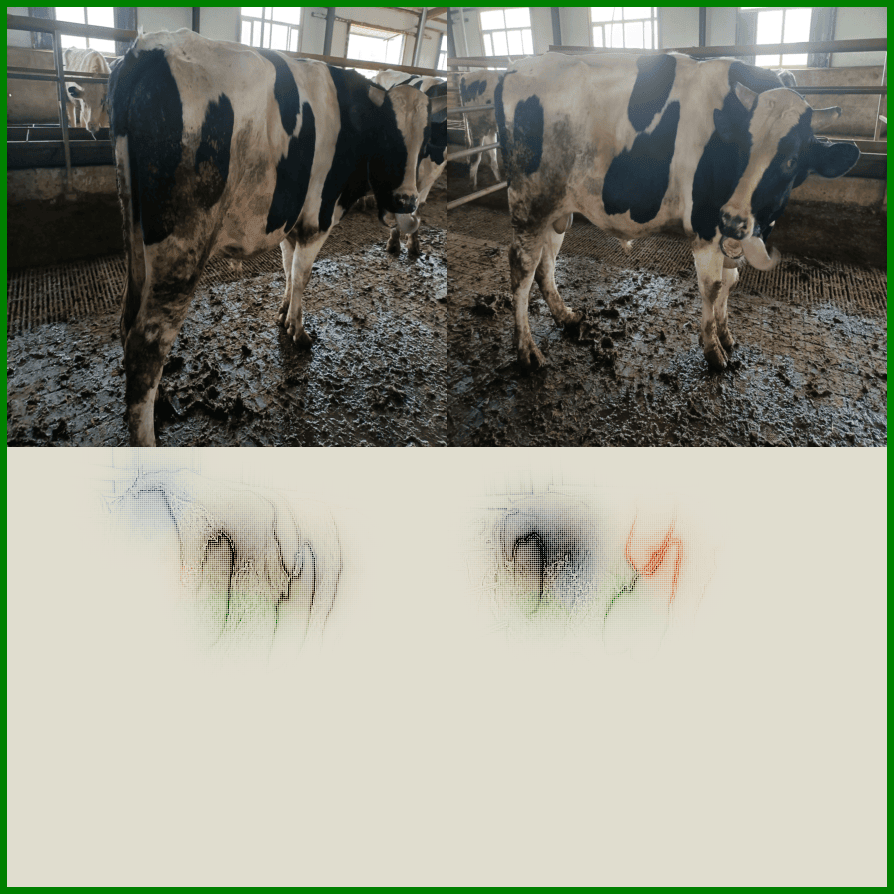}} \hspace{0.01\textwidth}
        \subfloat[Extreme pose difference]{\ToggleImages[width=0.235\textwidth]{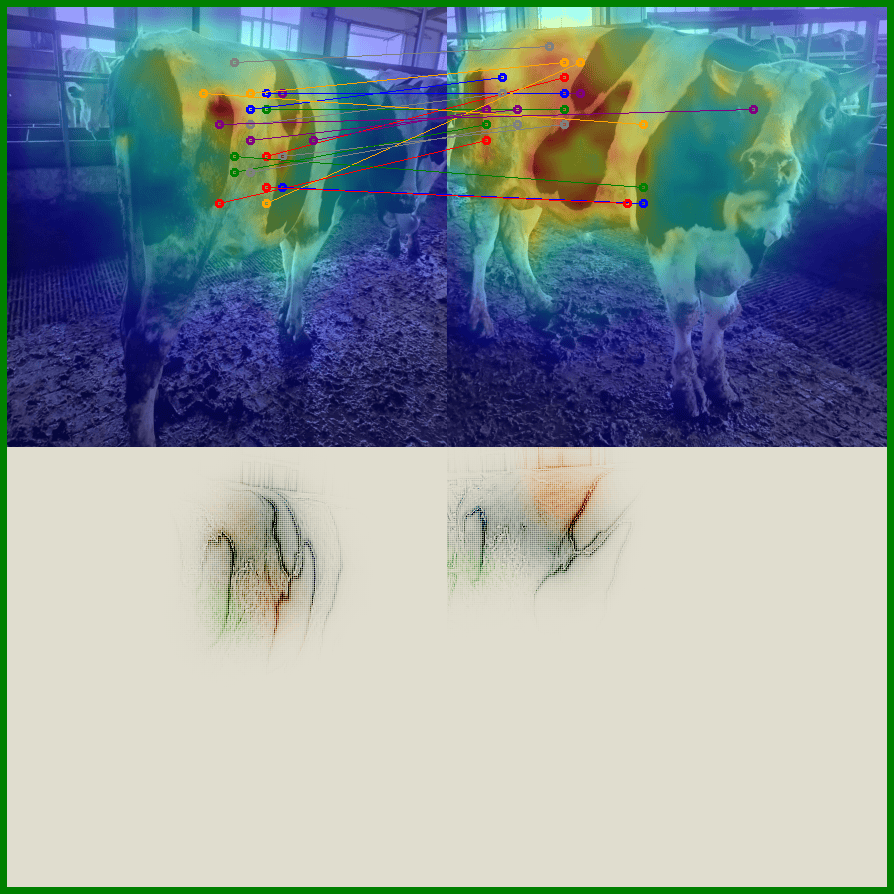}{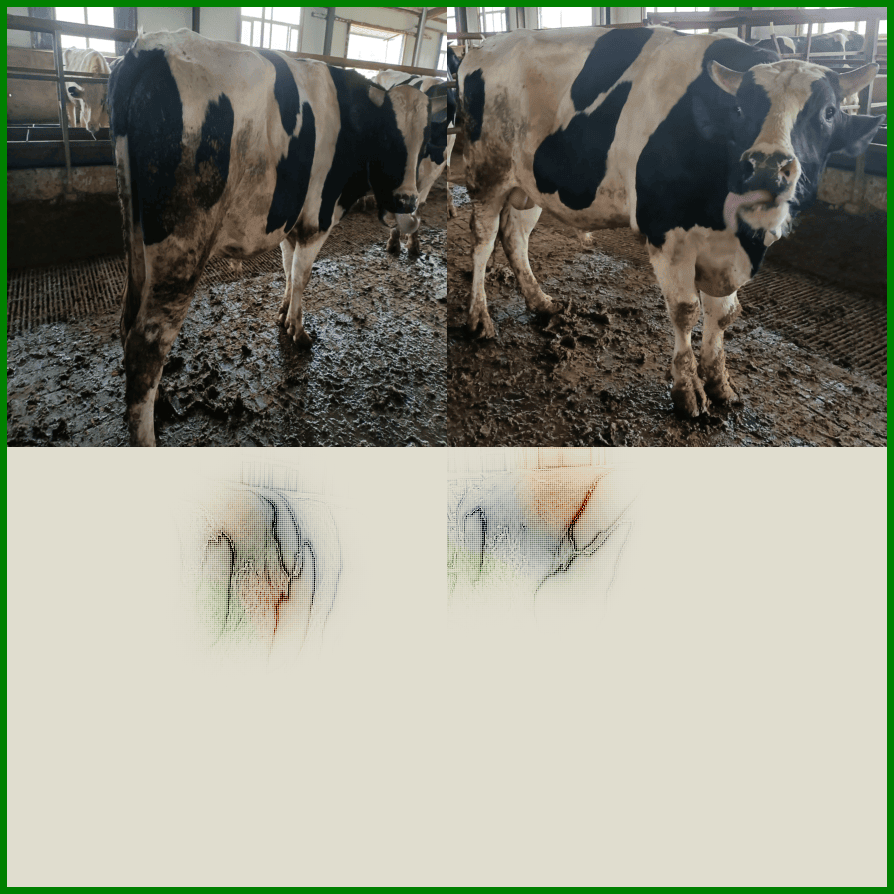}}
    \end{minipage}
    
    \caption{Qualitative analysis of trends in PAIR-X outputs.}
    \label{fig:species_ablation}
\end{figure*}

\section{Results}
\label{sec:results}



\begin{table}[h!]
\begin{threeparttable}
\caption{Quantitative Metrics Across Datasets}
\label{tab:dataset_ablation_quantitative}

\begin{tabular}{c|c c c c}
    \toprule
    \textbf{Dataset} & $\rho_{res}$ & $\Delta_{res}$ & $\rho_{mc}$ & $\Delta_{mc}$\\
    \midrule
AAUZebraFish~\cite{Haurum_2020_zebra_fish} & 0.73 & \cellcolor{green!56}0.56 & 0.75 & \cellcolor{green!78}0.78 \\
ATRW~\cite{ATRW} & 0.72 & \cellcolor{green!15}0.16 & 0.79 & \cellcolor{green!17}0.18 \\
BelugaIDv2~\cite{BelugaIDv2} & 0.27 & \cellcolor{red!0}-0.01 & 0.43 & \cellcolor{red!0}-0.00 \\
BirdIndividualID~\cite{BirdIndividualID} & 0.62 & \cellcolor{green!7}0.07 & 0.71 & \cellcolor{green!3}0.03 \\
CTai~\cite{CTai} & 0.24 & \cellcolor{green!8}0.09 & 0.57 & \cellcolor{green!7}0.08 \\
CZoo~\cite{CTai} & 0.27 & \cellcolor{green!13}0.14 & 0.36 & \cellcolor{green!23}0.24 \\
CatIndividualImages~\cite{CatIndividualImages} & 0.88 & \cellcolor{green!8}0.08 & 0.92 & \cellcolor{green!12}0.13 \\
CowDataset~\cite{CowDataset} & 0.70 & \cellcolor{green!39}0.39 & 0.85 & \cellcolor{green!77}0.78 \\
Cows2021v2~\cite{Cows2021} & 0.77 & \cellcolor{green!74}0.74 & 0.80 & \cellcolor{green!100}1.32 \\
\textbf{DogFaceNet}~\cite{DogFaceNet} & \textbf{0.48} & \cellcolor{red!0}\textbf{-0.01} & \textbf{0.70} & \cellcolor{green!1}\textbf{0.02} \\
ELPephants~\cite{ELPephants} & 0.21 & \cellcolor{green!8}0.09 & 0.32 & \cellcolor{green!16}0.16 \\
FriesianCattle2015v2~\cite{FriesianCattle2015v2} & 0.74 & \cellcolor{green!22}0.22 & 0.81 & \cellcolor{green!23}0.24 \\
FriesianCattle2017~\cite{FriesianCattle2017} & 0.74 & \cellcolor{green!7}0.08 & 0.89 & \cellcolor{green!5}0.06 \\
GiraffeZebraID~\cite{GiraffeZebraID} & 0.38 & \cellcolor{green!15}0.15 & 0.41 & \cellcolor{green!40}0.40 \\
\textbf{Giraffes}~\cite{Giraffes} & \textbf{0.74} & \cellcolor{green!56}\textbf{0.57} & \textbf{0.75} & \cellcolor{green!85}\textbf{0.85} \\
HyenaID2022~\cite{HyenaID2022} & 0.46 & \cellcolor{red!2}-0.03 & 0.63 & \cellcolor{green!1}0.02 \\
IPanda50~\cite{IPanda50} & 0.38 & \cellcolor{green!24}0.24 & 0.44 & \cellcolor{green!21}0.21 \\
LeopardID2022~\cite{LeopardID2022} & 0.50 & \cellcolor{green!3}0.04 & 0.54 & \cellcolor{green!1}0.01 \\
LionData~\cite{LionData} & 0.18 & \cellcolor{red!4}-0.05 & 0.28 & \cellcolor{green!3}0.03 \\
MacaqueFaces~\cite{MacaqueFaces} & 0.24 & \cellcolor{green!4}0.05 & 0.43 & \cellcolor{green!8}0.09 \\
NDD20v2~\cite{NDD20v2} & 0.44 & \cellcolor{green!4}0.04 & 0.56 & \cellcolor{green!9}0.10 \\
NOAARightWhale~\cite{NOAARightWhale} & 0.45 & \cellcolor{green!16}0.16 & 0.62 & \cellcolor{green!21}0.22 \\
NyalaData~\cite{NyalaData} & 0.30 & \cellcolor{green!23}0.23 & 0.62 & \cellcolor{green!13}0.14 \\
OpenCows2020~\cite{Andrew_2021} & 0.78 & \cellcolor{green!31}0.31 & 0.80 & \cellcolor{green!27}0.27 \\
ReunionTurtles~\cite{ReunionTurtles} & 0.50 & \cellcolor{green!18}0.18 & 0.74 & \cellcolor{green!12}0.13 \\
SMALST~\cite{SMALST} & 0.46 & \cellcolor{red!48}-0.48 & 0.67 & \cellcolor{red!18}-0.19 \\
SeaStarReID2023~\cite{SeaStarReID2023} & 0.64 & \cellcolor{green!38}0.38 & 0.68 & \cellcolor{green!28}0.29 \\
SeaTurtleID2022~\cite{SeaTurtleID2022} & 0.26 & \cellcolor{green!16}0.17 & 0.28 & \cellcolor{green!14}0.14 \\
SeaTurtleIDHeads~\cite{SeaTurtleID2022} & 0.26 & \cellcolor{green!22}0.22 & 0.31 & \cellcolor{green!15}0.16 \\
SealID~\cite{SealID} & 0.56 & \cellcolor{green!50}0.50 & 0.88 & \cellcolor{green!13}0.13 \\
SouthernProvinceTurtles~\cite{SouthernProvinceTurtles} & 0.69 & \cellcolor{green!14}0.14 & 0.75 & \cellcolor{green!42}0.43 \\
WhaleSharkID~\cite{WhaleSharkID} & 0.19 & \cellcolor{green!30}0.31 & 0.73 & \cellcolor{green!5}0.05 \\
ZakynthosTurtles~\cite{ZakynthosTurtles} & 0.48 & \cellcolor{green!0}0.00 & 0.87 & \cellcolor{green!0}0.00 \\
ZindiTurtleRecall~\cite{ZindiTurtleRecall} & 0.46 & \cellcolor{green!2}0.03 & 0.71 & \cellcolor{green!9}0.10 \\
\textbf{Oxford5k}~\cite{Oxford5k} & \textbf{0.64} & \cellcolor{green!36}\textbf{0.37} & \textbf{0.69} & \cellcolor{green!57}\textbf{0.58 }\\
    \bottomrule
\end{tabular}
\smallskip
\begin{flushleft}
\begin{tablenotes}
\small
\raggedright
\item Inverted residual mean ($res$) and relevance-weighted match coverage ($mc$) across datasets, aggregated within each dataset using both Spearman's rank correlation coefficient ($\rho$) between each metric and model match score, and binned Bhattacharyya distance ($\Delta$) of each metric between correct and incorrect matches. \colorbox{green}{Positive $\Delta$ values} indicate that PAIR-X improves separation for similarly scored matches. Metrics for \textbf{bolded} datasets are shown in more detail in Figure \ref{fig:scatterplots}.
\newline
\end{tablenotes}
\end{flushleft} 
\end{threeparttable}
\vspace{-1cm}
\end{table}

\begin{figure}[h!]
\centering
\captionsetup{aboveskip=16pt, belowskip=5pt, font=small}

\subfloat[Three possible distributions of correct and incorrect matches using our PAIR-X metrics.]{\fbox{\includegraphics[width=.9\columnwidth]{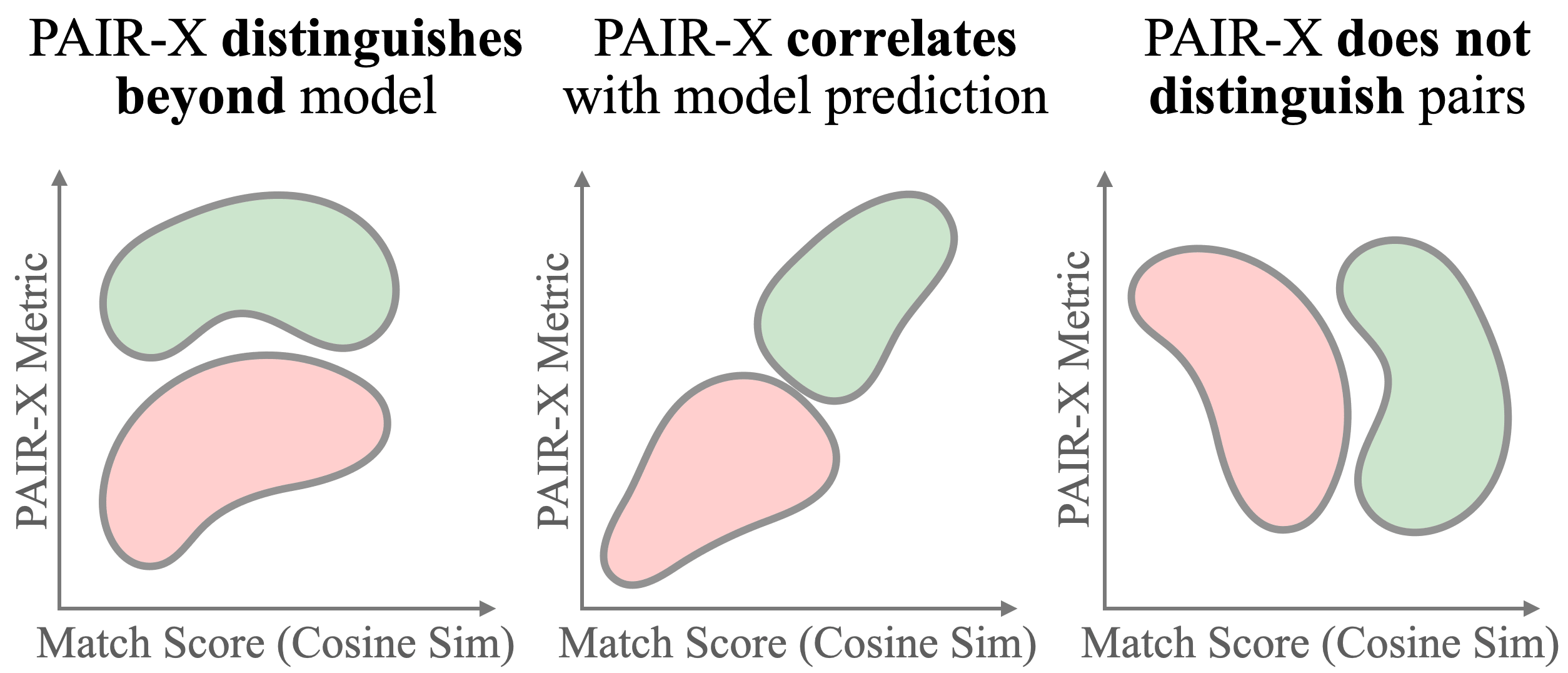}}}

\subfloat[The PAIR-X metric separates correctly and incorrectly matched pairs well for highly-patterned species such as giraffes, suggesting that PAIR-X proves highly explainable in these cases.]{\fbox{\includegraphics[width=.9\columnwidth]{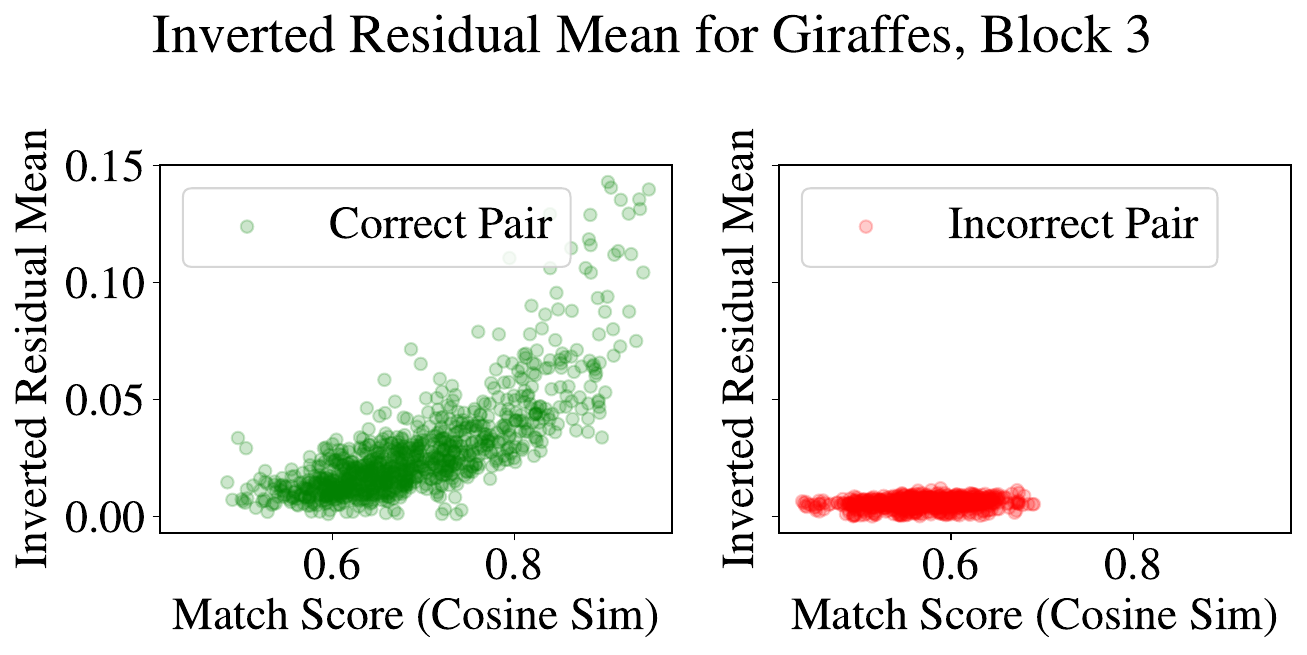}}} \\

\subfloat[The PAIR-X metric struggles to separate correctly and incorrectly matched pairs for less-highly-patterned species like dogs. However, this analysis did allow us to identify several outliers in the DogFaceNet dataset which were found to be mislabeled copies of the same images, with distortions such as rotations and cropping. The fact that PAIR-X is able to flag these mislabeled pairs is potentially useful. However, there are also incorrect pairs that receive moderately high scores from both the model and PAIR-X simply because they are similar-looking.]{
    \fbox{
        \begin{minipage}{.88\columnwidth}
            \includegraphics[width=\columnwidth]{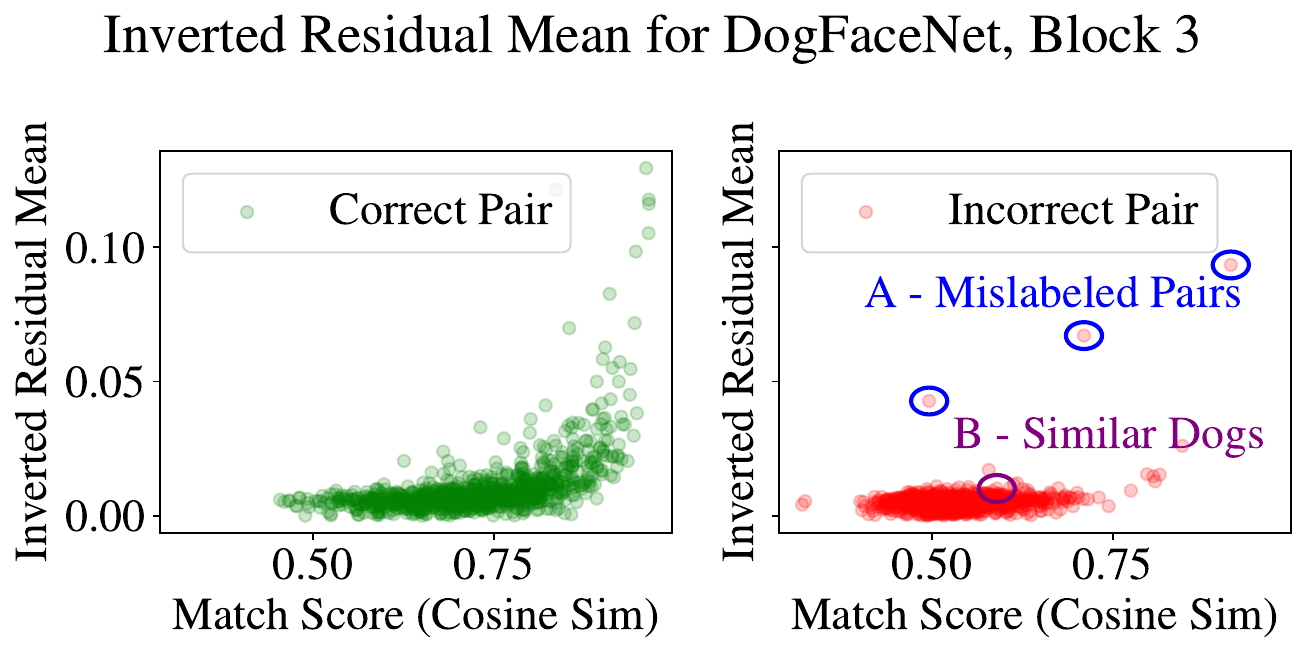} \\[5pt]
            \centering
            \hspace{.12\columnwidth}
            \subfloat[Mislabeled Pair]{\ToggleImages[width=0.38\columnwidth]{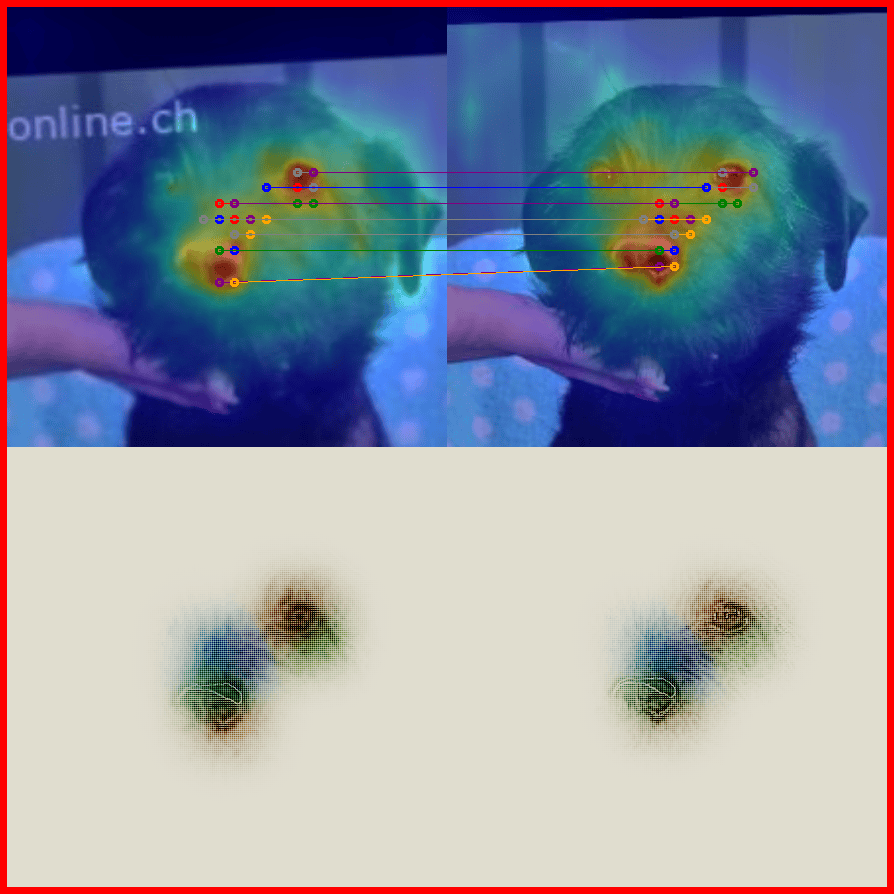}{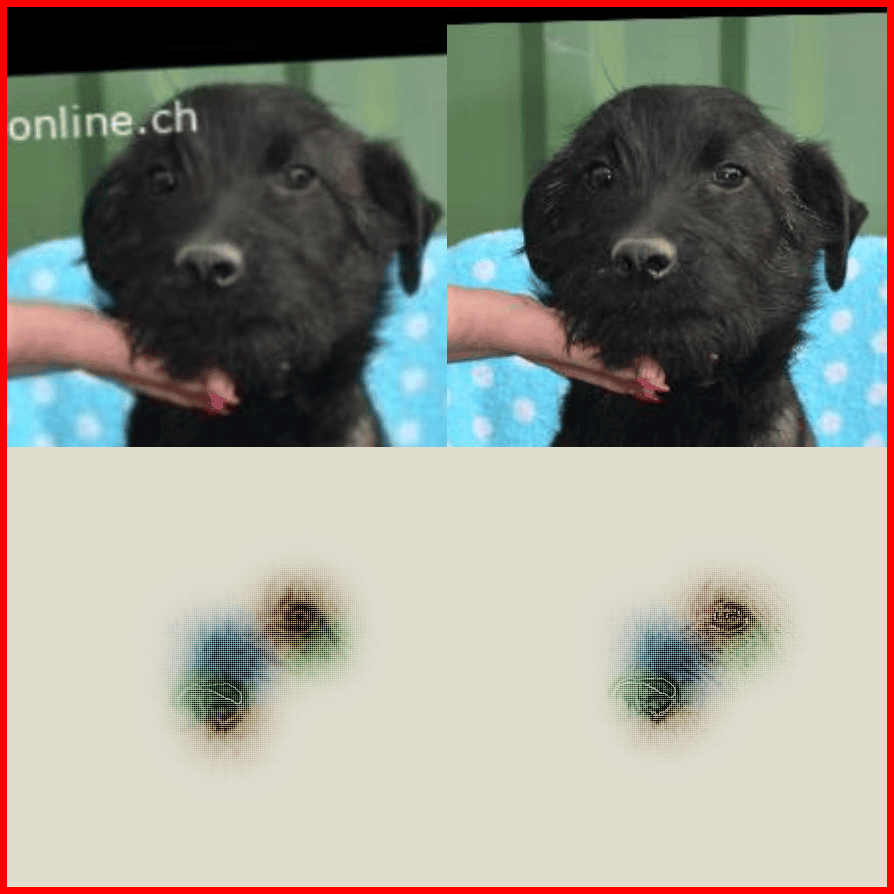}}
            \hspace{0.005\textwidth}
            \subfloat[Similar Dogs]{
            \ToggleImages[width=0.38\columnwidth]{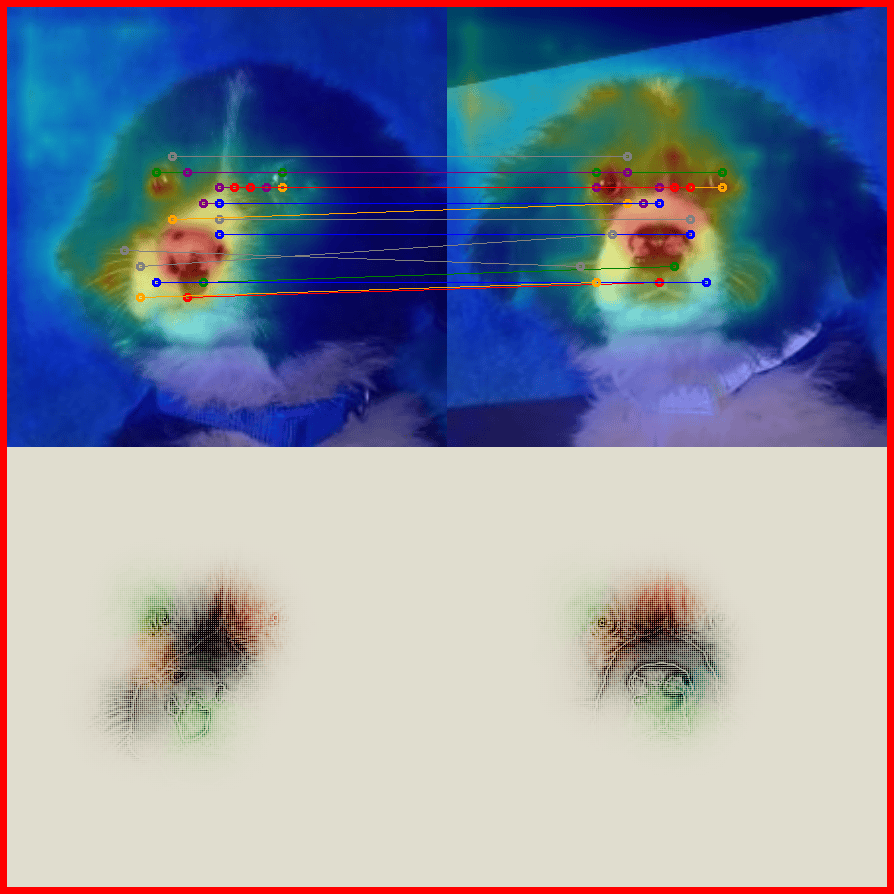}{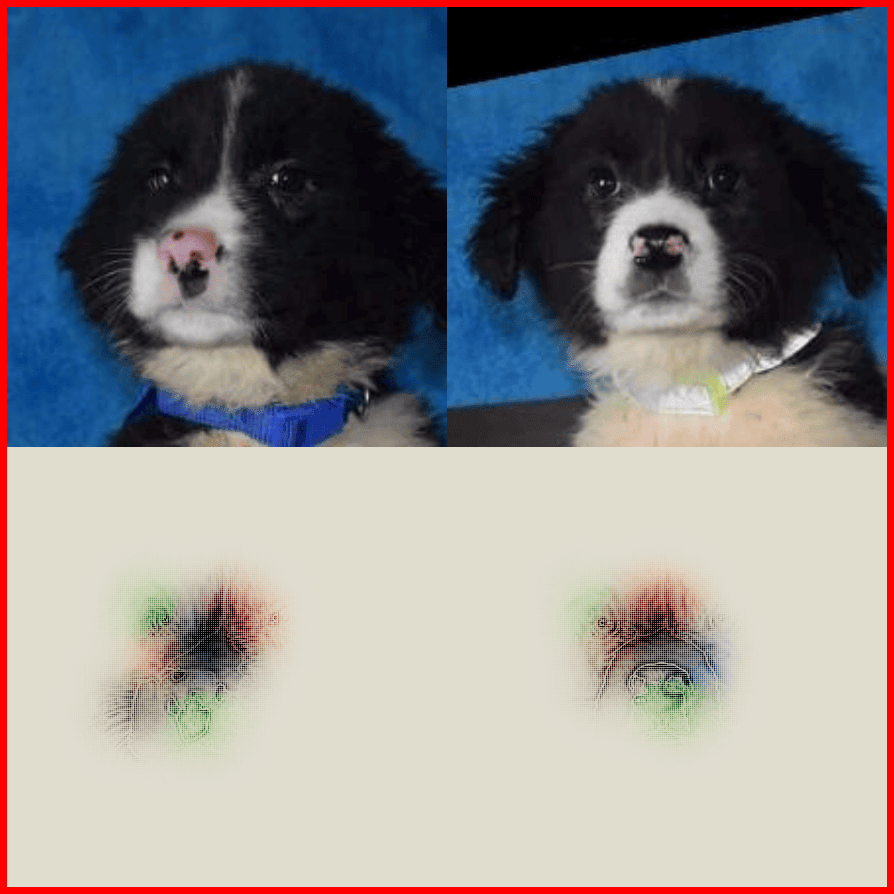}}
        \end{minipage}
    }
}
\vspace{-20pt}
\caption{We provide visualizations to build intuition for interpreting plots of PAIR-X metrics (inverted residual mean vs. match score), as well as real examples of these plots for a higher-performing (giraffe) and a lower-performing species (dogs).}
\label{fig:scatterplots}
\label{fig:diagram-scenarios}
\label{fig:dog_examples}
\end{figure}

Using the multispecies re-ID model Miew-ID as our deep metric model~\cite{otarashvili2024multispeciesanimalreidusing}, we evaluate PAIR-X across 34 public datasets from WildlifeDatasets~\cite{cermak2023wildlifedatasetsopensourcetoolkitanimal}, as well as the Oxford5k building dataset~\cite{Oxford5k}. See Suppl. Sec.~\ref{supsec:quantitative} for results on an additional model, and Suppl. Sec.~\ref{supsec:transformers} for a preliminary expansion to transformers. The metrics defined in Sec. \ref{sec:quantitative_metrics} (which we refer to as the PAIR-X metrics) are aggregated across each dataset using two additional values described below. Results are presented in Table \ref{tab:dataset_ablation_quantitative}. In Figure \ref{fig:scatterplots}, we visualize our metrics across individual image pairs in specific datasets.

\noindent \textbf{Dataset Metrics.} Both metrics are computed for a fixed set of image pairs for each dataset (see Suppl. Sec.~\ref{supsec:method} for details on pair selection). To aggregate these metrics across pairs within each dataset, we computed two additional values for each PAIR-X metric. First, we measure the rank correlation between the PAIR-X metric and the model similarity scores, to ascertain whether PAIR-X visualizations appear more plausible for pairs that the model scores more highly. This is done using Spearman's rank correlation coefficient $\rho$ (see Table \ref{tab:dataset_ablation_quantitative}). Second, we measure the separability of correct and incorrect pairs using the PAIR-X metric, while controlling for the model similarity score. The goal of this separability test is to ascertain whether, for correct and incorrect pairs that the model cannot distinguish, PAIR-X visualizations appear more plausible for correct matches than incorrect ones. This value is denoted as $\Delta$ in Table \ref{tab:dataset_ablation_quantitative}, and it is measured by binning over cosine similarity, then taking a weighted average of bin-wise Bhattacharyya distances (see~\cref{sec:metrics_appendix} for exact details).\\
In Figure \ref{fig:diagram-scenarios}, we visualize and discuss possible distributions of our metrics for correct and incorrect pairs, to develop intuition for the meaning of these metrics.


\vspace{10pt}
\paragraph{PAIR-X performs best on fine-grained tasks with highly-patterned or highly-localized features.} As an example, the giraffe data in WildlifeDatasets consists of high-quality, well-cropped images of Reticulated Giraffes, a species with dense, uniquely oriented patterns of highly-localized features. As shown in Fig.~\ref{fig:scatterplots}, the PAIR-X scores for this dataset follow two relevant trends. First, we see a strong positive correlation between match score and PAIR-X score, indicating that PAIR-X visualizations are more plausible for higher-scoring image pairs. It also suggests that PAIR-X is unlikely to produce highly plausible but misleading visualizations for image pairs with low model match scores. Second, we see that the PAIR-X scores show an additional dimension of separability between correct and incorrect pairs. For image pairs that the model assigns equivalent match scores, the PAIR-X metrics suggest that visualizations appear, on average, more plausible for correct than incorrect pairs. This is a promising result, and if additional separability between correct and incorrect pairs can be achieved through this type of method, it could perhaps be directly utilized to improve model accuracy.

Our method shows potential for fine-grained explainability beyond animals, particularly for other tasks with highly-structured and localized features such as building facades. We conduct a detailed analysis of performance on the Oxford5k dataset in Suppl. Sec.~\ref{sec:oxford5k}.


\noindent \textbf{PAIR-X is less well-suited for fine-grained tasks with less-localized or less-structured distinguishing features}. As an example, for images in DogFaceNet, we see a much lower degree of separability between correct and incorrect pairs. Qualitatively, we see that this is largely due to structural similarities between individuals: PAIR-X is, for instance, likely to find matches between eyes and noses of incorrectly-matched dog pairs, especially when comparing different individuals from the same breed. This raises the question of what optimal explainability would look like in this case, where identification may be more gestalt than localizeable (\ie subtle variations in relative spacing between eyes and nose, as opposed to unique patterns of stripes). As we see in Figure \ref{fig:dog_examples}, many of the incorrect pairs that receive high model similarity scores are of very similar-looking dogs. In these cases, PAIR-X provides informative visualizations of the features that contribute most to similarity (\eg eyes, nose). However, as these facial features contribute to similarity for both correct and incorrect matches, PAIR-X may produce misleading visualizations for incorrect matches, and is thus not as useful for manual review of model predictions.
 This is in contrast to highly patterned species, where highly-matched but uniquely-structured patterns for each individual are easily distinguished in our visualizations, and thus lead to higher PAIR-X metric scores.
Examples on additional less-patterned species (cats and starfish) are shown in Figure \ref{fig:species_ablation}.

\section{Discussion}
\subsection{Quantitative comparisons between methods} Ideally it would be possible to not only quantify the performance of our method across datasets, but also quantitatively compare our method to other explainability techniques. General-purpose quantifiable metrics for explainability are still out of reach, and our PAIR-X metrics assume explicit pairwise feature matching to calculate, and are thus not directly applicable to, e.g., CAM-based methods. Thus, we rely on qualitative comparisons and expert interviews to measure the relative interpretability and usefulness of explainations from different methods. That said, we want to highlight the value that our method-specific metrics provide. The ability to quantify explainability in PAIR-X allows a user to efficiently determine whether PAIR-X is a good fit for their task of interest.
\subsection{Applicability to real-world use cases}

In an envisioned use case for animal re-ID, PAIR-X visualizations could be used to more efficiently manually validate model predictions, especially in cases where the closest correct match and closest incorrect match are scored similarly. Qualitatively, PAIR-X visualizations help to isolate important information and to visually align images, reducing the manual labor required for match verification. As discussed in Section \ref{sec:results}, the difference in PAIR-X metrics between correct and incorrect pairs with similar match scores suggests that visualizations for correct pairs would, on average, appear more visually plausible. This is especially true for datasets with a high degree of separability, as measured by $\Delta$ in Table \ref{tab:dataset_ablation_quantitative}. We further analyze the applicability of PAIR-X to real-world use cases via expert interviews and a brief analysis of computational costs.


\paragraph{Expert interviews.} Because explainability is highly subjective, we found it important to collect perspectives from downstream model users about the usability of PAIR-X. Animal re-ID is a niche topic, with a very limited number of experts capable of manual re-ID on these datasets, which limited the pool of people from whom to collect feedback. However, we interviewed three experts in giraffe re-ID, and discuss a few key insights gained from those conversations.

In real-world deployments of giraffe re-ID models, experts are tasked with manually verifying large batches of image labels, which frequently requires reviewing between five and twenty top-ranked database matches per query image. In this setting, efficiency is very important. The experts we interviewed agreed that explainability visualizations that highlight relevant image regions are helpful for directing user attention and speeding up verification.

\vspace{10pt}
We asked experts about their preferences between PAIR-X, SIFT feature matching such as HotSpotter, and Grad-CAM++. While experts found Grad-CAM++ to be more useful than \textit{no} explainability, they found the fine-grained information provided by PAIR-X to be more helpful. Between classical SIFT feature matching and PAIR-X, experts were split. One expert had used HotSpotter extensively before switching to deep models, and thus had a preference for the classical feature-matching visualization, but recognized that its inability to scale to their current database rendered it no longer usable. Another expert, who had not previously used HotSpotter, found those visualizations to contain too many matches to be interpretable, and preferred PAIR-X for its filtered set of feature matches.

\paragraph{Computational efficiency.} Since experts are interactively reviewing large numbers of images (one of the experts we interviewed had reviewed more than 100,000 during their time in the field), low latency, and therefore computational efficiency, is essential. On a single A100 GPU, we find that creating a typical explanation for 10 backpropagated matches requires 5 seconds, which demonstrates the feasibility of using PAIR-X at scale. We expand upon the factors influencing computational efficiency in~\cref{sec:computational_efficiency}.

\section{Conclusion}

We present PAIR-X, a novel fine-grained explainability technique based on a combination of deep feature matching and layer-wise relevance propagation (LRP), which provides explanations of pairwise similarity based on pretrained deep metric learning models. We demonstrate promising results on a diverse collection of animal re-ID datasets, as well as the Oxford-5k building dataset. Qualitatively, the explanations produced by PAIR-X are finer-grained than existing CAM-based techniques, as well as easier to interpret thanks to explicit matching of relevant image features and the color-coded propogation of those features back into image space. We furthermore propose a set of quantitative metrics which show that PAIR-X is in many cases able to distinguish correct from similarly scoring incorrect (\ie confusing) matches. While PAIR-X may produce misleading explanations for species with a high degree of structural similarity, we show that there are many patterned species PAIR-X is applicable to. The experts we interviewed unanimously agreed that PAIR-X explanations are useful and informative, and WildMe\footnote{https://www.wildme.org/}, a cross-species animal re-identification platform, has expressed intent to deploy our method for all of its patterned species, emphasizing its applicability and usefulness in real-world settings.

\section*{Acknowledgements}

Research was sponsored by the Department of the Air Force Artificial Intelligence Accelerator and was accomplished under Cooperative Agreement Number FA8750-19-2-1000. The views and conclusions contained in this document are those of the authors and should not be interpreted as representing the official policies, either expressed or implied, of the Department of the Air Force or the U.S. Government. The U.S. Government is authorized to reproduce and distribute reprints for Government purposes notwithstanding any copyright notation herein. This work was supported in part by the MIT-IBM Watson AI Lab. This work was supported by the AI and Biodiversity Change (ABC) Global Center, which is funded by the US National Science Foundation under Award No. 2330423 and Natural Sciences and Engineering Research Council of Canada under Award No. 585136. This work draws on research supported in part by the Social Sciences and Humanities Research Council.

We also thank Jared Stabach from the Smithsonian Conservation Biology Institute, as well as Michael Brown and Courtney Marneweck from the Giraffe Conservation Foundation, for offering their expertise in the area of giraffe re-ID to provide feedback on our work. We also thank Lukas Picek and Vojtěch Čermák for providing us with trained MegaDescriptor re-ID models in different architectures for analysis.

{
    \small
    \bibliographystyle{ieeenat_fullname}
    \bibliography{main}
}

\newpage
\appendix
\section{Comparison of capabilities to prior work}

\begin{table*}
\centering
\resizebox{\textwidth}{!}{
\begin{threeparttable}

\caption{Comparison of explainability techniques}
\label{comparison_of_capabilities}

\begin{tabular}{@{} l *{4}{C{2.5cm}} @{}}
\toprule
&  Fine-grained resolution &  Pairwise spatial correspondences
&  Single visualization & Explains deep models \\
\midrule
Explainability for local feature matching \cite{hotspotter,curvrank} & \cmark & \cmark & \cmark & \xmark \\
CAM-based~\cite{gradcam, gradcam++, eigencam, hirescam, kpcacam, layercam, hirescam, ablationcam, scorecam, xgradcam, fullgrad} & \xmark & \xmark & \cmark & \cmark \\
SHAP~\cite{shap} & \cmark & \xmark & \cmark & \cmark \\
LRP~\cite{LRP} & \cmark & \xmark & \cmark & \cmark \\
CRP~\cite{CRP} & \cmark & \xmark & \xmark & \cmark\\
Point-to-point correspondences \cite{zhu2021visualexplanationdeepmetric} & \xmark & \cmark & \xmark & \cmark \\
PAIR-X & \cmark & \cmark & \cmark & \cmark\\
\bottomrule
\end{tabular}

\smallskip
\begin{tablenotes}[para, flushleft] 
\small
    Comparison of explainability techniques across four criteria. \textbf{Fine-grained resolution:} is the resolution of the explainability visualization finer than that of the final convolutional layer of a deep model? \textbf{Pairwise spatial correspondences:} are the spatial regions deemed similar by the model explicitly visualized? \textbf{Single visualization:} can results across spatial regions or concepts be encapsulated in a single visualization? \textbf{Explains deep models:} Is the explainability derived directly from a deep learning model?
\end{tablenotes}

\end{threeparttable}
}
\end{table*}
For an overview of key qualities of PAIR-X versus baselines, see Table~\ref{comparison_of_capabilities}.
\subsection{Choice of baselines}
\label{sec:baseline_ablation}

To select the explainability baselines presented in \cref{fig:methods_ablation}, we performed a wider ablation across a variety of CAM-based explainability techniques. We also performed a hyperparameter search for Kernel SHAP, and manually selected the query pixels used to visualize point-to-point activation intensities~\cite{zhu2021visualexplanationdeepmetric, shap}.

\begin{figure*}

    \centering
    \begin{tabular}{m{.30\textwidth}|m{.20\textwidth}m{.20\textwidth}m{.20\textwidth}}
        \toprule
        \textbf{Technique} & \textbf{Giraffes} & \textbf{Cows} & \textbf{Buildings} \\
        \midrule
\textbf{Grad-CAM}~\cite{gradcam, jacobgilpytorchcam} &
\includegraphics[width=.19\textwidth]{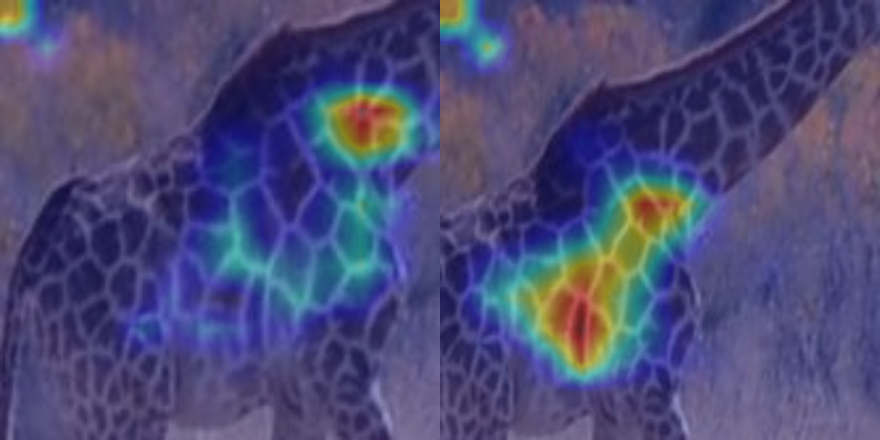} &
\includegraphics[width=.19\textwidth]{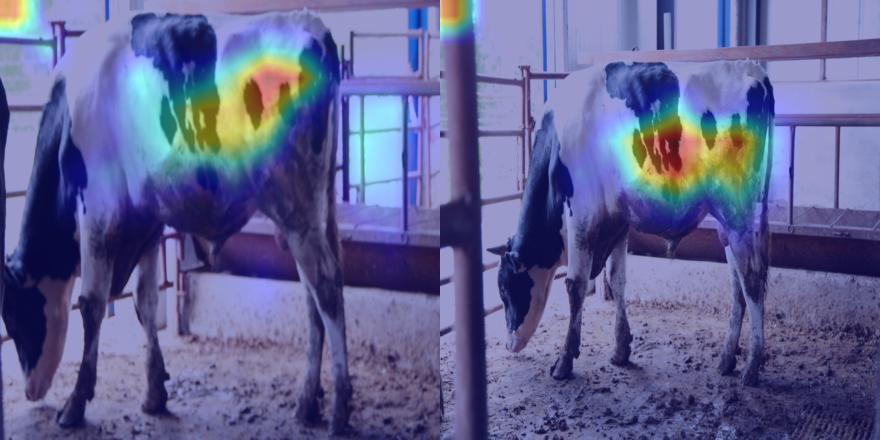} &
\includegraphics[width=.19\textwidth]{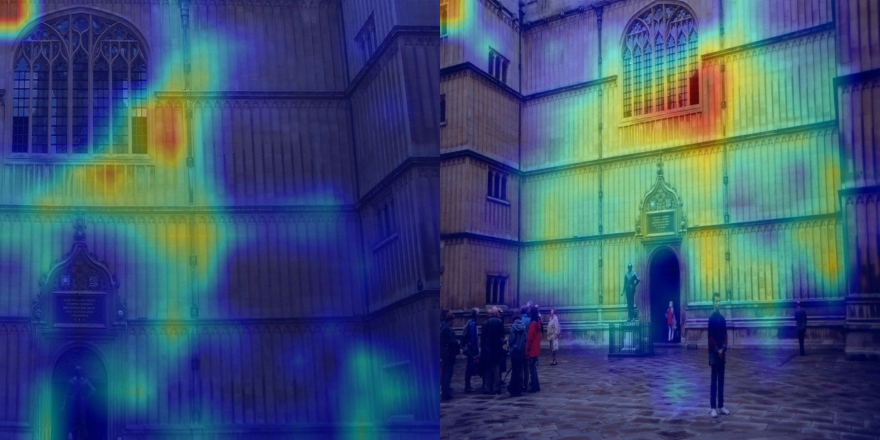} \\
\textbf{HiResCAM}~\cite{hirescam, jacobgilpytorchcam} &
\includegraphics[width=.19\textwidth]{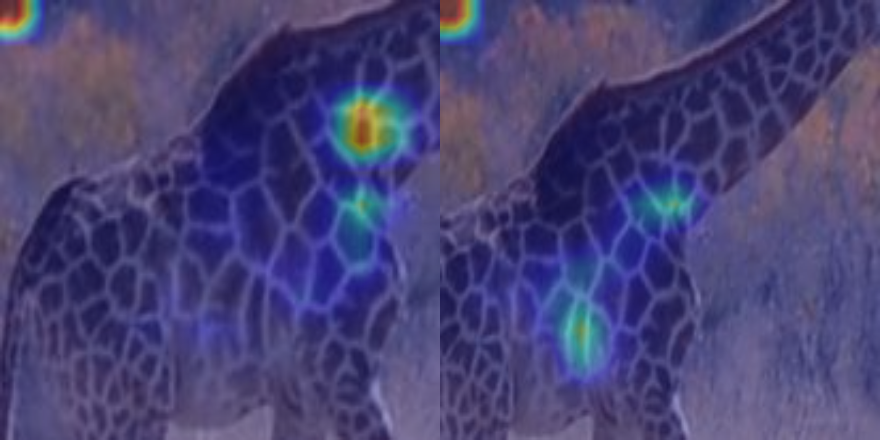} &
\includegraphics[width=.19\textwidth]{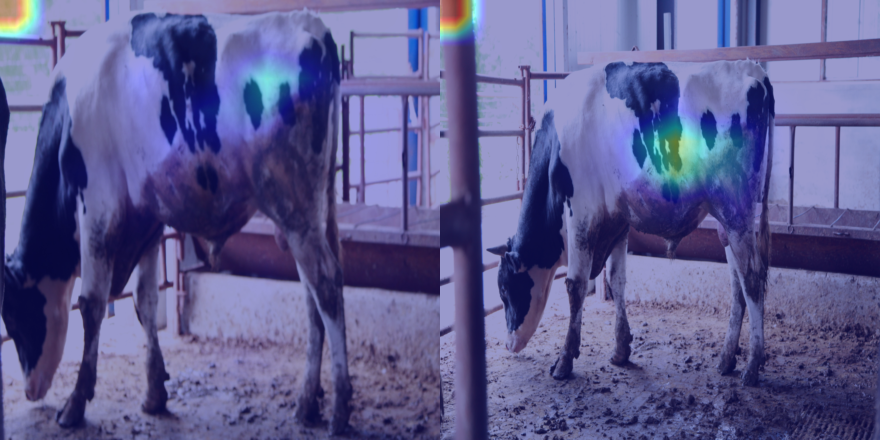} &
\includegraphics[width=.19\textwidth]{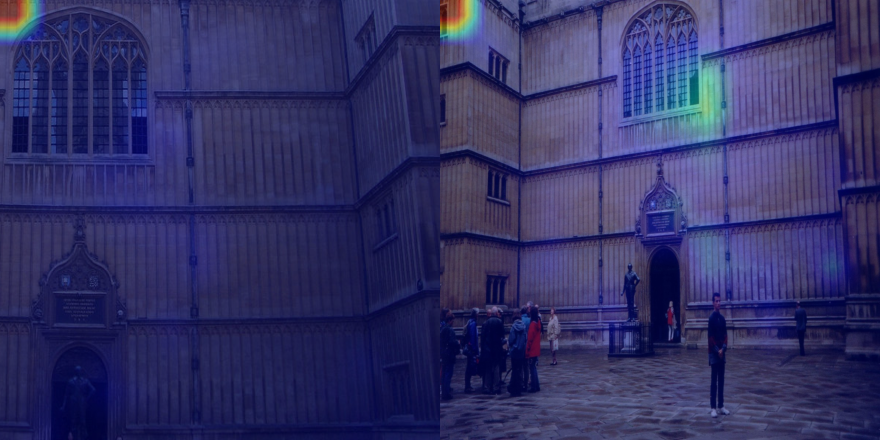} \\
\textbf{Score-CAM}~\cite{scorecam, jacobgilpytorchcam} &
\includegraphics[width=.19\textwidth]{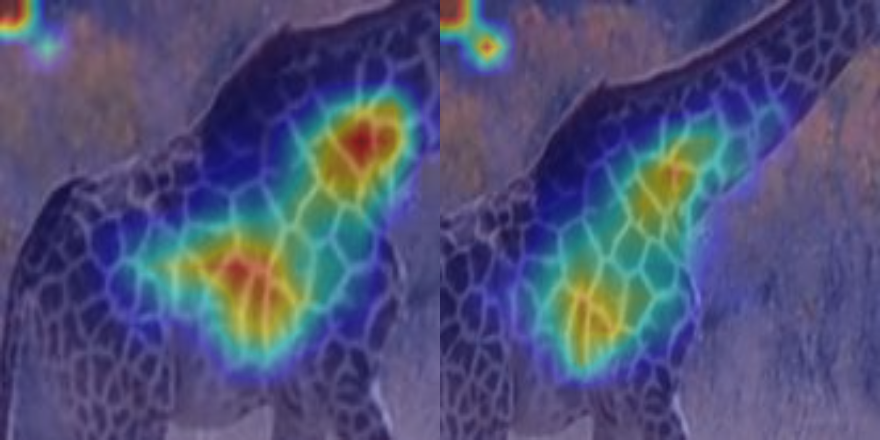} &
\includegraphics[width=.19\textwidth]{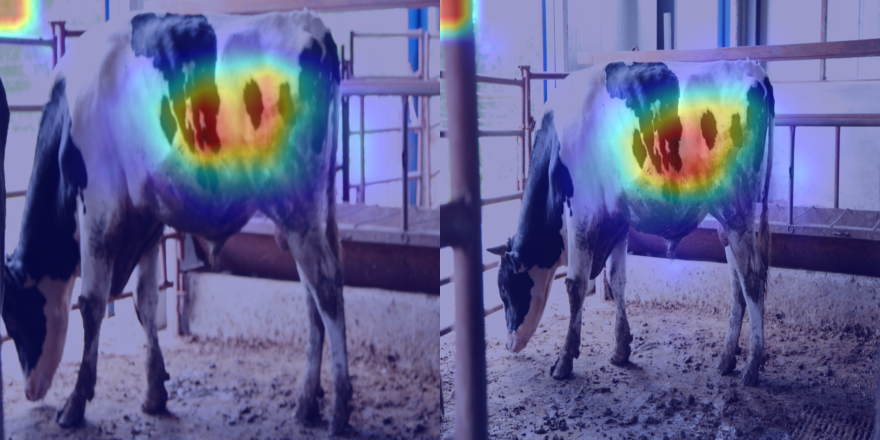} &
\includegraphics[width=.19\textwidth]{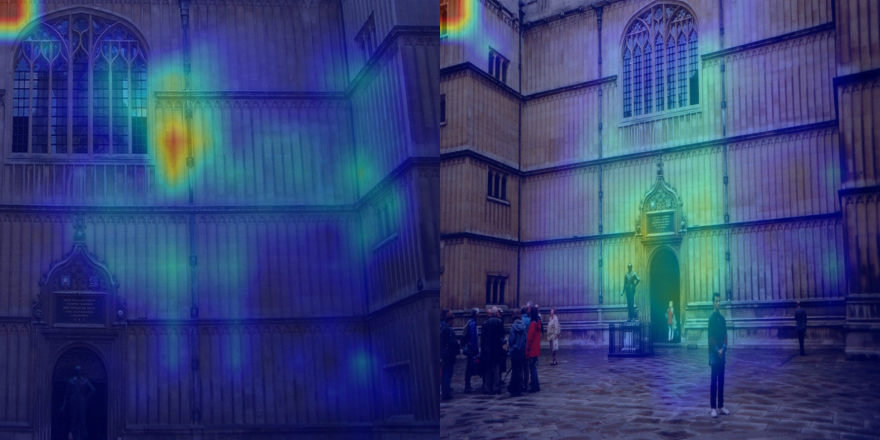} \\
\textbf{Grad-CAM++}~\cite{gradcam++, jacobgilpytorchcam} &
\includegraphics[width=.19\textwidth]{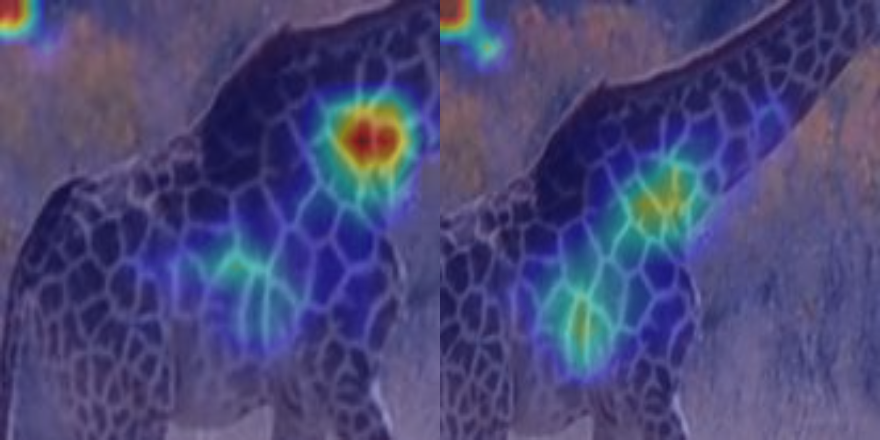} &
\includegraphics[width=.19\textwidth]{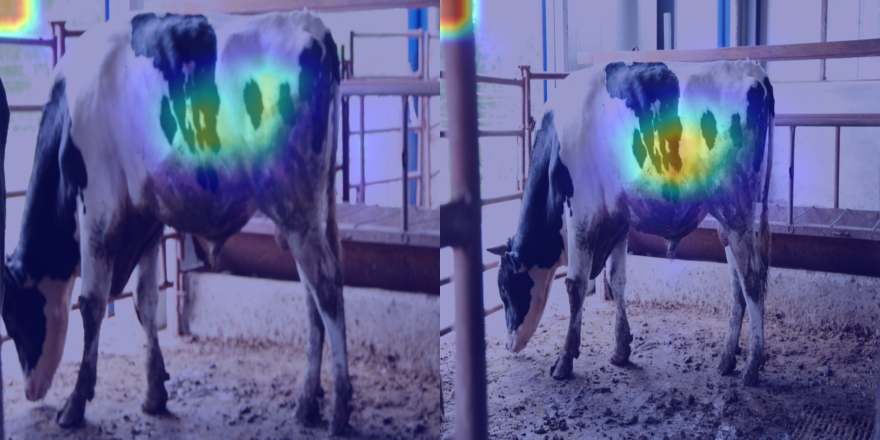} &
\includegraphics[width=.19\textwidth]{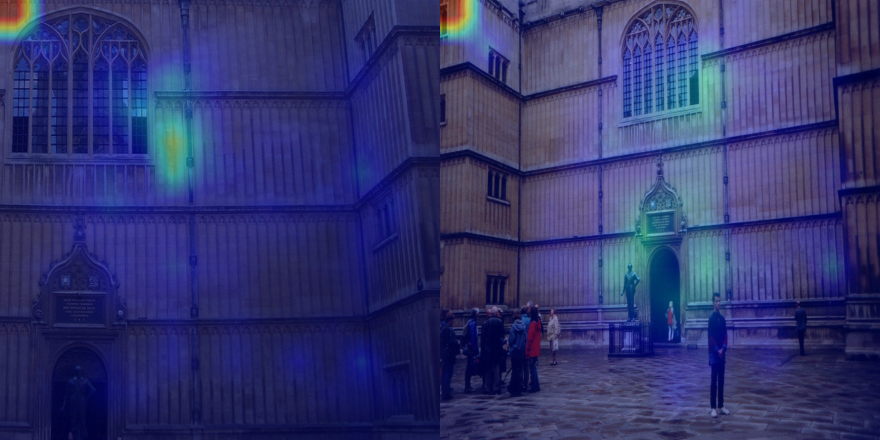} \\
\textbf{Ablation-CAM}~\cite{ablationcam, jacobgilpytorchcam} &
\includegraphics[width=.19\textwidth]{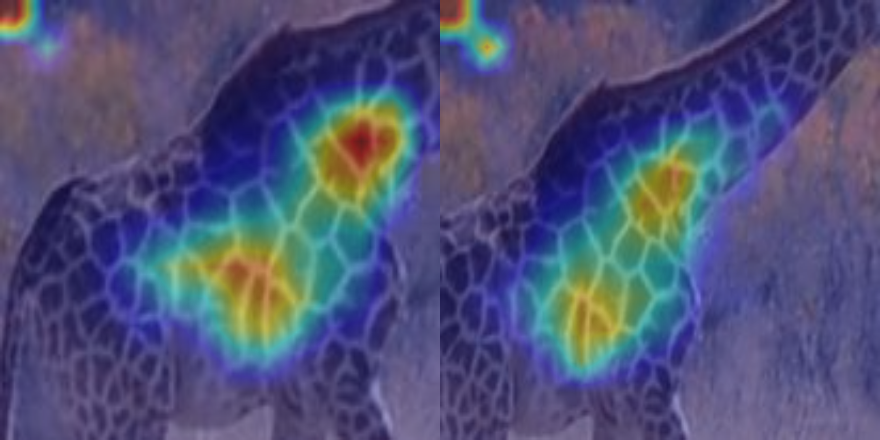} &
\includegraphics[width=.19\textwidth]{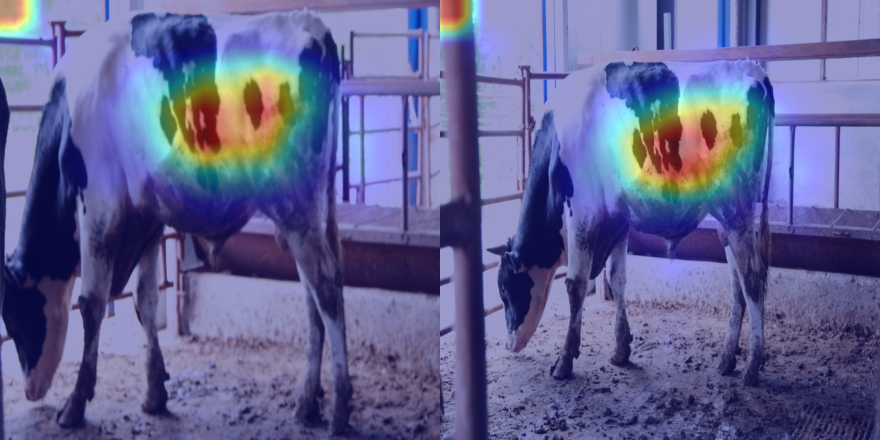} &
\includegraphics[width=.19\textwidth]{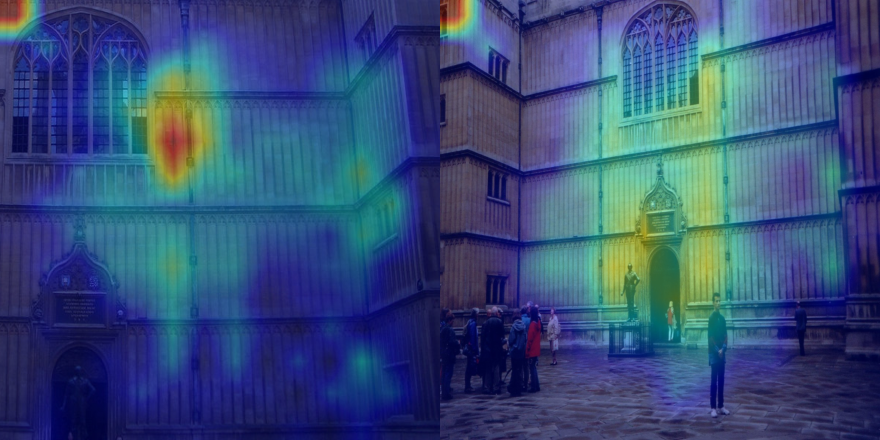} \\
\textbf{XGrad-CAM}~\cite{xgradcam, jacobgilpytorchcam} &
\includegraphics[width=.19\textwidth]{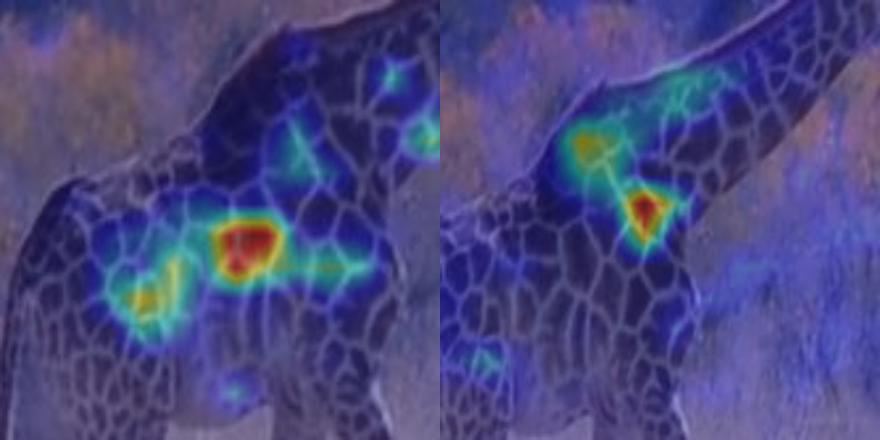} &
\includegraphics[width=.19\textwidth]{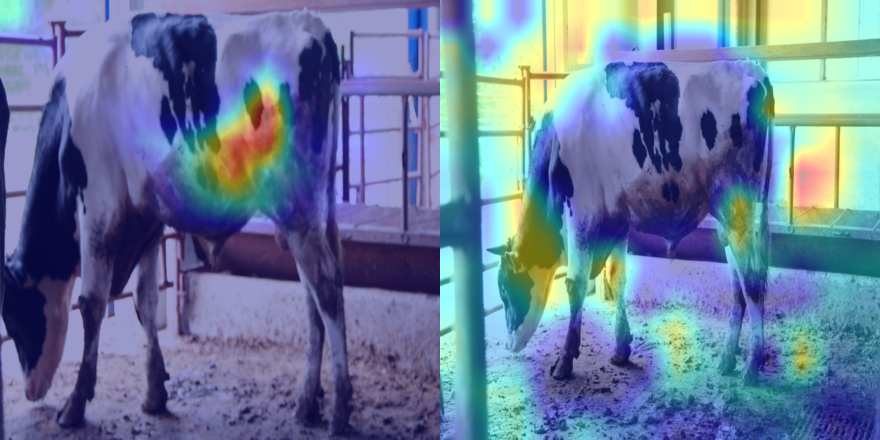} &
\includegraphics[width=.19\textwidth]{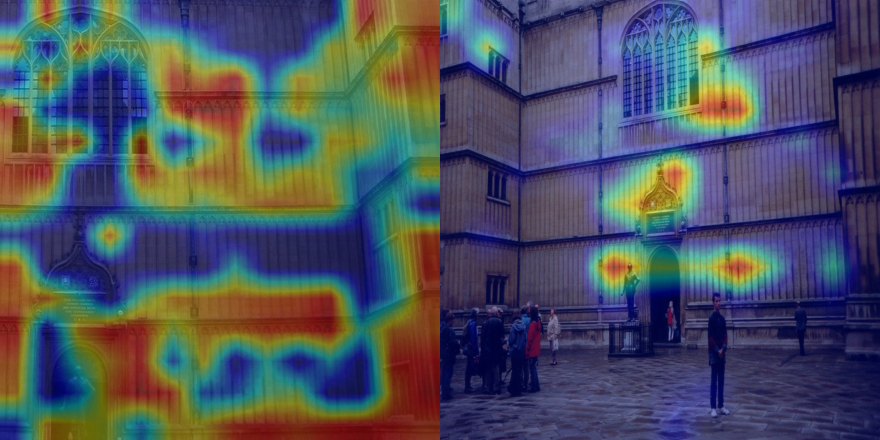} \\
\textbf{Eigen-CAM}~\cite{eigencam, jacobgilpytorchcam} &
\includegraphics[width=.19\textwidth]{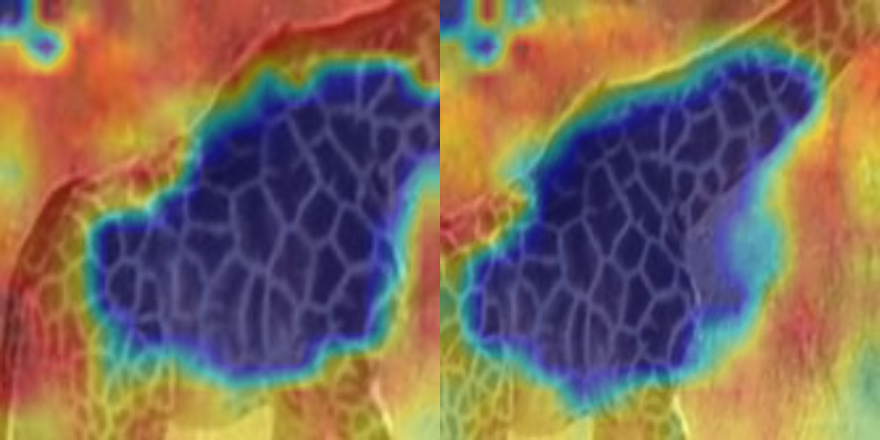} &
\includegraphics[width=.19\textwidth]{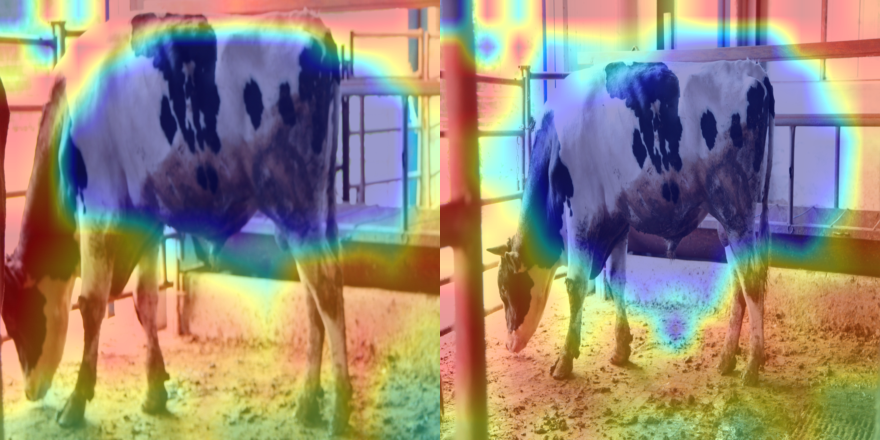} &
\includegraphics[width=.19\textwidth]{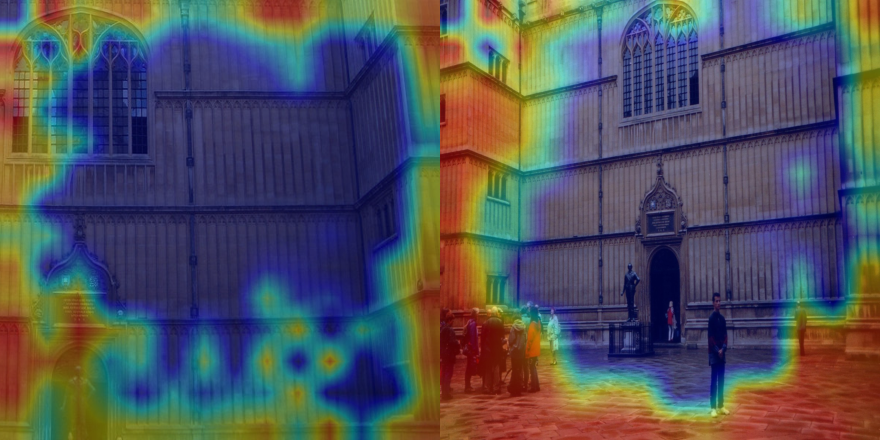} \\
\textbf{EigenGradCAM}~\cite{eigencam, jacobgilpytorchcam} &
\includegraphics[width=.19\textwidth]{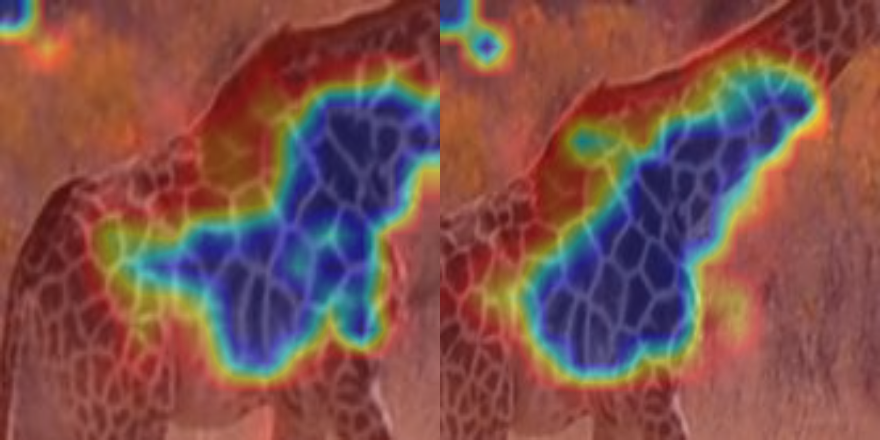} &
\includegraphics[width=.19\textwidth]{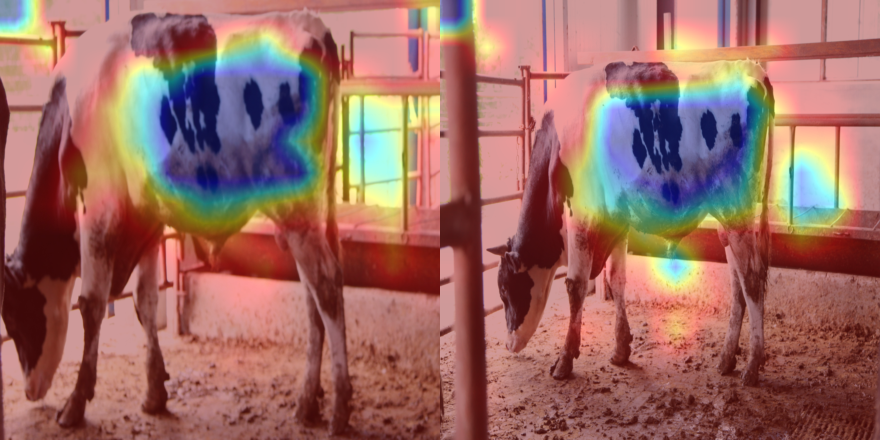} &
\includegraphics[width=.19\textwidth]{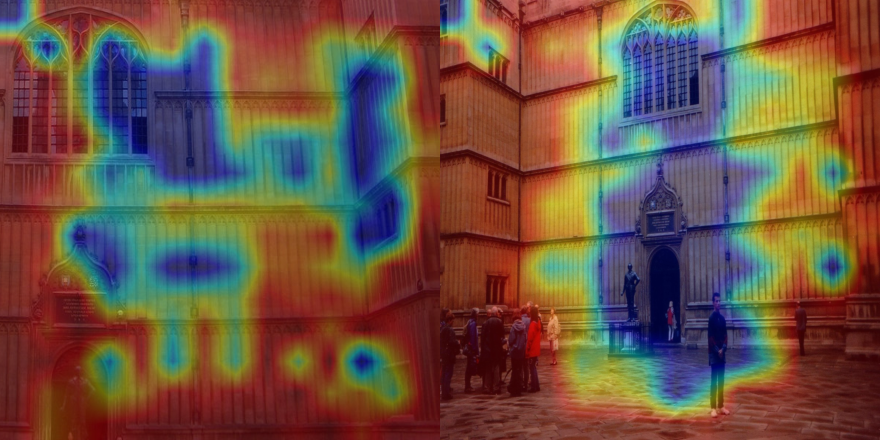} \\
\textbf{LayerCAM}~\cite{layercam, jacobgilpytorchcam} &
\includegraphics[width=.19\textwidth]{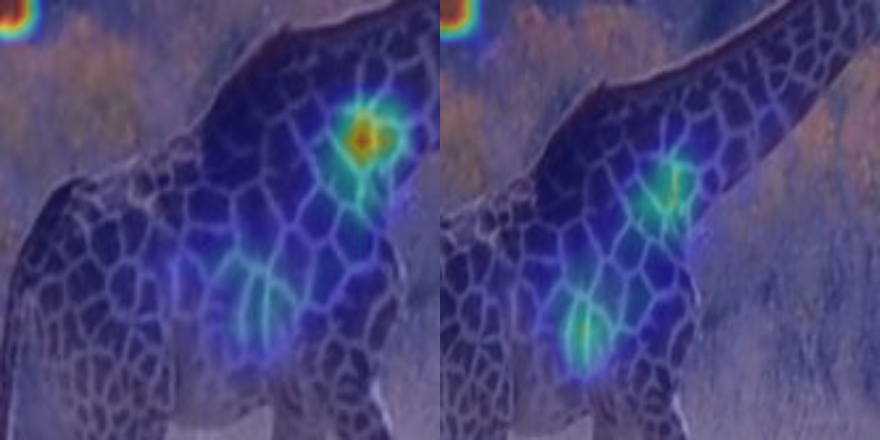} &
\includegraphics[width=.19\textwidth]{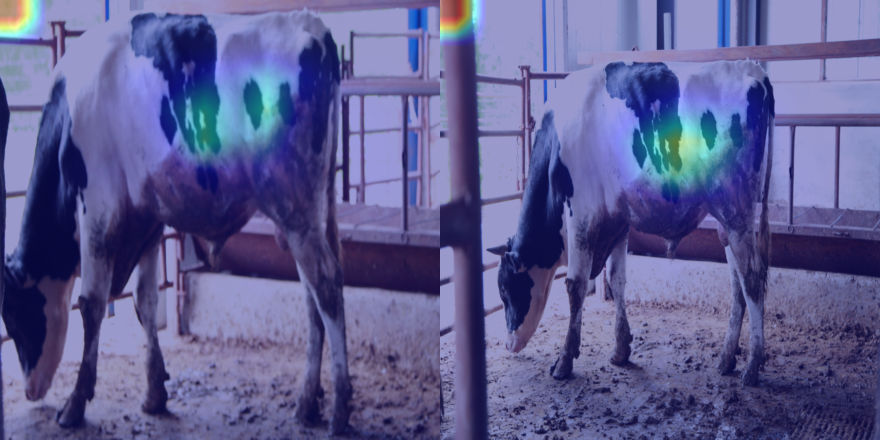} &
\includegraphics[width=.19\textwidth]{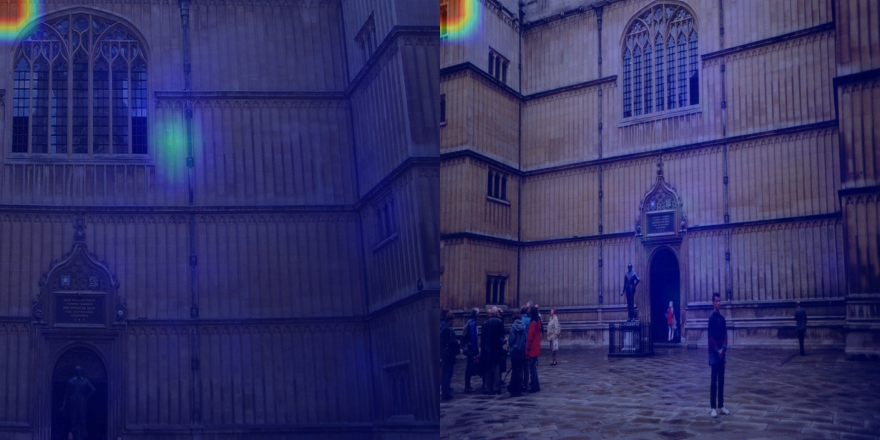} \\
\textbf{FullGrad}~\cite{fullgrad, jacobgilpytorchcam} &
\includegraphics[width=.19\textwidth]{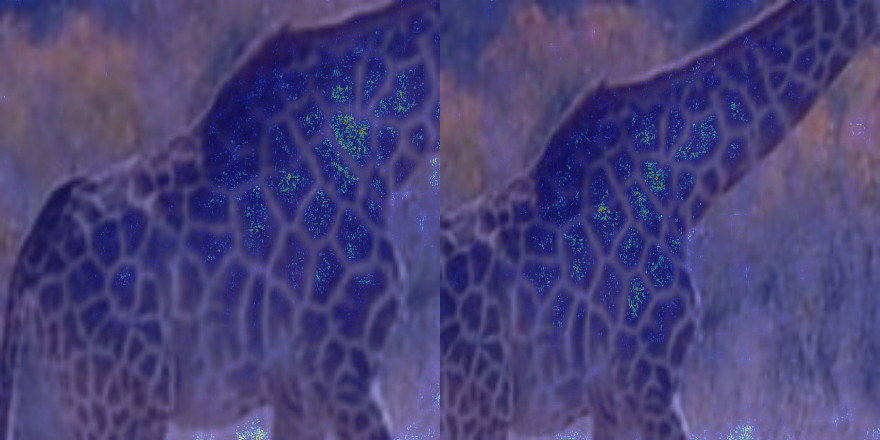} &
\includegraphics[width=.19\textwidth]{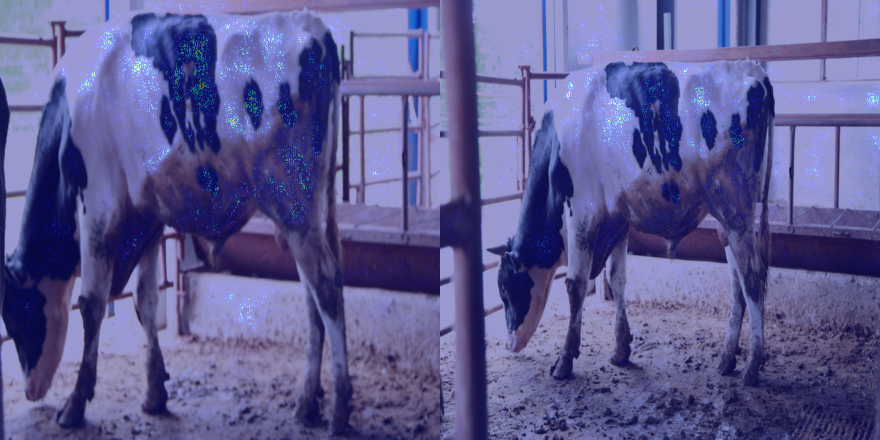} &
\includegraphics[width=.19\textwidth]{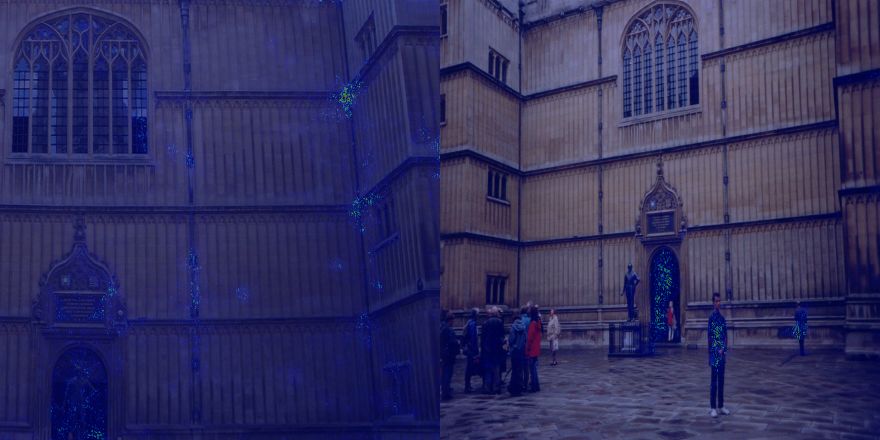} \\
\textbf{GradCAMElementWise}~\cite{gradcam, jacobgilpytorchcam} &
\includegraphics[width=.19\textwidth]{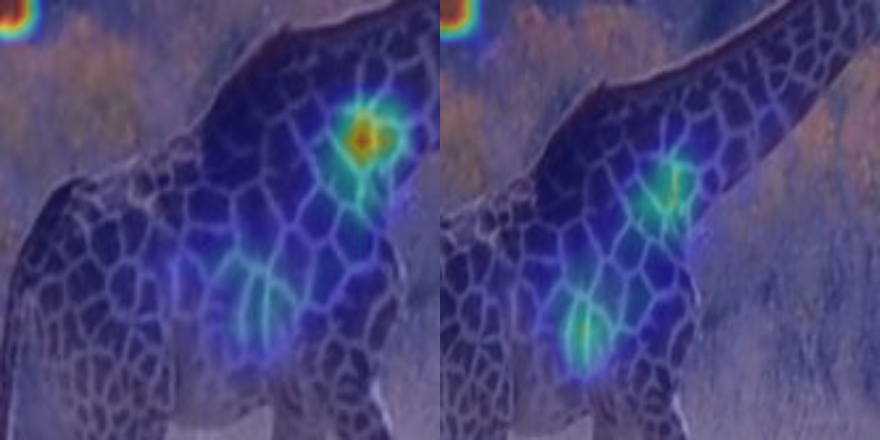} &
\includegraphics[width=.19\textwidth]{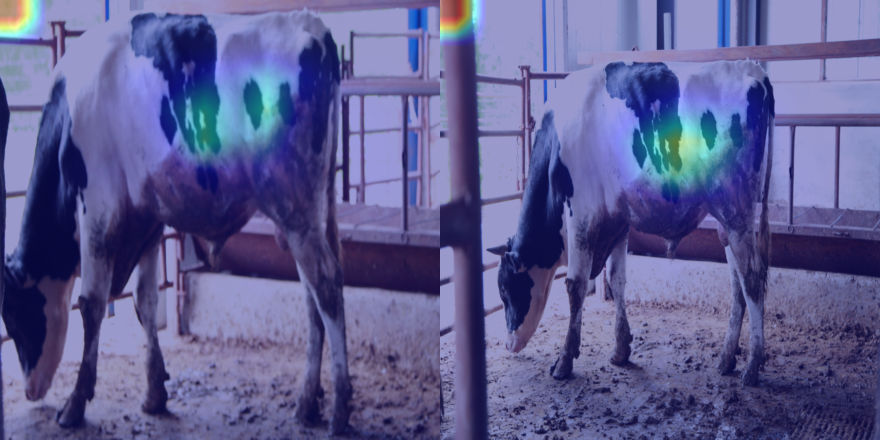} &
\includegraphics[width=.19\textwidth]{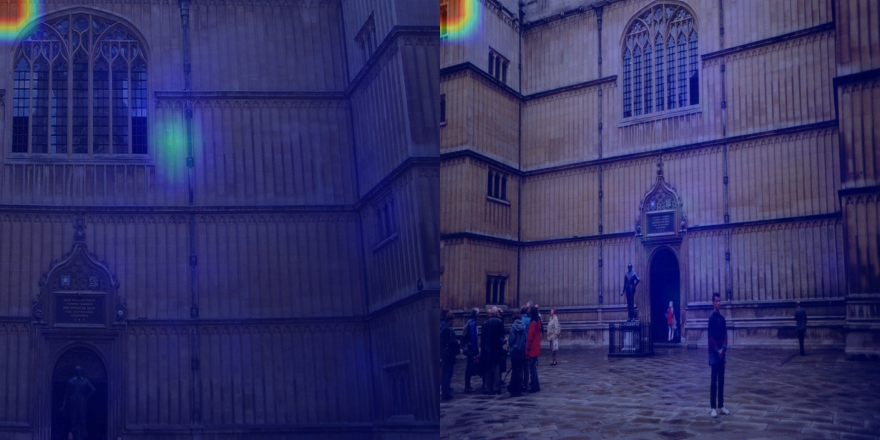} \\
\textbf{KPCA-CAM}~\cite{kpcacam, jacobgilpytorchcam} &
\includegraphics[width=.19\textwidth]{figures/methods_ablation/baseline_ablations/Giraffes_kpcacam.png} &
\includegraphics[width=.19\textwidth]{figures/methods_ablation/baseline_ablations/CowDataset_kpcacam.png} &
\includegraphics[width=.19\textwidth]{figures/methods_ablation/baseline_ablations/Oxford5k_kpcacam.png} \\
        \hline
    \end{tabular}
    \caption{We used~\cite{jacobgilpytorchcam} to compare twelve CAM-based techniques and qualitatively select the best-performing technique for our tasks. Based on this analysis, we chose to present KPCA-CAM~\cite{kpcacam} as our primary CAM-based baseline in Figure \ref{fig:methods_ablation}.}
    \label{fig:cam_baselines}

\end{figure*}

\noindent \textbf{CAM-based techniques.} We qualitatively compared twelve CAM-based techniques, using implementations from~\cite{jacobgilpytorchcam}. The results of this ablation are shown in Figure \ref{fig:cam_baselines}. Based on the ablation, we selected \textbf{KPCA-CAM} as the primary CAM-based technique for comparison.

\noindent\textbf{Perturbation-based techniques.} To directly measure the impact of image regions on a prediction, perturbation-based techniques such as LIME and SHAP work by repeatedly perturbing an image in different ways, then measuring the impact on a model’s output~\cite{lime, shap}.
While these methods can produce more fine-grained attribution maps than CAM-based techniques, they are computationally expensive, frequently requiring hundreds of evaluations to produce fine-grained results. Additionally, they are designed to measure the impact of small individual image regions without adequately considering the interactions between regions. For animal re-ID, where identification is based on the relative arrangement of fine-grained patterns across an image (\eg, an animal's stripe pattern), this can fail to capture important information and yield confusing, potentially misleading results, as shown in Figure \ref{fig:methods_ablation}.

For our implementation of Kernel SHAP, we segmented the images using SLIC superpixels~\cite{slic_superpixels, scikit-image}, then applied Kernel SHAP to measure the contribution of each image region~\cite{shap, kokhlikyan2020captum}. We experimented with the number of segments and the number of samples, but we found that this technique did not adequately capture interactions between image regions, and thus did not produce meaningful visualizations. In the main methods ablation, we chose to use 500 segments and 1000 samples.

\section{Methodological details}
\label{supsec:method}

\subsection{Feature matching implementation}
We experimented with different techniques for computing matches between keypoint-descriptor sets, as in Section \ref{sec:deep_feature_matching}, using~\cite{opencv_library}. We found that brute-force matching with cross-checking was both simple and effective.

\subsection{LRP implementation}

To effectively backpropagate relevance, LRP requires a set of rules to be defined for different layer types. For LRP on CNNs, we used the EpsilonPlus composite defined by~\cite{zennit}, which uses the Epsilon rule for linear layers and the ZPlus rule for convolutional layers~\cite{Montavon2019}. To canonize different model architectures, we used~\cite{zennit} and~\cite{xai_canonization}. For LRP on transformer architectures, we used~\cite{achtibat2024attnlrpattentionawarelayerwiserelevance}, though we still struggled to achieve numerical stability on vision transformers.

\subsection{Metric for correct/incorrect separability}\label{sec:metrics_appendix}

\begin{figure}[hbt!]
    \centering
    \captionsetup{aboveskip=16pt, belowskip=-12pt, font=small}
    \includegraphics[width=0.45\textwidth]{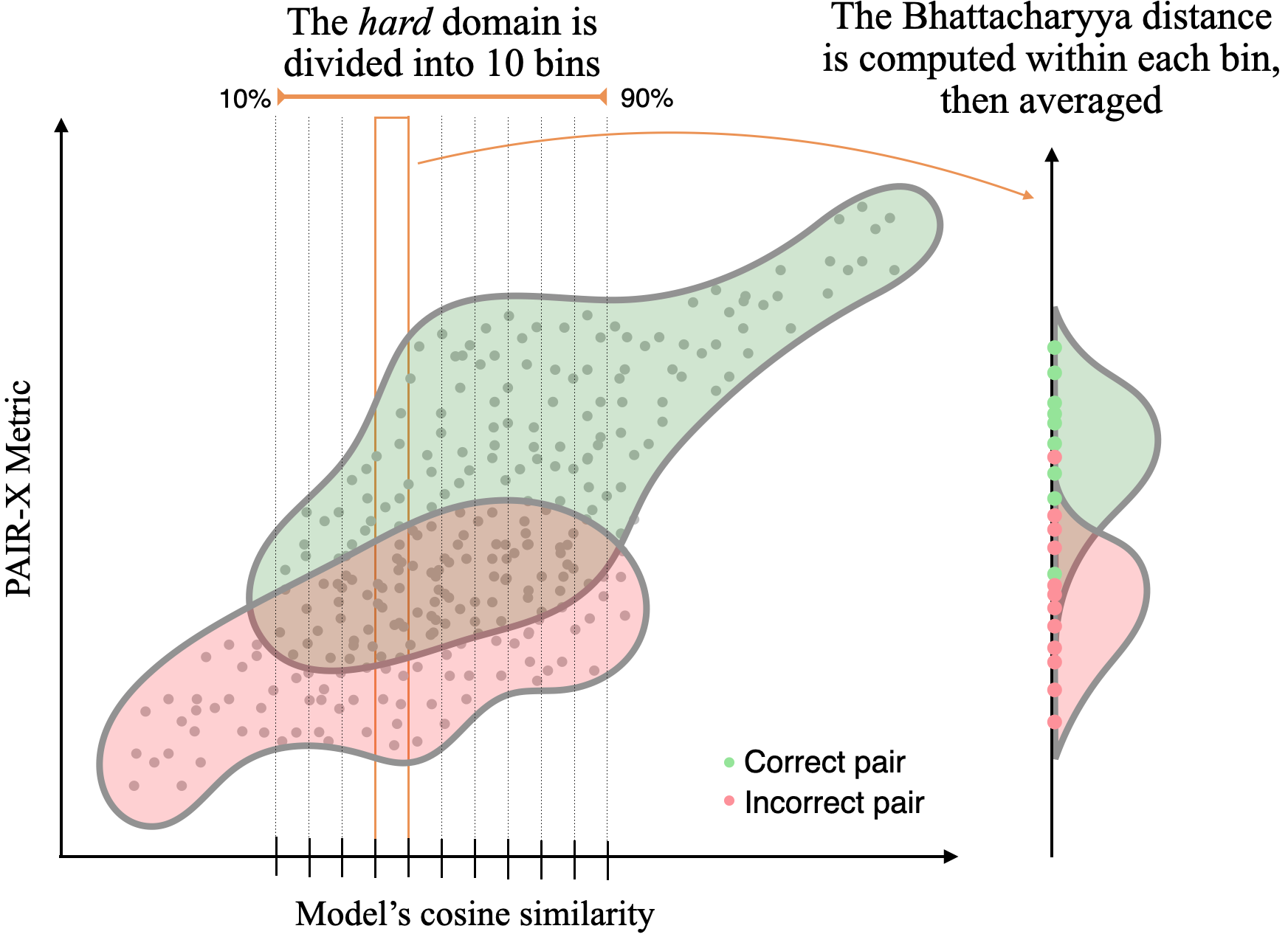}
    \caption{The \textit{hard} domain is divided into 10 bins. Within each bin, the Bhattacharyya distance is computed along the PAIR-X metric, then averaged across the 10 bins with respect to the number of pairs in each bin.}
    \label{fig:metric-pooling}
\end{figure}

As described in Section~\ref{sec:results}, we found it valuable to define a metric assessing the separability of correct and incorrect pairs according to our PAIR-X metrics, while controlling for cosine similarity. The goal of this metric was to assess whether the visualizations typically appeared more plausible for correct than incorrect pairs even when the model scored them similarly, because this would be an important quality when using PAIR-X to validate model predictions. Figure~\ref{fig:metric-pooling} illustrates the high-level idea behing the binned Bhattacharyya distance metric.

The \textbf{Bhattacharyya distance} \cite{Bhattacharyya} measures the overlap between the distributions of the PAIR-X metric for positive pairs ${p_i}$ and negative pairs ${n_i}$, making it useful for evaluating separability of the two distributions in each bin. For positive (\( p \)) and negative (\( n \)) pair distributions, it is given by:

\begin{equation*}
    \text{Bhattacharyya}(p, n) = -\ln \sum_i \sqrt{p_i n_i}
\end{equation*}

To compare the separability of correct and incorrect image pairs across datasets while controlling for the model match score, we first filter to the set of pairs that are hard to classify, by taking the overlap between the middle 95\% of correct pairs and the middle 95\% of incorrect pairs with respect to the cosine similarity. Then, in order to control for the cosine similarity of the image representations, we split this window with respect to cosine similarity into 10 equally-sized bins. Within each bin, we compute KDE with respect to the PAIR-X score for correct and incorrect points, then take the symmetric Bhattacharyya distance between the distributions. Finally, we take an average of these distances, weighted according to the number of points in each bin. This is defined in Algorithm~\ref{alg:binned_bd}.

\begin{algorithm*}
    \small
    \caption{Binned Bhattacharyya distance calculation}
    \label{alg:binned_bd}
    \begin{algorithmic}
        \State $\text{left\_bd} \gets \max(\text{pct}_{2.5}(\text{corr\_cos\_sims}), \text{pct}_{2.5}(\text{incorr\_cos\_sims}))$
        \State $\text{right\_bd} \gets \min(\text{pct}_{97.5}(\text{corr\_cos\_sims}), \text{pct}_{97.5}(\text{incorr\_cos\_sims}))$
        \State bin\_size $\gets$ (\text{right\_bd} - \text{left\_bd}) / 10
        \State
        \State $\text{weighted\_sum} \gets 0$
        \State $\text{points\_counted} \gets 0$
        
        \For{$i = 0$ to $9$}
            \State bin\_start $\gets$ left\_bd + i $\times$ bin\_size
            \State bin\_end $\gets$ bin\_start + bin\_size

            \State $\text{corr\_binned\_pairx\_scores} \gets \{ \text{pairx\_score(pt)} \mid pt \in \text{corr\_pts}, \text{bin\_start} < \text{cos\_sim(pt)} < \text{bin\_end} \}$

            \State $\text{incorr\_binned\_pairx\_scores} \gets \{ \text{pairx\_score(pt)} \mid pt \in \text{incorr\_pts}, \text{bin\_start} < \text{cos\_sim(pt)} < \text{bin\_end} \}$
            \State
            \If{$|\text{corr\_binned\_pairx\_scores}| < 3$ \textbf{or} $|\text{incorr\_binned\_pairx\_scores}| < 3$}
                \State \textbf{continue}
            \EndIf
            \State
            \State $\text{max\_pairx\_score} \gets \max(\max(\text{corr\_binned\_pairx\_scores}), \max(\text{incorr\_binned\_pairx\_scores}))$
            \State $\text{min\_pairx\_score} \gets \min(\min(\text{corr\_binned\_pairx\_scores}), \min(\text{incorr\_binned\_pairx\_scores}))$
            \State
            \State $\text{corr\_kde\_values} \gets \text{kernel\_density\_estimation}(\text{corr\_binned\_pairx\_scores}, \text{min\_pairx\_score}, \text{max\_pairx\_score})$
            \State $\text{incorr\_kde\_values} \gets \text{kernel\_density\_estimation}(\text{incorr\_binned\_pairx\_scores}, \text{min\_pairx\_score}, \text{max\_pairx\_score})$
            \State
            \State $\text{bd} \gets \text{bhattacharyya\_distance}(\text{corr\_kde\_values}, \text{incorr\_kde\_values})$
            \State
            \If{$\text{mean}(\text{corr\_binned\_pairx\_scores}) < \text{mean}(\text{incorr\_binned\_pairx\_scores})$}
                \State $\text{bd} \gets -\text{bd}$
            \EndIf
            \State
            \State $\text{weighted\_sum} \gets \text{weighted\_sum} + \text{bd} \times (|\text{corr\_binned\_pairx\_scores}| + |\text{incorr\_binned\_pairx\_scores}|)$
            \State $\text{points\_counted} \gets \text{points\_counted} + |\text{corr\_binned\_pairx\_scores}| + |\text{incorr\_binned\_pairx\_scores}|$
        \EndFor
        \State
        \State \Return \text{weighted\_sum} / \text{points\_counted}
    \end{algorithmic}
\end{algorithm*}

\subsection{Pair and layer selection for quantitative metrics}
\label{sec:pair_selection_appendix}

For our quantitative evaluation, we were interested in comparing pairs that the model scored highly, both for correct and incorrect matches. To construct a set of pairs to select from, we used our test set as query images and the train set as our gallery. We took the top $k$ matching gallery images for each query image, producing a set of pairs to select from, then randomly sampled 1000 correct pairs and 1000 incorrect pairs. In general, we set $k=5$, but for smaller datasets where this did not produce enough pairs, we iteratively increased $k$ until the number of pairs was sufficient. To limit the analysis to high-scoring pairs, we capped $k$ at $20$. For very small datasets where this was still not enough to produce 2000 pairs, we used a smaller sample size.

The optimal layer for evaluation also varied across datasets, primarily based on the resolution of meaningful local features. A visual analysis of layer choice is presented in Figure \ref{fig:layer_ablation}. To select the layer used for quantitative evaluation for each dataset, we used a smaller random sample of 500 pairs from the training set (using the train set as both query and gallery), then selected the layer with the best value for $\rho_{res}$. We chose to use the inverted residual mean metric to select layers rather than the match coverage metric because it was not as affected by the resolution of the feature maps. In later layers, where feature maps have coarser resolution, the match coverage score was typically slightly better, simply because there were fewer features to match.

\section{Additional qualitative results}

Various qualitative trends in the performance of PAIR-X on different datasets are demonstrated and described in Figure \ref{fig:species_ablation}. Generally, we saw strong performance on patterned or otherwise individually marked species. For species with a high degree of structural similarity (\eg, face-based species like dogs, cats, or lions), we saw that the PAIR-X results did explain model predictions, but produced potentially misleading results for incorrect matches. This is also demonstrated in \ref{fig:dog_examples}.

\subsection{Additional results and analysis on buildings}\label{sec:oxford5k}

\begin{figure}[h!]
    \centering
    \includegraphics[width=\columnwidth]{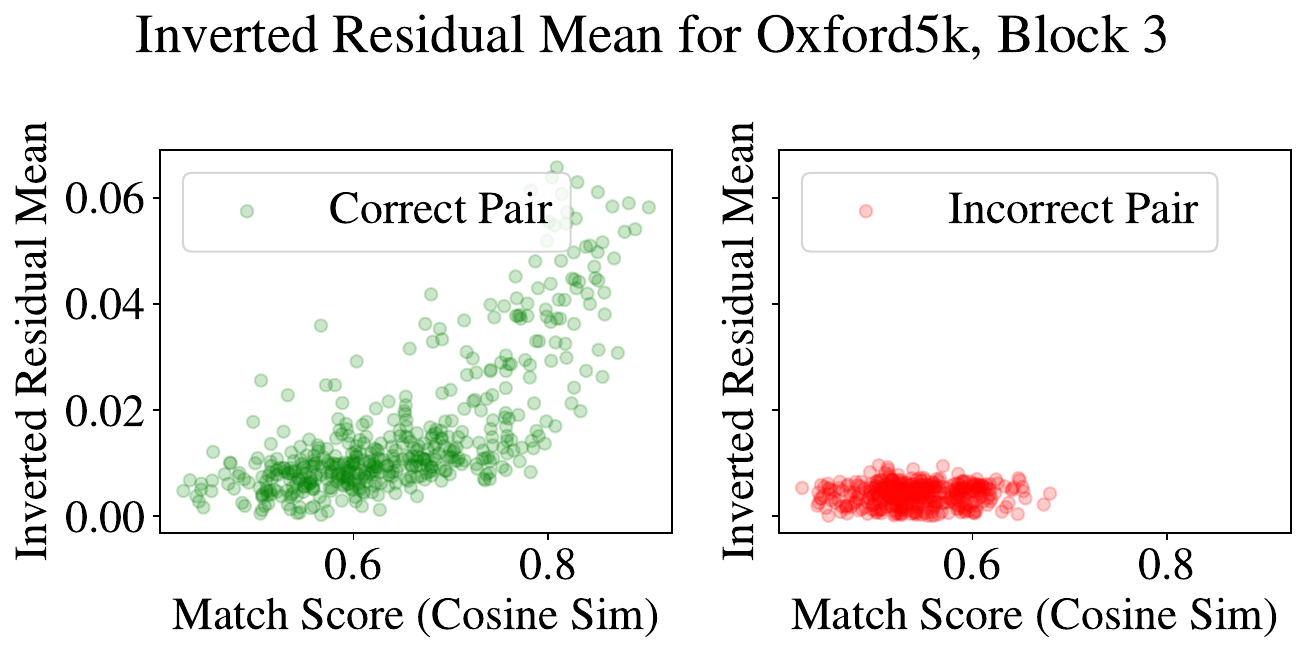}
    \caption{Distribution of PAIR-X scores for correct and incorrect image pairs in the Oxford-5k building dataset}
    \label{fig:enter-label}
\end{figure}

\subsection{Choice of intermediate model layer}
\captionsetup[subfloat]{labelformat=empty}
\begin{figure*}
    \centering
    \subfloat[Block 1]{\ToggleImages[width=0.19\textwidth]{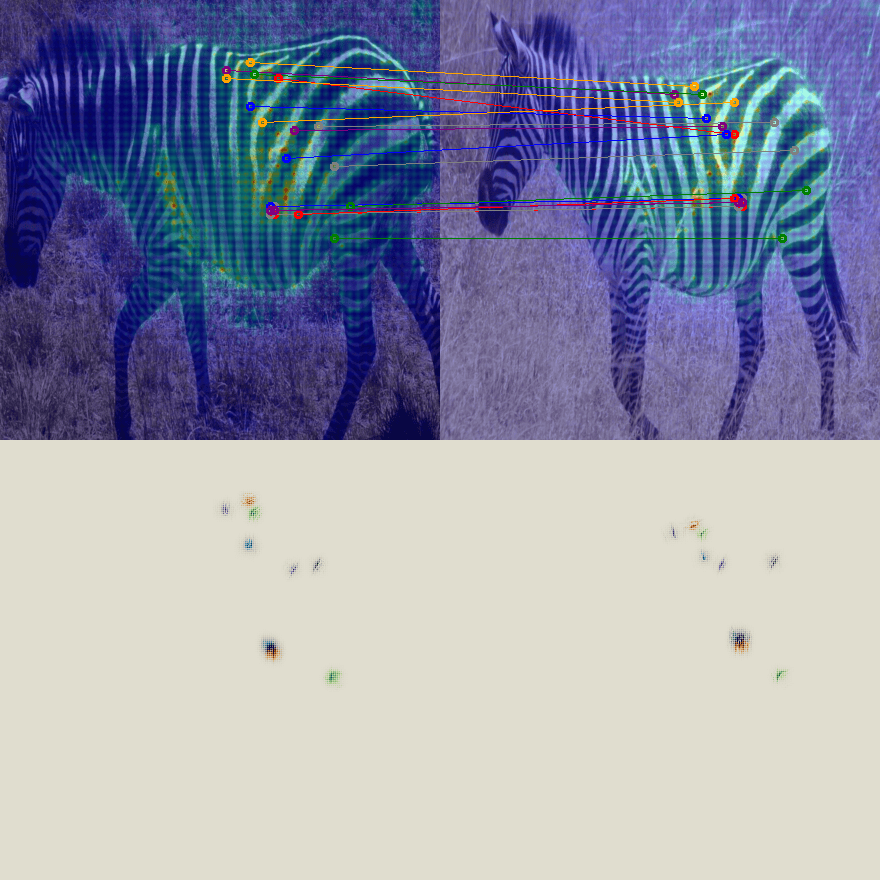}{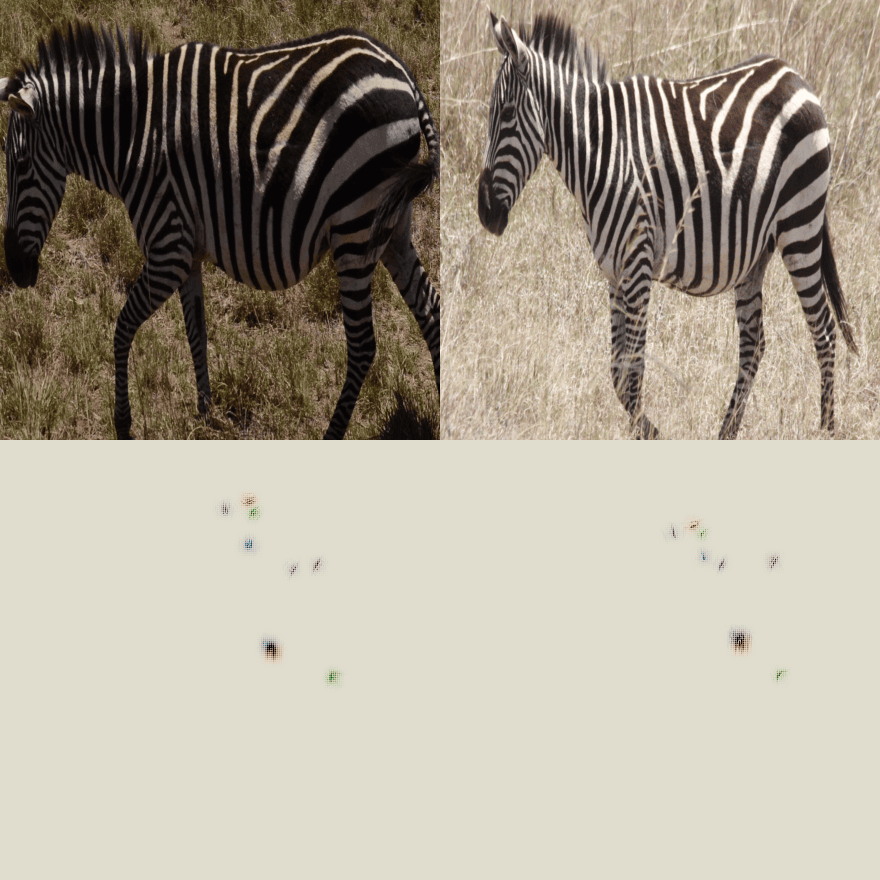}} \hspace{0.005\textwidth}
    \subfloat[Block 2]{
    \ToggleImages[width=0.19\textwidth]{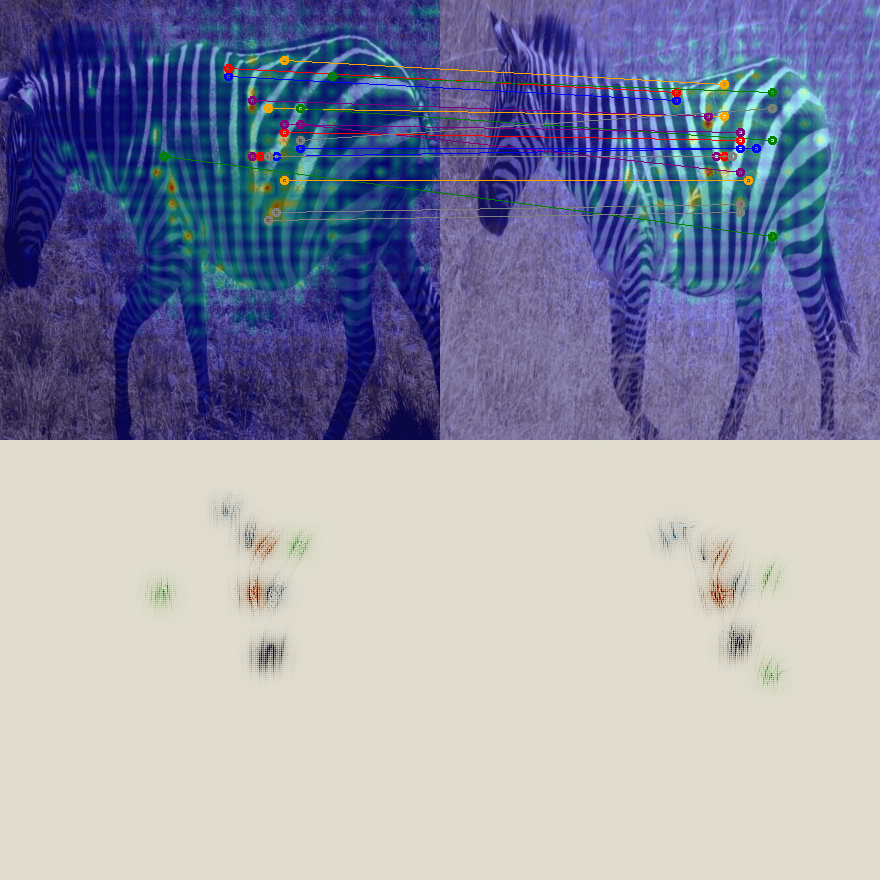}{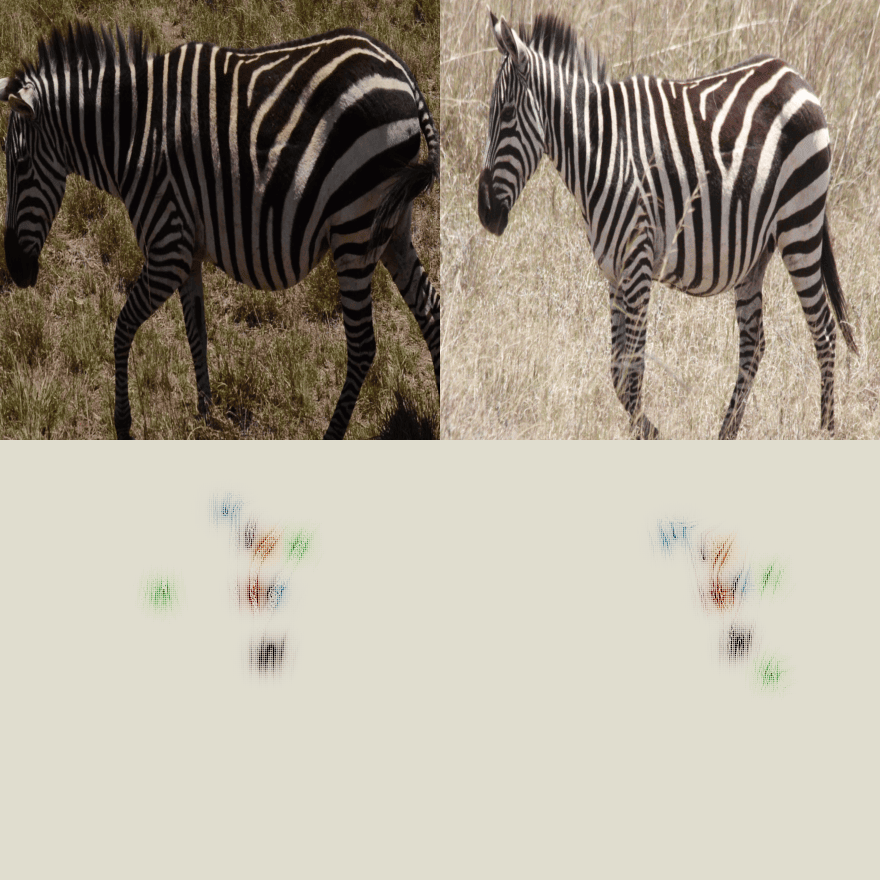}} \hspace{0.005\textwidth}
    \subfloat[Block 3]{\ToggleImages[width=0.19\textwidth]{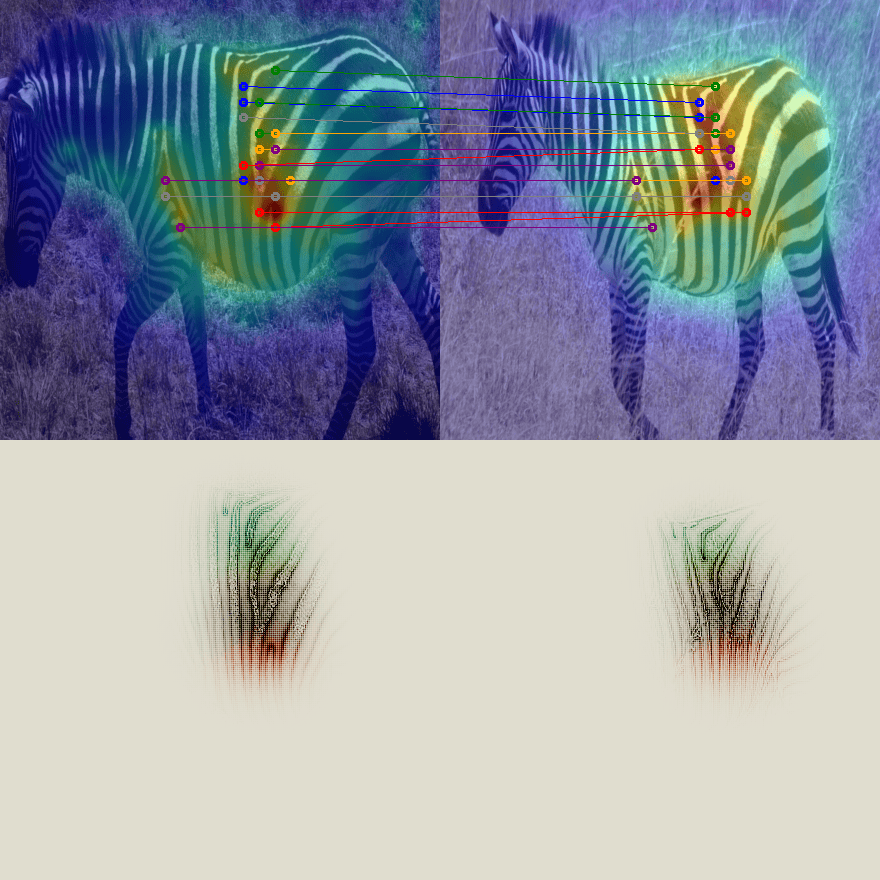}{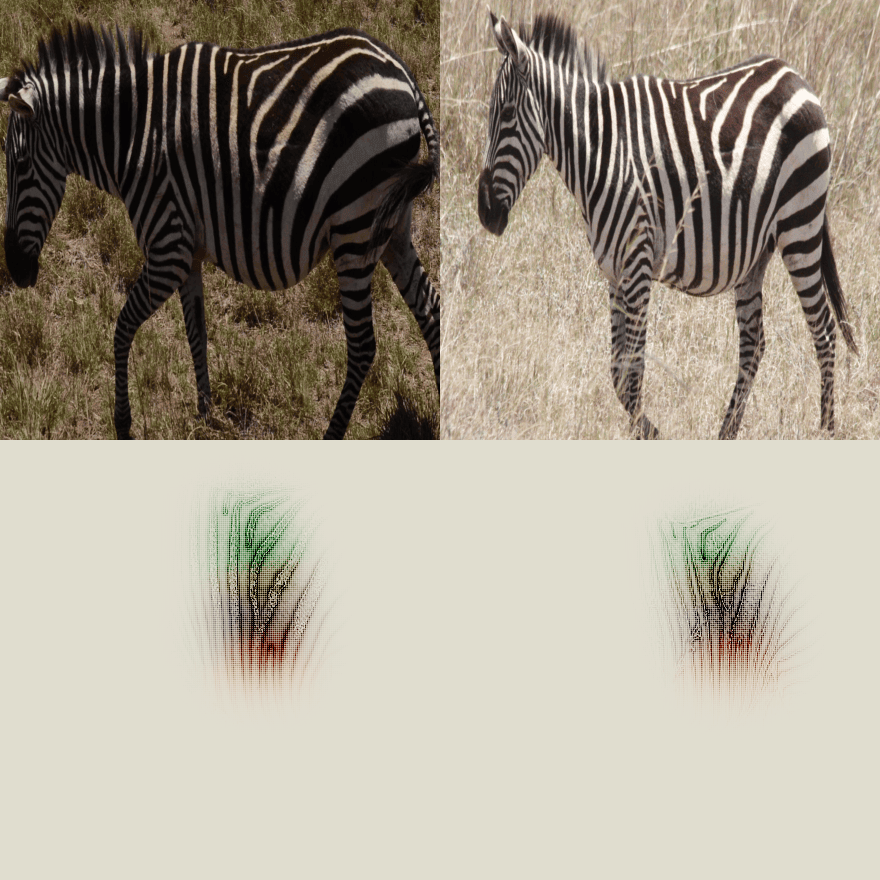}} \hspace{0.005\textwidth}
    \subfloat[Block 4]{\ToggleImages[width=0.19\textwidth]{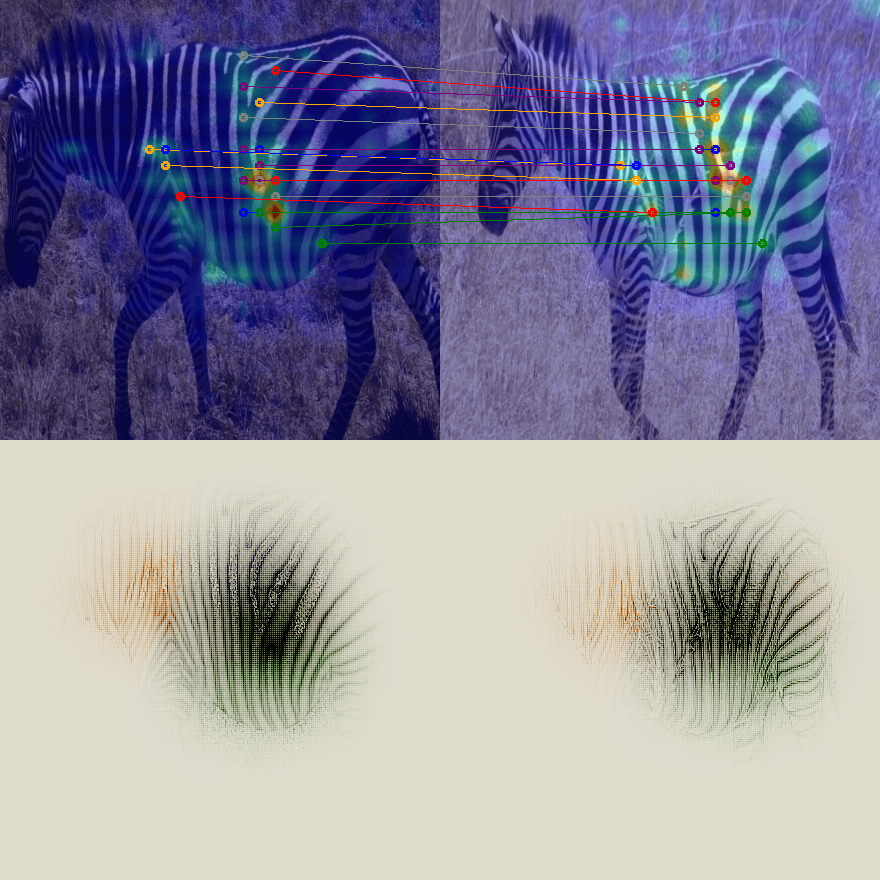}{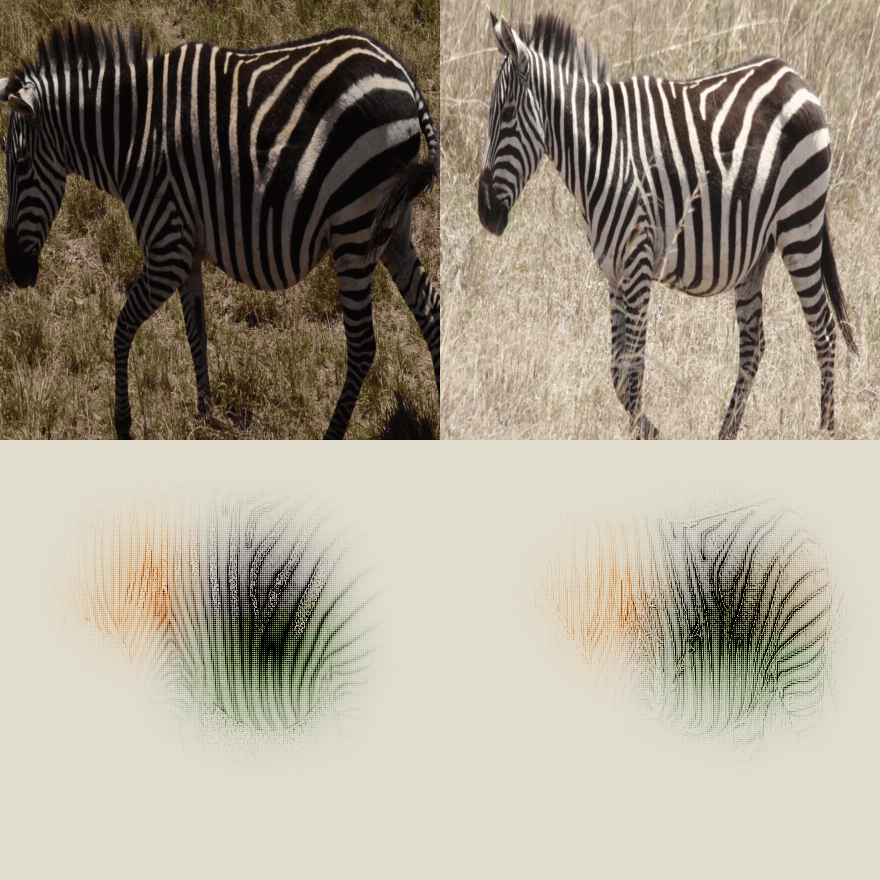}}
    \hspace{0.005\textwidth}
    \subfloat[Block 5]{\ToggleImages[width=0.19\textwidth]{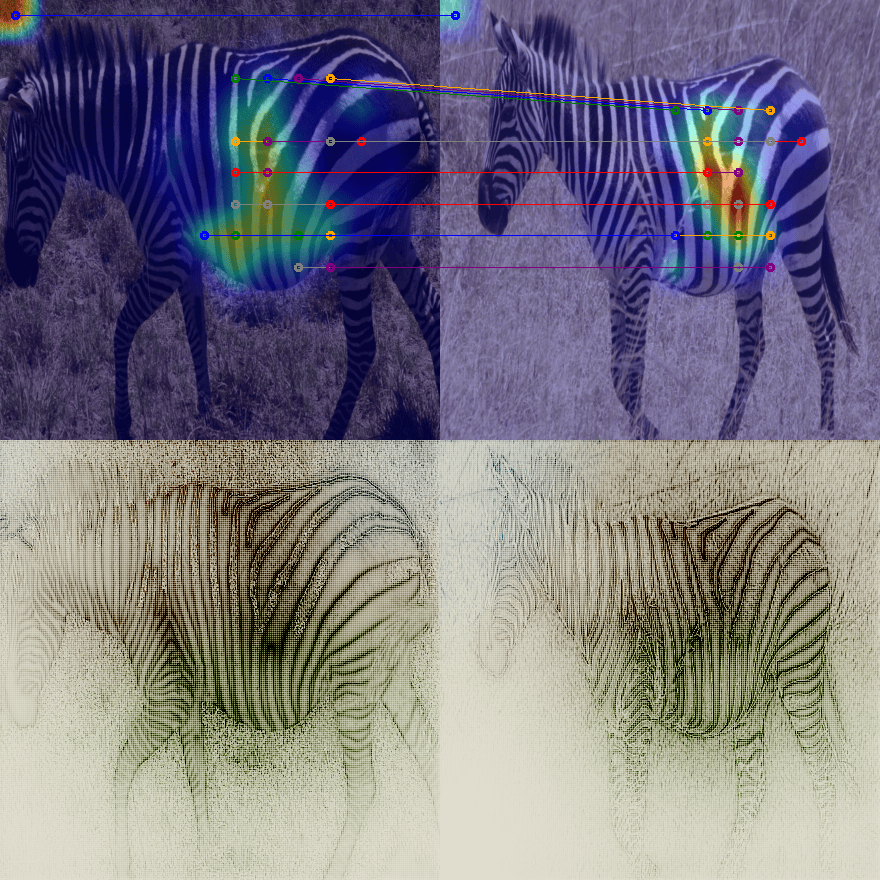}{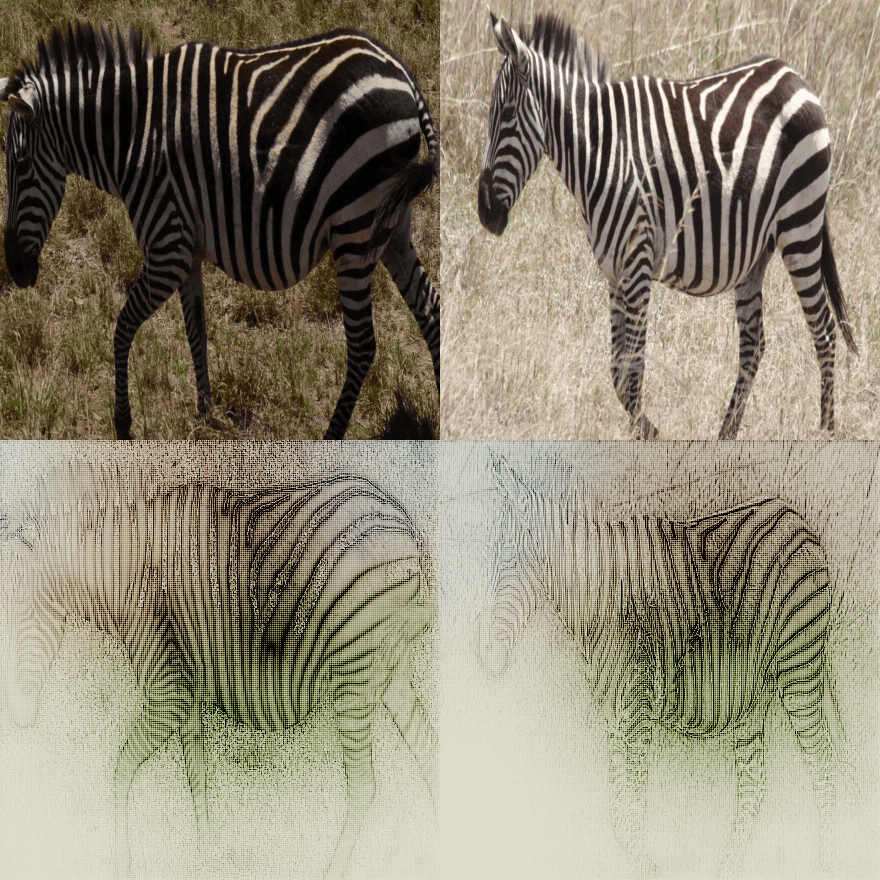}}

     \begin{tikzpicture}[overlay, remember picture]
        \draw node[below right, xshift=-0.5\textwidth, yshift=-1.3] {Shallow};
        \draw node[below left, xshift=0.5\textwidth, yshift=-1.3] {Deep};
        \draw[thick, ->] (-0.5\textwidth,-0.05) -- (0.5\textwidth,-0.05);
    \end{tikzpicture}

    \vspace{.7cm}
    
    \caption{\textbf{As we move deeper into the model, the receptive field incorporates more spatially distant information.} We find that layers close to the middle of the model provide the most qualitatively useful visualizations.}
    \label{fig:layer_ablation}
\end{figure*}

As shown in Figure \ref{fig:layer_ablation}, we see that the choice of intermediate layer has a significant impact on the results of the visualization. While the optimal choice of layer varies between datasets, we generally found layers close to the middle of the model to offer the best qualitative results. According to the metrics defined in Section \ref{sec:quantitative_metrics}, the later layers in a model typically offer the best performance. Because they incorporate more spatial context, these layers are more likely to produce accurate matches.

\section{Additional quantitative results}\label{supsec:quantitative}
In addition to the multispecies Miew-ID model, we compared against a CNN version of MegaDescriptor~\cite{cermak2023wildlifedatasetsopensourcetoolkitanimal, otarashvili2024multispeciesanimalreidusing}. The results of this comparison were similar to the results with multispecies Miew-ID, with small variations across datasets, and are presented in Table \ref{tab:megadescriptor_ablation}.

\begin{table}
\begin{threeparttable}
\caption{Quantitative Metrics For MegaDescriptor CNN Model}
\label{tab:megadescriptor_ablation}

\begin{tabular}{c|c c c c}
    \toprule
    \textbf{Dataset} & $\rho_{res}$ & $\Delta_{res}$ & $\rho_{mc}$ & $\Delta_{mc}$\\
    \midrule
AAUZebraFish~\cite{Haurum_2020_zebra_fish} & 0.66 & \cellcolor{green!100}1.14 & 0.73 & \cellcolor{green!62}0.63 \\
ATRW~\cite{ATRW} & 0.52 & \cellcolor{green!14}0.15 & 0.66 & \cellcolor{green!7}0.07 \\
BelugaIDv2~\cite{BelugaIDv2} & 0.08 & \cellcolor{red!0}-0.01 & 0.08 & \cellcolor{green!1}0.02 \\
BirdIndividualID~\cite{BirdIndividualID} & 0.70 & \cellcolor{green!13}0.14 & 0.74 & \cellcolor{green!12}0.13 \\
CTai~\cite{CTai} & 0.16 & \cellcolor{green!2}0.03 & 0.26 & \cellcolor{green!0}0.01 \\
CZoo~\cite{CTai} & 0.35 & \cellcolor{green!0}0.00 & 0.50 & \cellcolor{green!0}0.00 \\
CatIndividualImages~\cite{CatIndividualImages} & 0.76 & \cellcolor{green!55}0.55 & 0.78 & \cellcolor{green!62}0.63 \\
CowDataset~\cite{CowDataset} & 0.49 & \cellcolor{green!70}0.70 & 0.48 & \cellcolor{green!66}0.66 \\
Cows2021v2~\cite{Cows2021} & 0.72 & \cellcolor{green!0}0.00 & 0.78 & \cellcolor{green!0}0.00 \\
DogFaceNet~\cite{DogFaceNet} & 0.27 & \cellcolor{green!61}0.62 & 0.30 & \cellcolor{green!58}0.59 \\
ELPephants~\cite{ELPephants} & 0.24 & \cellcolor{green!16}0.16 & 0.26 & \cellcolor{green!6}0.07 \\
FriesianCattle2015v2~\cite{FriesianCattle2015v2} & 0.71 & \cellcolor{green!43}0.44 & 0.76 & \cellcolor{green!22}0.23 \\
FriesianCattle2017~\cite{FriesianCattle2017} & 0.63 & \cellcolor{red!12}-0.12 & 0.77 & \cellcolor{red!5}-0.06 \\
GiraffeZebraID~\cite{GiraffeZebraID} & 0.00 & \cellcolor{green!4}0.05 & -0.00 & \cellcolor{green!4}0.04 \\
Giraffes~\cite{Giraffes} & 0.62 & \cellcolor{green!46}0.46 & 0.63 & \cellcolor{green!34}0.34 \\
HyenaID2022~\cite{HyenaID2022} & 0.37 & \cellcolor{green!21}0.22 & 0.43 & \cellcolor{green!17}0.17 \\
IPanda50~\cite{IPanda50} & 0.19 & \cellcolor{green!2}0.03 & 0.34 & \cellcolor{green!3}0.04 \\
LeopardID2022~\cite{LeopardID2022} & 0.52 & \cellcolor{green!11}0.12 & 0.52 & \cellcolor{green!10}0.11 \\
LionData~\cite{LionData} & 0.12 & \cellcolor{red!5}-0.05 & 0.39 & \cellcolor{green!4}0.04 \\
MacaqueFaces~\cite{MacaqueFaces} & 0.19 & \cellcolor{green!3}0.03 & 0.46 & \cellcolor{green!3}0.04 \\
NDD20v2~\cite{NDD20v2} & 0.25 & \cellcolor{green!10}0.10 & 0.34 & \cellcolor{green!5}0.06 \\
NOAARightWhale~\cite{NOAARightWhale} & 0.32 & \cellcolor{green!26}0.26 & 0.45 & \cellcolor{green!23}0.23 \\
NyalaData~\cite{NyalaData} & 0.19 & \cellcolor{green!7}0.07 & 0.12 & \cellcolor{green!3}0.04 \\
OpenCows2020~\cite{Andrew_2021} & 0.64 & \cellcolor{green!5}0.05 & 0.73 & \cellcolor{green!2}0.02 \\
ReunionTurtles~\cite{ReunionTurtles} & 0.29 & \cellcolor{green!22}0.22 & 0.41 & \cellcolor{green!41}0.42 \\
SMALST~\cite{SMALST} & 0.34 & \cellcolor{red!19}-0.19 & 0.44 & \cellcolor{red!1}-0.01 \\
SeaStarReID2023~\cite{SeaStarReID2023} & 0.38 & \cellcolor{green!44}0.44 & 0.37 & \cellcolor{green!30}0.31 \\
SeaTurtleID2022~\cite{SeaTurtleID2022} & 0.26 & \cellcolor{green!25}0.26 & 0.41 & \cellcolor{green!15}0.15 \\
SeaTurtleIDHeads~\cite{SeaTurtleID2022} & 0.29 & \cellcolor{green!25}0.25 & 0.40 & \cellcolor{green!17}0.18 \\
SealID~\cite{SealID} & 0.35 & \cellcolor{green!100}1.04 & 0.42 & \cellcolor{green!48}0.48 \\
SouthernProvinceTurtles~\cite{SouthernProvinceTurtles} & 0.49 & \cellcolor{green!36}0.36 & 0.47 & \cellcolor{green!28}0.29 \\
WhaleSharkID~\cite{WhaleSharkID} & 0.16 & \cellcolor{green!16}0.16 & 0.31 & \cellcolor{green!2}0.03 \\
ZakynthosTurtles~\cite{ZakynthosTurtles} & 0.50 & \cellcolor{green!37}0.38 & 0.61 & \cellcolor{green!14}0.15 \\
ZindiTurtleRecall~\cite{ZindiTurtleRecall} & 0.28 & \cellcolor{green!13}0.13 & 0.37 & \cellcolor{green!9}0.09 \\

    \bottomrule
\end{tabular}

\smallskip
\begin{flushleft} 
\begin{tablenotes}
\small
\raggedright
    \item Inverted residual mean ($res$) and relevance-weighted match coverage ($mc$) across datasets, aggregated within each dataset using both Spearman's rank correlation coefficient ($\rho$) between each metric and model match score, and binned Bhattacharyya distance ($\Delta$) of each metric between correct and incorrect matches. \colorbox{green}{Positive $\Delta$ values} indicate that PAIR-X improves separation for similarly scored matches.
\end{tablenotes}
\end{flushleft} 
\end{threeparttable}
\end{table}

\section{Computational efficiency}\label{sec:computational_efficiency}
In general, PAIR-X is slightly more expensive than techniques such as Grad-CAM, which require a single partial backpropagation, but it is much cheaper than perturbation-based techniques such as LIME and SHAP. The first stage of PAIR-X, feature matching, requires one full forward pass, where model activations are saved off at the selected intermediate layer $l$. The cost of the matching operation itself is negligible. The second stage of PAIR-X, where matches are filtered according to intermediate relevances, requires one backward pass, where the relevance scores for the intermediate layer are saved off. Finally, the third stage of PAIR-X, where pixel-wise relevance scores are computed to create color maps, requires $k$ partial backward passes (from the intermediate layer to the original input), where $k$ is the number of matches included in the color map.

We do not find these costs to be prohibitive, even in cases where users are running PAIR-X without a GPU. For the multispecies Miew-ID model, which uses an EfficientNet backbone with ~51M parameters, we find that running PAIR-X on the third intermediate layer with 10 backpropagated matches took about 5 seconds on GPU and 17 seconds on CPU. Results are likely to vary according to model architecture, selected intermediate layer, computational resources, and number of matches, but these experiments demonstrate the feasibility of using PAIR-X at scale.

\section{Experiments on Vision Transformers}\label{supsec:transformers}

\begin{figure}
    \subfloat[Zebras]{\includegraphics[width=.49\columnwidth]{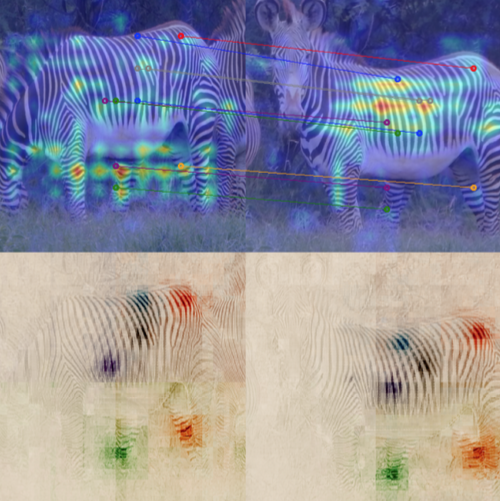}}
    \subfloat[Buildings]{\includegraphics[width=.49\columnwidth]{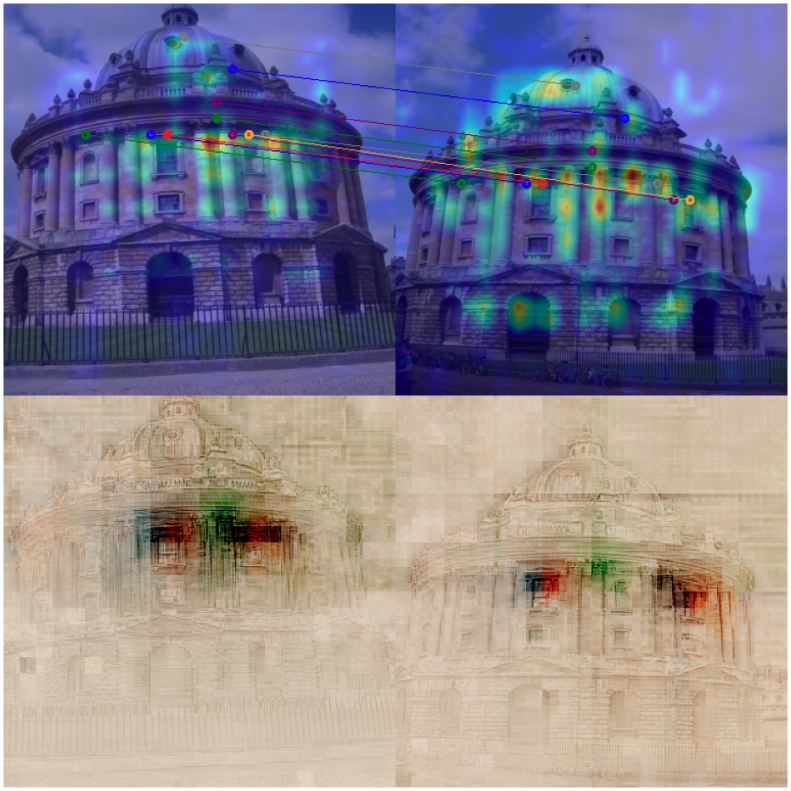}}

    \caption{Preliminary PAIR-X results on a SwinTransformer model show roughly correct matches, but with lower precision than for CNNs, likely due to the nonlocality of transformer receptive fields. We also observe significant block artifacts in the receptive field visualizations, stemming from the model's patch-based architecture. Additionally, the heatmaps retain artifacts and often fail to highlight identifiable features, which we attribute to limitations in our current LRP implementation for transformers.}
    \label{fig:swin_examples}
\end{figure}

Preliminary experiments on vision transformers, such as the multispecies MegaDescriptor SwinTransformer model~\cite{cermak2023wildlifedatasetsopensourcetoolkitanimal}, suggest that PAIR-X shows some promise for these architectures, but requires further development to match its performance on CNNs. Figure~\ref{fig:swin_examples} shows PAIR-X results on MegaDescriptor for a zebra and a building example, illustrating current capabilities and highlighting areas for future improvement.

A primary challenge in adapting PAIR-X to Vision Transformers is LRP, which was initially developed for CNNs. While there has been work on expanding LRP to transformers, the structures for doing so are much less established than for CNNs.~\cite{achtibat2024attnlrpattentionawarelayerwiserelevance} presents a strong framework for the application of LRP to transformers. They find that noisy attributions and gradient-shattering are a significant problem for ViTs, and address this by using the $\gamma$-LRP rule~\cite{Montavon2019}, with manual tuning for the $\gamma$ hyperparameter. However, they acknowledge that the manual tuning of this hyperparameter is critical to produce accurate attributions, and an ongoing limitation. In our own experiments, we found it difficult to produce faithful explanations while preventing gradient-shattering. Second, ViT features are inherently nonlocal, incorporating information from spatially distant locations even at early model layers. While we found reasonably promising results via simple spatial decomposition into descriptor vectors (like with CNNs), we found that the feature matches were not quite as faithful, likely due to this nonlocality. Other work on deriving localized features from ViTs has sought to address this problem via the incorporation of information from neighbor feature patches~\cite{amir2022deepvitfeaturesdense}, which might help to increase match fidelity. There remains room for future work to address each of these challenges.

\end{document}